\documentclass[Afour,sageh,times]{sagej}

\usepackage{moreverb,url}
\usepackage{mathtools}

\usepackage[colorlinks,bookmarksopen,bookmarksnumbered]{hyperref}
\usepackage{algorithm}
\usepackage{algpseudocode}
\usepackage[table,xcdraw]{xcolor}
\usepackage{subcaption}
\usepackage{multirow}
\usepackage{multicol}
\usepackage{tabularx}
\usepackage{siunitx}
\usepackage{prettyref}

\newrefformat{fig}{Figure~\ref{#1}}
\newrefformat{sec}{Section~\ref{#1}}
\newrefformat{tab}{Table~\ref{#1}}
\newrefformat{alg}{Algorithm~\ref{#1}}
\newrefformat{eq}{Equation~\ref{#1}}
\newrefformat{apd}{Appendix~\ref{#1}}

\newcommand{\pluseq}{\mathrel{+}=}
\newcommand{\textnameSMAT}{\text{S$^2$MAT}}

\newcommand\BibTeX{{\rmfamily B\kern-.05em \textsc{i\kern-.025em b}\kern-.08em
T\kern-.1667em\lower.7ex\hbox{E}\kern-.125emX}}

\def\volumeyear{2023}

\begin{document}

\runninghead{\textnameSMAT{}}

\title{\textnameSMAT{}: Simultaneous and Self-Reinforced Mapping and Tracking in Dynamic Urban Scenarios}

\author{Tingxiang Fan\affilnum{1*}, Bowen Shen\affilnum{2*}, Yinqiang Zhang\affilnum{1*}, Chuye Zhang\affilnum{2}, Lei Yang\affilnum{1}, Hua Chen\affilnum{2}, Wei Zhang\affilnum{2} and Jia Pan\affilnum{1}}

\affiliation{* These authors contributed equally. \\
\affilnum{1}Department of Computer Science, the University of Hong Kong, Hong Kong, China \\
\affilnum{2}Department of Mechanical and Energy Engineering, Southern University of Science and Technology, China}

\corrauth{Jia Pan (the University of Hong Kong), Wei Zhang (Southern University of Science and Technology)}
\email{jpan@cs.hku.hk}

\begin{abstract}
Despite the increasing prevalence of robots in daily life, their navigation capabilities are still limited to environments with prior knowledge, such as a global map. To fully unlock the potential of robots, it is crucial to enable them to navigate in large-scale unknown and changing unstructured scenarios. This requires the robot to construct an accurate static map in real-time as it explores, while filtering out moving objects to ensure mapping accuracy and, if possible, achieving high-quality pedestrian tracking and collision avoidance.
While existing methods can achieve individual goals of spatial mapping or dynamic object detection and tracking, there has been limited research on effectively integrating these two tasks, which are actually coupled and reciprocal. In this work, we propose a solution called \textnameSMAT{} (Simultaneous and Self-Reinforced Mapping and Tracking) that integrates a front-end dynamic object detection and tracking module with a back-end static mapping module. \textnameSMAT{} leverages the close and reciprocal interplay between these two modules to efficiently and effectively solve the open problem of simultaneous tracking and mapping in highly dynamic scenarios. 
We conducted extensive experiments using widely-used datasets and simulations, providing both qualitative and quantitative results to demonstrate \textnameSMAT{}'s state-of-the-art performance in dynamic object detection, tracking, and high-quality static structure mapping. Additionally, we performed long-range robotic navigation in real-world urban scenarios spanning over \SI{7}{km}, which included challenging obstacles like pedestrians and other traffic agents. The successful navigation provides a comprehensive test of \textnameSMAT{}'s robustness, scalability, efficiency, quality, and its ability to benefit autonomous robots in wild scenarios without pre-built maps.
\end{abstract}


\maketitle

\section{Introduction}
\label{sec:intro}

\begin{figure}[ht]
\centering
\includegraphics[width=1.0\linewidth]{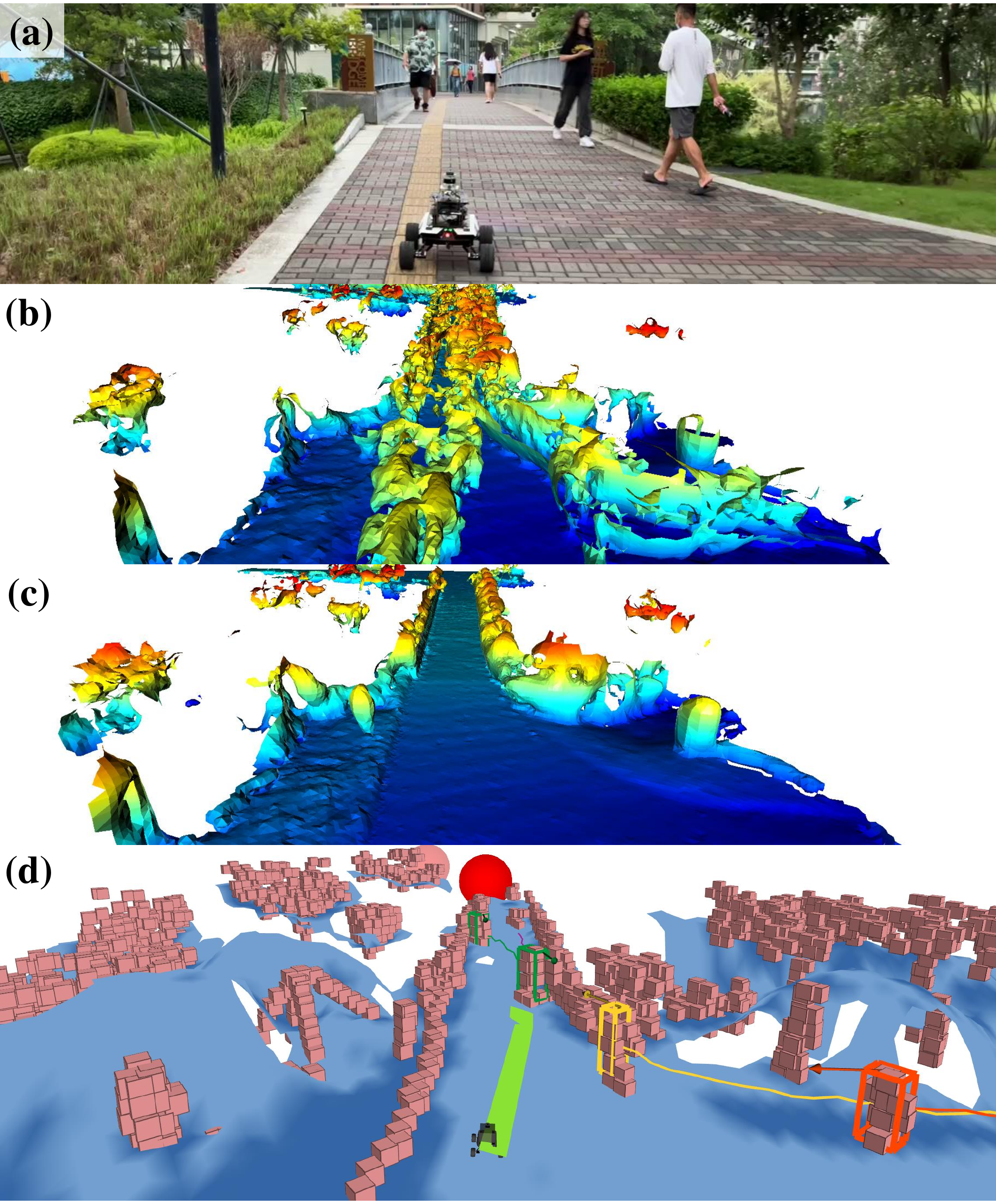}
\caption{(a) A robot is crossing a bridge filled with pedestrians. (b) State-of-the-art SLAM methods \protect \cite{zhang2014loam, shan2020lio, xu2021fast} fail to filter out dynamic points and produce a map contaminated with pedestrian points, rendering the bridge non-traversable. (c) Our \textnameSMAT{} algorithm accurately recovers the bridge structure and generates a high-quality drivable map. (d) It also accurately tracks pedestrians (boxes). The robot then can plan a safe path (green) and move towards a local goal (red ball).}
\label{fig: bridge-passing}
\end{figure}

The ability to navigate efficiently and effectively through human-populated environments has long been a goal of intelligent robotics. This is because humans can seamlessly blend into or move against pedestrian flow without much disturbance. Therefore, robots, aiming to mimic human behavior and intelligence, are expected to possess a similar skill. In addition to this biomimetic contribution, enabling a robot to navigate smoothly through a crowd also has important applications in the service robot industry, which eagerly awaits a breakthrough in developing a product that can robustly function in urban scenarios such as halls, classrooms, or busy streets~\cite{manyika2017jobs}. 

Existing robot navigation systems typically require detailed prior knowledge of the environment's stable features, known as the global map. This includes the location, geometry, and semantics of landmarks, static obstacles, and traversable areas. The global map is crucial for a robot to infer its own position and make sophisticated decisions, such as trajectory planning and collision avoidance, when navigating in a dynamic scenario. Unfortunately, constructing and maintaining such a global map for a large region is time-consuming, resource-intensive, and must be conducted during off-peak periods or when human-populated scenarios are closing, because it is designed for mapping open scenarios with only static obstacles or a few dynamic obstacles. Consequently, the global map cannot be updated in time for ever-changing urban regions, and this outdated map can lead to safety issues during autonomous navigation. For example, the robot may encounter a road construction site or collide with temporary market stalls. To achieve safe and long-horizon navigation in unknown and large-scale urban scenarios, it would be desirable for the robot to navigate without a prior global map but instead conduct an \emph{online and reliable understanding of the environment’s stable features in human-populated scenarios}, as shown in \prettyref{fig: bridge-passing}a. 

Though state-of-the-art Simultaneous Localization and Mapping (SLAM) methods~\cite{zhang2014loam,shan2020lio,xu2021fast} can already achieve high-quality online mapping of unknown environments, their performance in human-populated urban scenarios is poor due to their assumption that the scene is \textit{static}. Specifically, the presence of dynamic objects can occlude or confuse the robot's perception of important spatial structural features, resulting in mistakes when the robot infers its own position, identifies traversable areas, and explores frontiers (see Figure \ref{fig: bridge-passing}b for an example). Additionally, the complex background of urban landmarks makes it difficult for the robot to reliably track moving objects, presenting challenges for collision avoidance and human-robot collaboration. Interestingly, these two difficulties are closely \emph{coupled}, creating a \emph{chicken-and-egg problem}. On one hand, if the robot can accurately \emph{distinguish dynamic objects from static objects} in the scene, the robot can avoid confusing dynamic objects with static ones, and can track dynamic objects better. On the other hand, if the robot can reliably \emph{track all dynamic objects}, it can filter out the dynamic objects from the raw sensory data and only use static data for high-quality online mapping of stable spatial structures.

However, most existing methods focus solely on either separating dynamic and static objects or tracking dynamic objects, with very few attempts to address both challenges simultaneously. For example, some methods remove dynamic objects from raw scan data by identifying inconsistencies among multiple observations~\cite{hornung2013octomap, schauer2018peopleremover, pomerleau2014long, yoon2019mapless, kim2020remove}. However, they do not effectively utilize motion information from dynamic objects for robust removal, often overly eliminating points associated with dynamic objects. Consequently, there is a significant chance of erroneously removing static point clouds that are crucial for robot navigation, such as those corresponding to traversal regions. Conversely, other methods focus on detecting and tracking dynamic objects based on appearance models~\cite{cortinhal2020salsanext, milioto2019rangenet++, chen2021lidar-mos} or motion clues \cite{kaestner2012generative, dewan2016rigid, yan2017online}, and then eliminate the dynamic object point clouds for online mapping. However, none of them can guarantee perfect detection and tracking of dynamic objects in all situations, and even a small number of missed or false detections can significantly degrade the subsequent mapping quality.

\begin{figure*}
\centering
\includegraphics[width=1.0\linewidth]{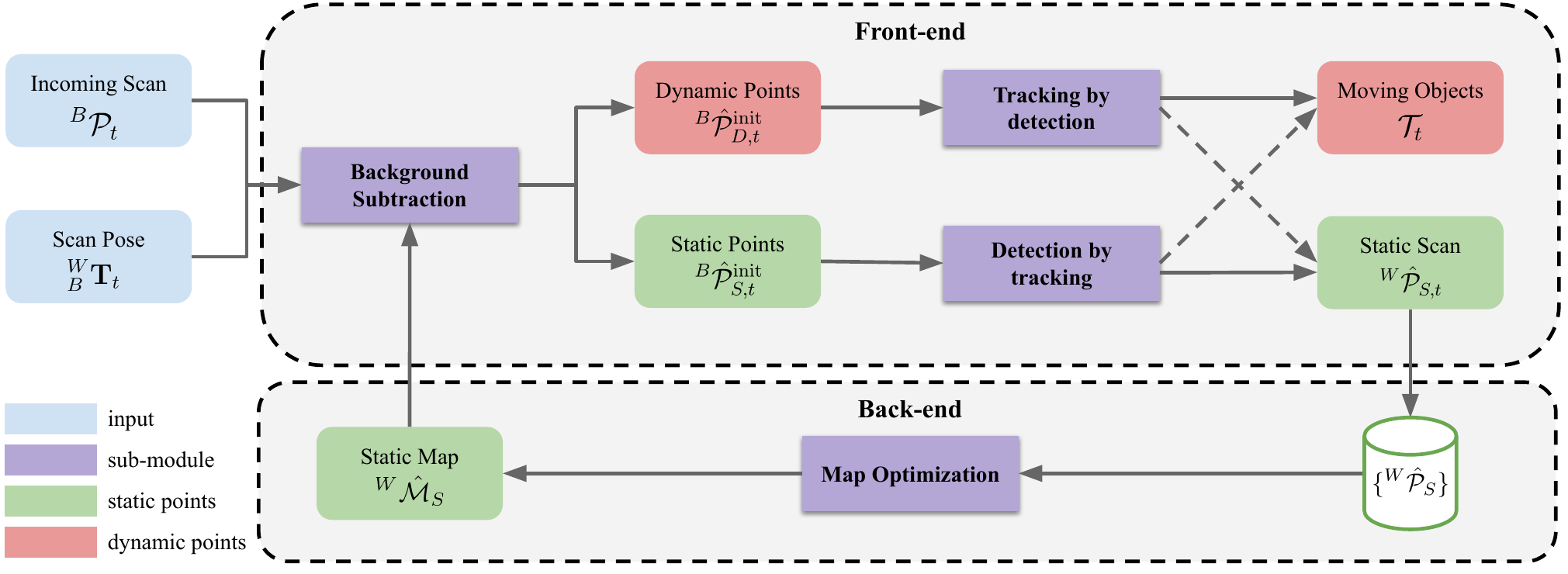}
\caption{\textnameSMAT{} overview. \textnameSMAT{} consists of two modules that interact with and reinforce each other. The front-end module is responsible for detecting and tracking dynamic objects in the incoming scan. It begins by subtracting the static map generated by the back-end module from the scan, resulting in two sets of points: static and dynamic. To enhance the accuracy of this separation, a multi-object tracking system is applied. However, the static scan generated by the front-end may still contain dynamic points. To address this, the back-end module optimizes and merges the static scan into the static map. This improvement in the static map facilitates better static/dynamic separation in the next iteration. This iterative process, known as the \emph{self-reinforcing mechanism}, gradually improves both mapping and tracking performances.
}
\label{fig:framework_overview}
\end{figure*}

In this paper, we present \textnameSMAT{} (\textbf{S}imultaneous and \textbf{S}elf-reinforced \textbf{M}apping \textbf{a}nd \textbf{T}racking), a novel solution that simultaneously tracks dynamic objects while constructing an online map in human-populated scenarios. As shown in \prettyref{fig:framework_overview}, \textnameSMAT{} comprises two tightly coupled modules, with the output of one module serving as the input of the other. The high-frequency front-end module detects and tracks potential dynamic points, while the more computational expensive back-end module reconstructs static spatial structures at a lower frequency. Initially, both modules provide low-quality results, with the front-end misclassifying many static points as dynamic and vice versa, and the back-end producing a local map with incorrect occupancy. \textnameSMAT{} leverages the interaction between the two modules to improve each other's results. The front-end uses \emph{tracking-by-detection} to remove incorrect static points from dynamic points and \emph{detection-by-tracking} to remove incorrect dynamic points from static points. The more accurate separation of dynamic and static points enables more precise tracking of multiple dynamic objects and more accurate separation of static points. The back-end then uses the refined set of static points to update voxel occupancy, providing a more accurate prior of the static spatial structure for the front-end in the next round. This \emph{self-reinforcing} mechanism combines the strengths of both modules, resulting in greater accuracy for dynamic object detection and online mapping.

We validated the proposed \textnameSMAT{} pipeline using diverse public datasets collected from vehicles and social robots. \textnameSMAT{} exhibited state-of-the-art performance on these datasets, as evidenced by mapping and tracking metrics. Additionally, we conducted extensive long-range experiments in real-world urban environments to further assess the effectiveness of our approach. In these experiments, a robot autonomously navigated through large human-populated scenes without relying on a pre-built global map but only utilizing a single LiDAR, a CPU-only onboard computer, and a consumer-level GPS receiver. The results of these experiments demonstrated the robustness, accuracy, scalability, and flexibility of \textnameSMAT{} in traversing and mapping various urban scenarios.


The rest of this paper is organized as follows. Section~\ref{sec:related} briefly reviews related works. Section~\ref{sec:perc_method} describes our self-reinforcing \textnameSMAT{} mechanism to solve the chicken-and-egg problem of tracking and mapping in dynamic scenarios. Section~\ref{sec:eval} evaluates \textnameSMAT{} in simulation scenarios and datasets. Section~\ref{sec:real_exp} evaluates \textnameSMAT{} by integrating it with the long-range navigation system and testing it in large-scale urban scenarios.

\section{Related works}
\label{sec:related}

In this section, we review the literature on spatial-temporal perception in dynamic scenarios. We categorize these studies into three types: spatial structure mapping, detection and tracking of dynamic objects, and simultaneous mapping and dynamic object detection. We also discuss their significance in long-range navigation in urban scenarios.

\subsection{Spatial structure mapping}
The most popular solution for spatial structure mapping is SLAM~\cite{zhang2014loam,shan2020lio,xu2021fast}, which integrates multiple sparse point clouds into a dense and complete point cloud map. But since it is designed for static environments with few obstacles, SLAM may perform poorly in dynamic environments because moving object point clouds can create ``ghost" tracks in the map, which can disrupt the spatial structure and compromises autonomous navigation.

To address this issue, one straightforward solution is to remove parts corresponding to dynamic objects from the raw sensor data and the remaining part can then be fed into SLAM for high-quality mapping. This removal process involves estimating the voxel occupancy probability in the 3D scene by tracking each emitted 3D LiDAR ray~\cite{hornung2013octomap}. A voxel is considered free if the ray passes through it and occupied if the ray stops at it. However, 3D LiDAR ray tracing is computationally expensive, and processing massive 3D data online is challenging, even with engineering optimization~\cite{schauer2018peopleremover,pagad2020robust}. Some other methods, such as visibility-based checking~\cite{pomerleau2014long,yoon2019mapless}, compute visibility difference to identify dynamic points. If a point is occluded in the line of sight of a previously observed point, it is labeled as dynamic. While visibility-based checking is more efficient than ray tracing, it simplifies the sensor model, resulting in lower mapping accuracy compared to ray tracing-based methods~\cite{lim2021erasor}. Another solution involves the removing-and-reverting mechanism~\cite{kim2020remove}, which iteratively retains static points from mistakenly removed points. But it relies on visibility for reverting and thus inherits the visibility-checking’s low performance. In addition, all these prior methods are offline and can mistakenly remove many static objects as dynamic objects.

In our previous work~\cite{fan2022dynamicfilter}, we combined the strengths of visibility-based checking and voxel occupancy checking for more efficient and accurate dynamic object removal than using each individual checking method alone. We utilized visibility-based checking as the high-frequency front-end and voxel occupancy checking as the low-frequency back-end.

\subsection{Detection and tracking of dynamic objects}
Various methods have been proposed for detecting and tracking dynamic objects. Recent advances in deep learning enable the detection of dynamic objects according to their appearance~\cite{lang2019pointpillars,cortinhal2020salsanext, milioto2019rangenet++}. However, these methods only identify movable objects rather than moving objects, and their ability to generalize to unknown object categories beyond the training datasets is limited. Another approach is to use sequential range images to identify dynamic points based on inconsistency~\cite{chen2021lidar-mos,mersch2022receding}. Some learning-based approaches estimate the scene flow from sequential point clouds~\cite{liu2019flownet3d,wu2020pointpwc,huang2022dynamic}. But they rely on manually labeled datasets and may not effectively generalize to unseen scenarios in real-world applications.

Some model-free approaches use motion clues to detect and track moving objects. For example, \cite{kaestner2012generative} utilizes a generative Bayesian method, but it is limited to stationary sensors and does not work with moving sensors. Another method~\cite{dewan2016motion} combines RANSAC and a Bayesian method for segmenting and tracking moving objects. It can also estimate scene flow from sequential frames~\cite{dewan2016rigid}. While these approaches are effective, they are suboptimal for tracking moving objects with low velocities, like crowds of pedestrians.

The performance of moving object detection can be significantly improved by combining it with spatial structure mapping. For example, inconsistencies between the existing static map and incoming scans can be utilized to detect dynamic objects~\cite{azim2012detection}. Occupancy map can also propose coarse candidates of moving objects, enabling accurate estimation of dynamic points through learning-based approaches~\cite{ushani2017learning}. Offline spatial structure mapping is employed in~\cite{pfreundschuh2021dynamic, chen2022automatic} to label dynamic objects, serving as the ground truth for learning-based online dynamic object detection. However, none of these methods acknowledge the potential reciprocal benefits between mapping quality and dynamic object tracking.

\subsection{Simultaneous mapping and dynamic object detection and tracking}

Some works have recognized the importance of simultaneously estimating the static map and the motion of dynamic objects. The earliest effort, SLAMMOT~\cite{wang2007simultaneous}, utilizes a joint probabilistic model to estimate the motion of moving objects and the robot pose. However, its efficiency and robustness have not been evaluated in 3D scenarios. In~\cite{moosmann2013joint}, point clouds are initially segmented into dynamic and static parts, which are then tracked using joint estimation. The decoupling between segmentation and tracking prevents this method from leveraging the interplay between spatial structure mapping and moving object detection. Another method~\cite{tanzmeister2014grid} detects dynamic objects of grid cells using online estimated occupancy maps. Similarly,~\cite{wang2015model} associates each incoming scan with a local static map and dynamic objects. Unfortunately, these methods are only effective for 2D LiDAR data, not 3D LiDAR data.

\subsection{Long-range navigation in urban scenarios}

Long-range navigation in unstructured urban environments has garnered considerable attention in the research community, particularly through the development of autonomous driving systems during the DARPA Urban Challenges~\cite{urmson2008autonomous, montemerlo2008junior}. Unlike autonomous vehicles that primarily traverse well-defined streets and lanes, these robots must navigate through less-structured areas with a significant human presence. In~\cite{kummerle2015autonomous}, a navigation system is proposed that enables a mobile robot to autonomously travel \SI{7.4}{km} through city centers, successfully navigating pedestrian zones and densely populated areas. ~\cite{francis2020long} combines reinforcement learning and sampling-based methods to achieve efficient global planning for long-range navigation. Additionally, \cite{wang2022motion} discusses the successful deployment and operation of last-mile delivery systems in various urban environments. Unlike these approaches that rely on pre-built global maps, our method enables the robot to navigate through urban environments without the need for a precomputed high-quality global map.

\section{Simultaneous and self-reinforced mapping and tracking}
\label{sec:perc_method}

\begin{figure*}
\centering
\includegraphics[width=1.0\linewidth]{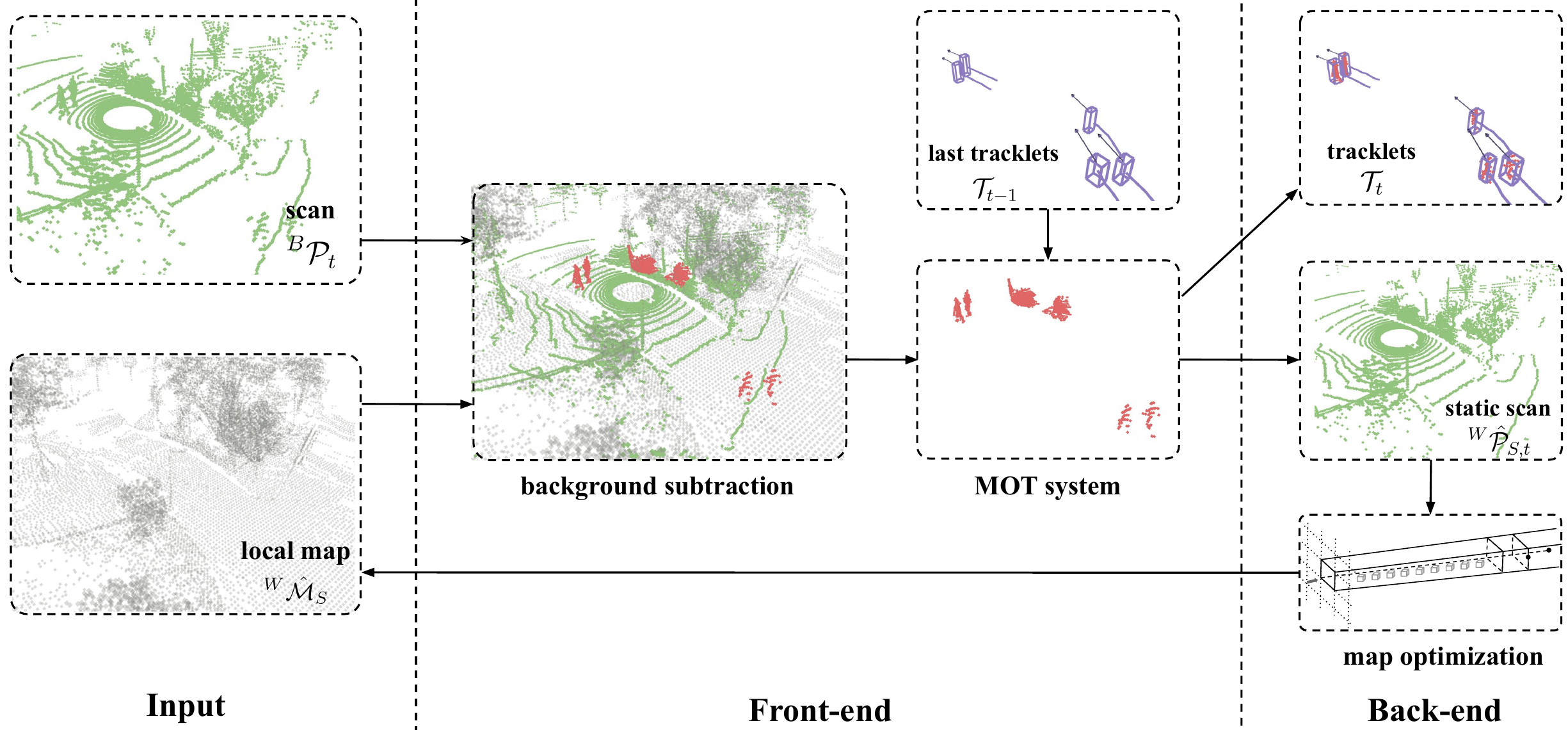}
\caption{
Illustration about how an incoming LiDAR scan is processed through the front-end and back-end modules to accomplish moving object tracking and environmental mapping.}
\label{fig:framework_pipeline}
\end{figure*}

This section presents the \textnameSMAT{} framework for online mapping and tracking in a dynamic environment. The framework consists of two modules: a front-end module that runs at a high frequency for detecting and tracking moving objects, and a back-end module that runs at a relatively lower frequency for creating a static spatial map. The performance of both modules is gradually improved through a self-reinforcing mechanism that encourages the interplay between them. We will first provide an overview of the problem formulation and framework pipeline, followed by a detailed discussion of the front-end, back-end, and the self-reinforcing mechanism.

\subsection{Framework overview}
\label{sec:framework_overview}

\textnameSMAT{} has two main objectives: detecting and tracking moving objects in consecutive 3D LiDAR scans and generating spatial structure maps. To aid comprehension, we define several concepts as follows:

\begin{itemize}
    \item LiDAR scan $^B\!\mathcal{P}_t$: a point cloud captured at time step $t$ in the local sensor frame $B$. It can be divided into two parts: the static scan and the dynamic scan, corresponding to point clouds of static and moving objects, respectively. 
    \item Spatial structure map $^W\!\mathcal{M}$ under the global coordinate system $W$: it is obtained by accumulating scans within the global frame.
    \item  Tracklet ${\mathcal{T}_t}$: a short track segment that captures the moving objects up to time $t$. Each tracklet is uniquely identified and contains information about the state of the corresponding object at each tracked moment, including its speed, position, and bounding box dimensions.
\end{itemize}

\textnameSMAT{} consists of a front-end and a back-end, as shown in \prettyref{fig:framework_overview}. The pipeline begins with the front-end, which uses the a priori static map $^W\!\hat{\mathcal{M}}_S$ and the current scan to detect and track moving objects $\mathcal{T}_t$. It then generates a static scan $^W\!\hat{\mathcal{P}}_{S, t}$ by filtering out points from dynamic objects. The back-end module operates at a lower frequency compared to the front-end and is executed when the robot moves a certain distance or for a certain period. The back-end integrates multiple static scans $\{^W\!\hat{\mathcal{P}}_S\}$ from the front-end and optimizes the estimated static map $^W\!\hat{\mathcal{M}}_S$ to serve as the prior map for the front-end in the next round. The resulting iteration cycle is called the \emph{self-reinforcing} cycle: the front-end uses the refined prior static map generated by the back-end to better identify dynamic points and improve dynamic object detection, while the back-end uses the improved separation of static and dynamic points for more accurate map updates. In \prettyref{fig:framework_pipeline}, we also provide a concrete example explaining how the sensory data flows through the entire pipeline. Next, we are going to introduce \textnameSMAT{}'s each module in more detail.

\begin{figure*}
\centering
\includegraphics[width=1.0\linewidth]{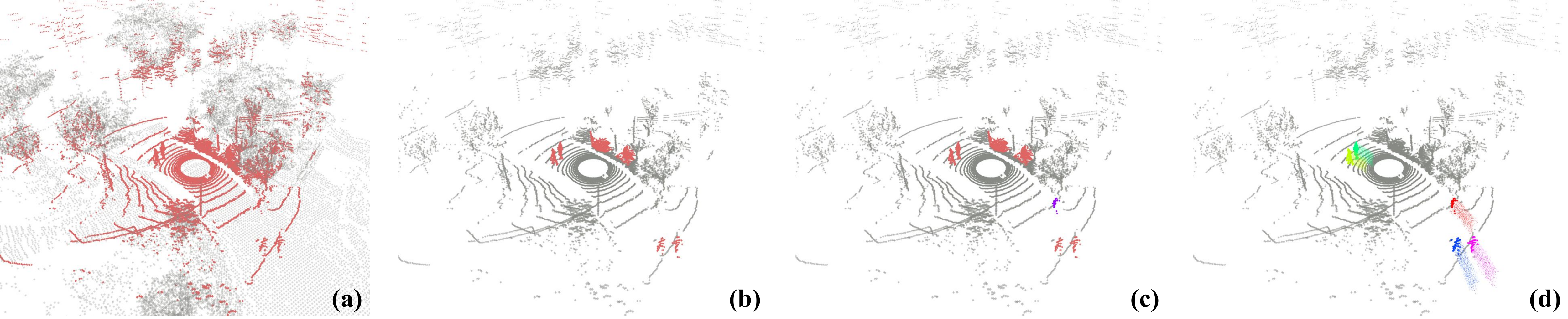}
\caption{
Front-end results. Given the current scan (red points) and the prior static map (grey points) from the back-end as shown in (a), the background subtraction generates the scan’s initial separation of static scan (grey points) and dynamic scan (red points) as shown in (b). Then, \textit{detection by tracking} identifies dynamic points misclassified as static, which are marked as purple points in (c). Finally, in (d) \textit{tracking by detection} identifies static points misclassified as dynamic and only the successfully associated and tracked points are considered dynamic. We use different colors to signify different moving instances, with the historical points of each tracklet also displayed.}
\label{fig:front_end_demo}
\end{figure*}

\subsection{Detection and tracking in the front-end}
\label{sec:frontend}

The front-end module has two parts: background subtraction and multiple object tracking (MOT), and is responsible for detecting and tracking dynamic objects in each scan. At time step $t$, the front-end receives a combined input, which includes 1) the newly acquired scan $^B\!\mathcal{P}_t$ and sensor's current pose in the workspace, 2) the prior static map $^W\!\hat{\mathcal{M}}_S$ estimated by the back-end, and 3) a set of dynamic object tracklets ${\mathcal{T}_{t-1}}$ computed by MOT in the last time step.

\noindent\textbf{Static background subtraction.}
In this step, we extract potential dynamic points in the current scan $^B\!\mathcal{P}_t$ by referencing the static map $^W\!\hat{\mathcal{M}}_S$ provided by the back-end, as shown in \prettyref{fig:front_end_demo}. The extraction results may contain two types of errors: \emph{type I errors}, where static points are incorrectly identified as dynamic, and \emph{type II errors}, where dynamic points are incorrectly identified as static.

We first transform the local scan $^B\!\mathcal{P}_t$ into the global frame $W$ and get $^W\!\mathcal{P}_t$, where the transform can be computed using LiDAR-based SLAM approaches, such as \cite{shan2020lio, xu2021fast}. Then, we crop the global workspace map $^W\!\hat{\mathcal{M}}_S$ provided by the back-end to obtain a local static map $^W\!\hat{\mathcal{M}}_{S}$ that is within a radius of $R_{\text{bound}}$ around the sensor, where $R_{\text{bound}}$ is determined by the maximum range of the sensor.

The local static map $^W\!\hat{\mathcal{M}}_{S}$ is voxelized into multiple occupied voxels, which are used to distinguish static and dynamic points in the current scan $^W\!\mathcal{P}_t$. If the prior map were perfect, points inside an occupied voxel would be considered static, while points outside the voxel would be considered dynamic. And in cases where no prior map is available, such as when initializing the pipeline, all points in $^W\!\mathcal{P}_t$ are treated as dynamic points. This division results in two sets: the initial static scan $^W\!\hat{\mathcal{P}}_{S, t}^{\text{init}}$ and the initial dynamic scan $^W\!\hat{\mathcal{P}}_{D, t}^{\text{init}}$, as shown in \prettyref{fig:front_end_demo}b. However, since the actual prior map is not perfect, this initial division may contain both type I and type II errors. Type I errors can occur when a static point is located in a region of the prior map with sparse voxels or in a new area not covered by the prior map. Type II errors can occur when sensor noise or state estimation drift is significant, causing some occupied voxels in the prior map to be incorrectly identified as free.

We next use multi-object tracking (MOT) to reduce type I and II errors in the initial scan separation. The MOT system employs two components: \textit{tracking by detection} and \textit{detection by tracking}. In \textit{tracking by detection}, we verify the trackability of points in $^W\!\hat{\mathcal{P}}_{D, t}^{\text{init}}$ to reduce the type I error, while in \textit{detection by tracking}, we use historical tracking information to help find missed dynamic points within the initial static scan $^W\!\hat{\mathcal{P}}_{S, t}^{\text{init}}$ to reduce the type II error.

\noindent\textbf{Tracking-by-dectection.}
In this part, we utilize a clustering algorithm~\cite{bogoslavskyi2016fast} to segment the initial dynamic scan $^W\!\hat{\mathcal{P}}_{D, t}^{\text{init}}$ into multiple object hypotheses, where each hypothesis is a bounding box enclosing a cluster of points.

We then associate these object hypotheses with the most recent bounding boxes of tracked objects in ${\mathcal{T}_{t-1}}$ using a greedy nearest neighbors matching approach with L2 distances as the association metric. If an object hypothesis is matched with an existing tracked object, we update its state using an extended Kalman filter with a constant-velocity motion model and add it to the object's tracklet. If an object hypothesis is not matched with any existing object, we create a new tracklet for it. Tracklets in ${\mathcal{T}_{t-1}}$ that are not matched with any object hypotheses for a sufficiently long time will be removed.


However, not all associated hypotheses, but only those that exhibit stable association and tracking with consistent velocities, shall be considered as actual dynamic objects. To determine if a hypothesis has been stably tracked during the last $t_{\text{val}}$ period, we use the following heuristic criteria:
\begin{itemize}
    \item Success rate of data association $\rho > \rho_{\min}$, where $\rho = \frac{\# \text{frames that the object being successfully associated}}{\# \text{total frames from first tracked}}$;
    \item Average speed $v > v_{\min}$, where $v = \frac{d}{t_{\text{val}} \ * \ \rho}$, and $d$ is the observed displacement distance of the object;
    \item Volume change $\Delta \mathbb{V} < \Delta \mathbb{V}_{\min}$, where $\Delta \mathbb{V} = \mathbb{V}_{\max} - \mathbb{V}_{\min}$, and $\mathbb{V}_{\max}$ and $\mathbb{V}_{\min}$ are the maximum and minimum volumes of the object in this period.
\end{itemize}
If a hypothesis fails to meet these criteria, the ``dynamic" points within it are reclassified as static points. These points are removed from the dynamic scan $^W\!\hat{\mathcal{P}}_{D,t}^{\text{init}}$ and added to the static scan $^W\!\hat{\mathcal{P}}_{S,t}^{\text{init}}$.


\noindent\textbf{Detection-by-tracking.}
If an object in ${\mathcal{T}_{t-1}}$ has been stably tracked in previous scans but cannot be associated with any object hypotheses in the current scan, we employ \textit{detection-by-tracking} to determine whether it should be retained. This involves generating an object hypothesis through motion prediction using the extended Kalman filter and a consistent velocity model. First, we predict the position of the object in the current scan based on its velocity from the last tracklet, assuming that the object's size and orientation remain constant over a short period. If the generated hypothesis contains a sufficient number of points, it is considered a valid moving object, and all points within the bounding box of this hypothesis are reclassified as dynamic points. These points are then removed from the static scan $^W\!\hat{\mathcal{P}}_{S, t}^{\text{init}}$ and added back to the dynamic scan $^W\!\hat{\mathcal{P}}_{D, t}^{\text{init}}$. The result is demonstrated in \prettyref{fig:front_end_demo}c.

After \textit{tracking-by-detection} and \textit{detection-by-tracking}, the initial division of $^W\!\hat{\mathcal{P}}_{S, t}^{\text{init}}$ and $^W\!\hat{\mathcal{D}}_{S, t}^{\text{init}}$ is updated to the more accurate $^W\!\hat{\mathcal{P}}_{S, t}$ and $^W\!\hat{\mathcal{P}}_{D, t}$ with reduced type I and II errors. This process also generates improved motion estimation for the moving object, as shown in \prettyref{fig:front_end_demo}d.

After finishing the front-end, the current scan efficiently and accurately extracts most fast-moving dynamic objects. However, the static scan $^W\!\hat{\mathcal{P}}_{S, t}$ may still include dynamic points with subtle movements, leaving significant type II errors. Furthermore, due to the sparsity of individual scans, the robot must accumulate multiple scans to create a comprehensive static map. These issues will be addressed in the subsequent back-end.


\subsection{Spatial mapping in the back-end}

The back-end operates at a lower frequency than the front-end and is triggered after the robot has moved a certain distance or for a certain amount of time. All the static scans generated by the front-end, along with their corresponding poses, are stored in a buffer. The back-end combines these static scans, which still have type II errors with missing dynamic points, to create a high-quality map with lower type II errors. 

Initially, the back-end searches the buffer for static scans that have poses close to the current scan's pose, typically within a radius $R_{\text{localmap}}$. The searching results are transformed to the sensor's local frame $B$, forming a set of nearby static scans $\{^B\!\hat{\mathcal{P}}_{S, k}\}$ in the sensor's vicinity. For points in each $k$-th nearby static scan $^B\!\hat{\mathcal{P}}_{S, k}$, they can be viewed as the endpoints of a set of LiDAR rays originating from the center of the sensor when perceiving that scan. 

Next, we voxelize the space encompassing the points of these static scans. The occupancy probability of each voxel can be computed as $\frac{n_{\text{occ}}}{n_{\text{occ}} + n_{\text{free}}}$. Here, $n_{\text{occ}}$ is the number of static scans with at least one LiDAR ray terminating in this voxel, which can be efficiently determined. On the other hand, computing $n_{\text{free}}$ directly requires ray tracing, which is computationally expensive for real-time mapping. Therefore, we employ \textit{visibility checking} to efficiently estimate $\hat{n}_{\text{free}}$.

\noindent\textbf{Visibility checking.} 
To approximate ray tracing for the $k$-th nearby static scan, we first transform the voxels back into that scan's local frame $B_k$. We then project the points in $^{B}\!\hat{\mathcal{P}}_{S, k}$ onto the sensor's imaging plane using the LiDAR's field of view (FOV) as the projection FOV. This projection yields a range image $I_k = [I_{k, r,c}] \in \mathbf{R}^{m\times n}$, where $m$ and $n$ represent the image dimensions determined by the resolution per pixel and FOV ranges. The resolution per pixel is typically set to match the LiDAR's vertical and horizontal resolutions. $I_{k, r,c}$ denotes the pixel value at coordinate $(r, c)$ in the range image for the $k$-th scan and is computed as:
\begin{equation*}
I_{k, r, c} = \min_{\mathbf p \in \hat{\mathcal{P}}_{S, k}^{r,c}} \text{range}(\mathbf p),
\end{equation*}
where $\text{range}(\cdot)$ is the range or depth of a point $\mathbf{p}$ in the sensor's local frame, and $\hat{\mathcal{P}}_{S, k}^{r,c} =~^B\!\hat{\mathcal{P}}_{S, k} \cap \text{Cone}(r, c)$ is a subset points of the static scan inside $\text{Cone}(r, c)$, which is the view space cone with the pixel $(r, c)$ as its apex. 
Next, we address the centroids of voxels in the same way. For each voxel whose centroid is projected onto the pixel $(r, c)$, we compare its range $d$ with the value $I_{k, r, c}$ from the range image $I_k$. If $d < I_{k, r, c}$, it indicates that a LiDAR ray passes through the voxel, resulting in an increment of $\hat{n}_{\text{free}}$ by 1. Note that visibility checking can effectively remove dynamic points that are not detected in the front-end, leading to a lower type II error.


\begin{figure}
\centering
\includegraphics[width=0.9\linewidth]{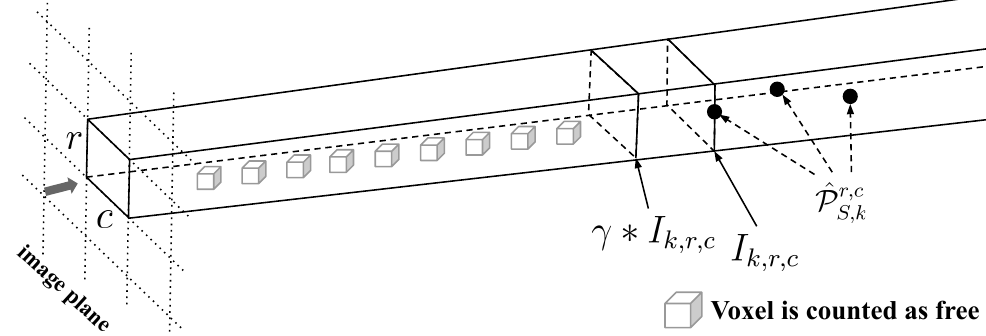}
\caption{Illustration of the \textit{visibility checking} in the back-end. The voxels inside the visibility cone are considered as passed through by LiDAR rays.}
\label{fig:visibility_check}
\end{figure}

However, using $I_{k, r, c}$ directly for \textit{visibility checking} can result in inaccuracies caused by occlusion and significant incident angle problems \cite{lim2021erasor}. To address this issue, we incorporate the concept of a safe sphere for range images, as proposed in \cite{schauer2018peopleremover}, which establishes a more conservative visibility boundary. Specifically, we modify the criterion for determining if a voxel is passed through by ray-tracing to $d < \gamma \cdot I_{k, r, c}$, where $\gamma \in [0.0, 1.0)$. Once we calculate the occupancy probability, we label voxels with a probability exceeding the threshold as occupied. This method enables us to create a voxelized static submap for the $k$-th nearby static scan.

The \textit{visibility checking} process is illustrated in \prettyref{fig:visibility_check} and its pseudocode is presented in \prettyref{alg:visibility_check}.

\begin{algorithm}
\caption{Visibility checking}
\label{alg:visibility_check}
\begin{algorithmic}[1]
\State \textbf{Input}: nearby static scans $\{ ^B\! \hat{\mathcal{P}}_{S, k}\}$, their corresponding poses $\{ ^W_B \mathbf T_k\}$, and occupancy threshold $p_{\text{occ}}$. 
\State // \textit{initialize submap}
\State $^W\!\mathcal{M} = \emptyset$
\For {$\text{scan } k = 1, 2,..., N$}
  \State // \textit{project scan to current image plane}
  \State $I_{k} \leftarrow \{d, r, c\} = \text{project}(^B\! \hat{\mathcal{P}}_{S, k})$
  \State // \textit{accumulate scan into the submap}
  \State $^W\!\mathcal{M} \pluseq ^W_B\mathbf T_k * ^B\! \hat{\mathcal{P}}_{S, k}$
\EndFor
\State // \textit{initialize static submap}
\State $^W\!\hat{\mathcal{M}}_S$ = $\emptyset$
\State $\{ ^W\! \mathbf{v}_i \}$ = voxelize($^W\!\mathcal{M}$)
\For {$\text{voxel } i = 1, 2, \cdots$}
    \State // \textit{initialize occupancy model}
    \State $n_{\text{occ}}$ = countOccupied($^W\!\mathcal{M}$, $^W\! \mathbf{v}_i$)
    \State $n_{\text{free}}$ = $0$
    \For {$\text{scan } k = 1, 2, \cdots$}
        \State // \textit{project voxel to local image plane}
        \State $^B\!\mathbf{v}$ = $^W_{B_k}\mathbf T_k^{-1} * ^W\! \mathbf{v}_i$
        \State $d, r, c$ = project($^B\!\mathbf{v}$)
        \If {$d < \gamma * I_{k} [r, c]$}
            \State $n_{\text{free}} \pluseq 1$
        \EndIf
    \EndFor
        \If {$\frac{n_{\text{occ}}}{n_{\text{occ}} + \hat{n}_{\text{free}}} > p_{\text{occ}}$}
        \State $^W\!\hat{\mathcal{M}}_S \pluseq ^W\! \mathbf{v}_i$
    \EndIf
\EndFor
\State \textbf{Output}: static submap $^W\!\hat{\mathcal{M}}_S$
\end{algorithmic}
\end{algorithm}

\noindent\textbf{Map merging.} 
We perform the \textit{visibility checking} for nearby static scans, generating a set of local static submaps that are limited to specific regions. The back-end then transforms these submaps into the global frame using their corresponding poses and merges them with the global static map. As there may be overlaps between multiple local submaps, different submaps may provide different occupancy estimations for the same voxel. To address this potential inconsistency, we utilize the fact that locations near the center of each submap are more accurately perceived. This is because LiDAR scans become sparser with increasing distances, and therefore, the \textit{visibility checking} of a static scan provides a more accurate occupancy estimate for voxels closer to the sensor's instantaneous location. Consequently, we designate a voxel as occupied if the submap where it is closest to the sensor predicts it as occupied. Otherwise, it will be predicted free. The centroids of these occupied voxels form our final down-sampled static map ${^W\!\hat{\mathcal{M}}_S}$. This map can serve as prior knowledge for the front-end in the next iteration.

\begin{figure}
\centering
\includegraphics[width=0.9\linewidth]{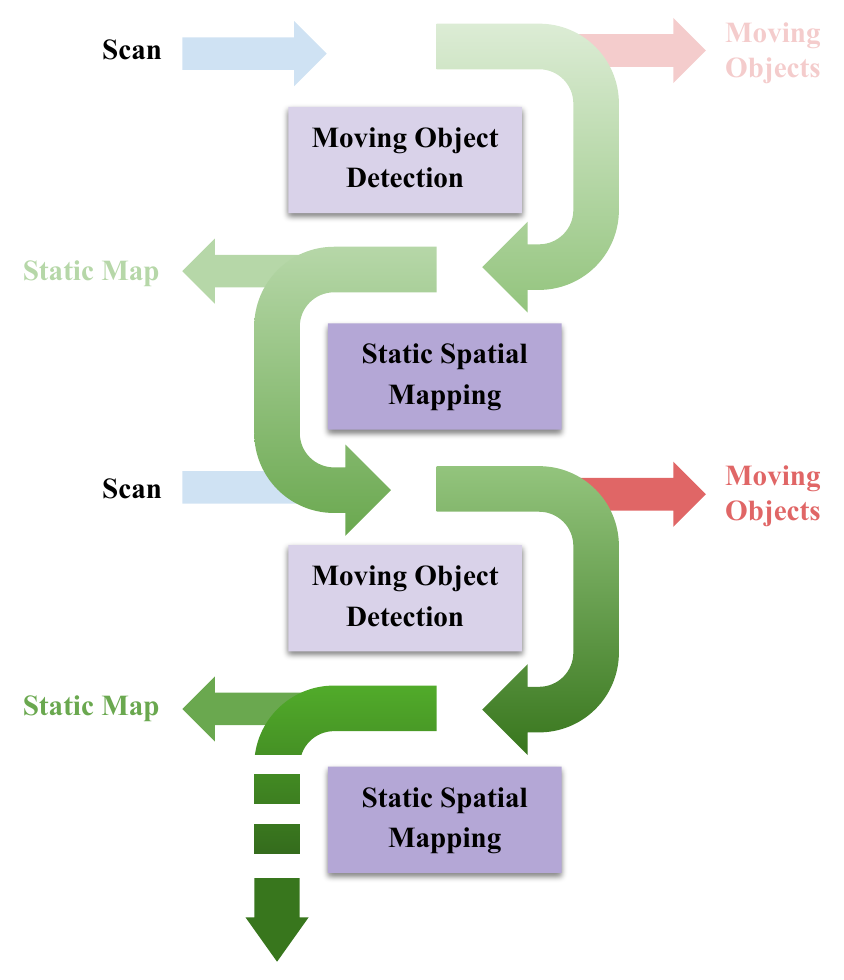}
\caption{\textnameSMAT{}'s self-reinforcing mechanism, where the shades of color indicate the accuracy of the scan or tracking. \textnameSMAT{}'s iterative optimization between moving object tracking and static spatial mapping, the performance for both tasks gradually improves.}
\label{fig:self-reinforcing}
\end{figure}

\subsection{Self-reinforcing mechanism}



After one round of the \textnameSMAT{} pipeline, the back-end generates a global static map. This map covers a larger region and provides a more accurate occupancy estimate with smaller type I error (i.e., some static voxel regions are missing) and type II error (i.e., some dynamic voxels are not filtered out) compared to the global static map from the previous round. It serves as a better prior for the next round's front-end. This continuous interplay between the front-end and back-end creates a self-reinforcing mechanism, forming a positive feedback loop. In this loop, the front-end operates at a high frequency to detect and track potential dynamic points, while the more computationally expensive back-end reconstructs static spatial structures at a lower frequency.


The benefit of the back-end and front-end's interplay can be explained using Bayesian estimation. The front-end begins with the background subtraction, which separates static and moving objects. It estimates the posterior $\mathbb{P}(\mathbf p = 0)$ based on the ideal likelihood knowledge that $\mathbb{P}(\mathbf p = 0 \mid \mathbf v = 1) = 0$ and the prior static map $\mathbb{P}(\mathbf v = 1)$ provided by the back-end. Here, $\mathbf p$ represents a point located in voxel $\mathbf v$, and both have two states: a point can be dynamic (0) or static (1), while a voxel can be free (0) or occupied (1).
However, due to noise, in practice $\mathbb{P}(\mathbf p = 0 \mid \mathbf v = 1) \neq 0$. To better describe the actual situation, we use Bayesian theorem $\mathbb{P}(\mathbf p = 0 \mid \mathbf v = 1) \propto \mathbb{P}(\mathbf v = 1 \mid \mathbf p = 0)\mathbb{P}(\mathbf p = 0)$ to compute a reasonable likelihood. The values for $\mathbb{P}(\mathbf p = 0)$ and $\mathbb{P}(\mathbf v = 1 \mid \mathbf p = 0)$ are adjusted using \emph{tracking-by-detection} and \emph{detection-by-tracking}, respectively.

Dynamic point removal in the front-end enhances the accuracy of voxel state observations. The back-end accumulates and utilizes multiple static scans from the font-end as denoised measurements $z$ of the ground-truth voxels. This allows the back-end to compute a conservative occupancy grid posterior $\mathbb{P}(\mathbf v = 1 \mid z)$. Consequently, the back-end can provide a robust prior to the front-end, even in highly dynamic scenes.

Our experiments will demonstrate that the interplay between the front-end and back-end can yield superior performance compared to the individual performance of either component.

\section{Evaluation}
\label{sec:eval}

In this section, we first quantitatively and qualitatively evaluate \textnameSMAT{} in terms of the mapping quality for spatial structure and the performance of detecting and tracking moving objects. We then conduct a detailed ablation study in dynamic simulation environments to further investigate the effectiveness of \textnameSMAT{}.

\subsection{Mapping quality in urban scenarios}

To evaluate the quality of the recovered static map after removing dynamic points, we utilize the preservation rate (PR) and rejection rate (RR) metrics proposed by~\cite{lim2021erasor}. These voxel-wise metrics are evaluated at a resolution of \SI{0.2}{\meter}. Specifically, the metrics are defined as:
\begin{itemize}
    \item PR: $\dfrac{\# \text{preserved static points by static map}}{\# \text{total static points on raw map}}$
    \item RR: $1 - \dfrac{\# \text{preserved dynamic points by static map}}{\# \text{total dynamic points on raw map}}$
\end{itemize}
In addition, we compute the $\text{F}_1$ score, which is the harmonic mean of precision and recall. Note that large PR and RR correspond to small type I and II errors in the separation of static and dynamic points, respectively. 

We compare \textnameSMAT{}'s performance to various representative solutions, including: i) OctoMap~\cite{hornung2013octomap}, a typical approach to occupancy maps; ii) Removert~\cite{kim2020remove}, a state-of-the-art visibility-based method; iii) ERASOR~\cite{lim2021erasor}, the current state-of-the-art in SemanticKITTI; iv) 4DMOS~\cite{mersch2022receding}, a state-of-the-art learning-based method for moving object segmentation; and v) DynamicFilter~\cite{fan2022dynamicfilter}, our previous work that only conducts dynamic object removal. As 4DMOS only distinguishes between static and dynamic points within individual scans, we aggregate its static scans to obtain the final map.

Among all methods, \textnameSMAT{} is the most suitable solution for real-time navigation in large-scale unknown dynamic scenarios. It has a low computational cost and can be deployed on onboard computers with just a single CPU, making it highly practical for real-world deployment. In contrast, other methods have different limitations. Removert and ERASOR are offline methods that require a pre-built map, limiting their applicability to explored scenarios. OctoMap has high computational costs that increase rapidly with scene scale, making it challenging to deploy on robots with limited onboard computational resources. 4DMOS relies on GPUs, making it unsuitable for CPU-only onboard computers. Additionally, it utilizes multiple scans before and after the current scan to identify static and dynamic points in the current scan, resulting in computational delay.


\noindent{\textbf{Comparison on SemanticKITTI dataset.}}
We utilize the SemanticKITTI dataset~\cite{geiger2012KITTI,behley2019semantickitti} as our benchmark. It was collected in an urban environment using a vehicle and contains manually labeled moving objects. We follow the experimental setup of~\cite{lim2021erasor} and perform quantitative experiments on specific segments containing multiple dynamic objects.

As summarized in \prettyref{tab:kitti}, \textnameSMAT{} achieves the highest $\text{F}_1$ score across almost all sequences, indicating its superior performance compared to others. The only exception is sequence 02, where 4DMOS outperforms \textnameSMAT{}. However, this can be attributed to 4DMOS being trained specifically on sequence 02, which may result in overfitting, particularly in terms of single-frame filtering effectiveness. But 4DMOS does not optimize multi-frame mapping, leading to the incorrect inclusion of dynamic obstacles in the map that cannot be removed, resulting in a decrease in RR. \textnameSMAT{} demonstrates superior performance in terms of PR and RR, highlighting its effectiveness in simultaneously filtering dynamic points while preserving static features. This further emphasizes the benefits of self-reinforcing mechanisms.

\begin{figure*}
\centering
\begin{minipage}[t]{0.30\linewidth}
\centering
\includegraphics[width=1\linewidth]{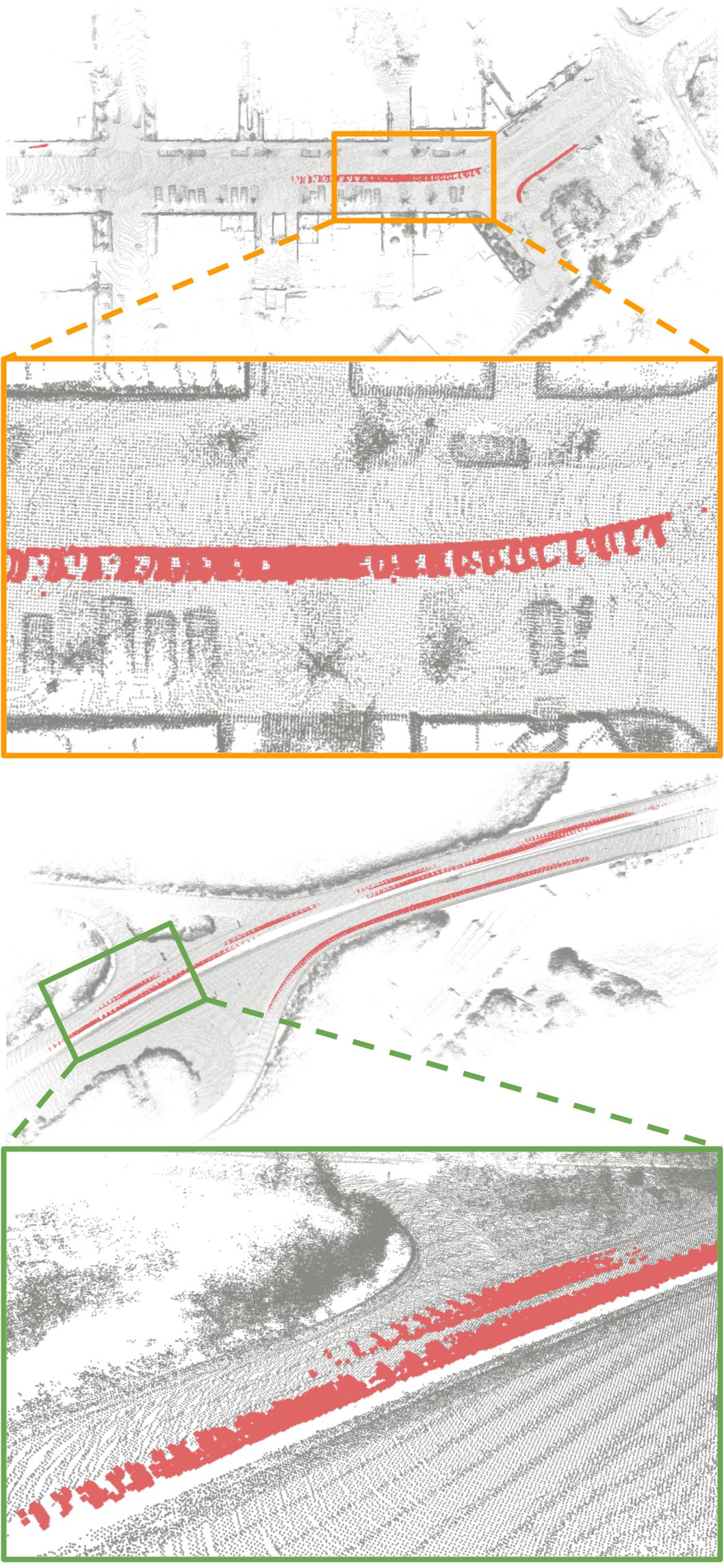}
\subcaption{Original Map\centering}
\end{minipage}
\begin{minipage}[t]{0.30\linewidth}
\centering
\includegraphics[width=1\linewidth]{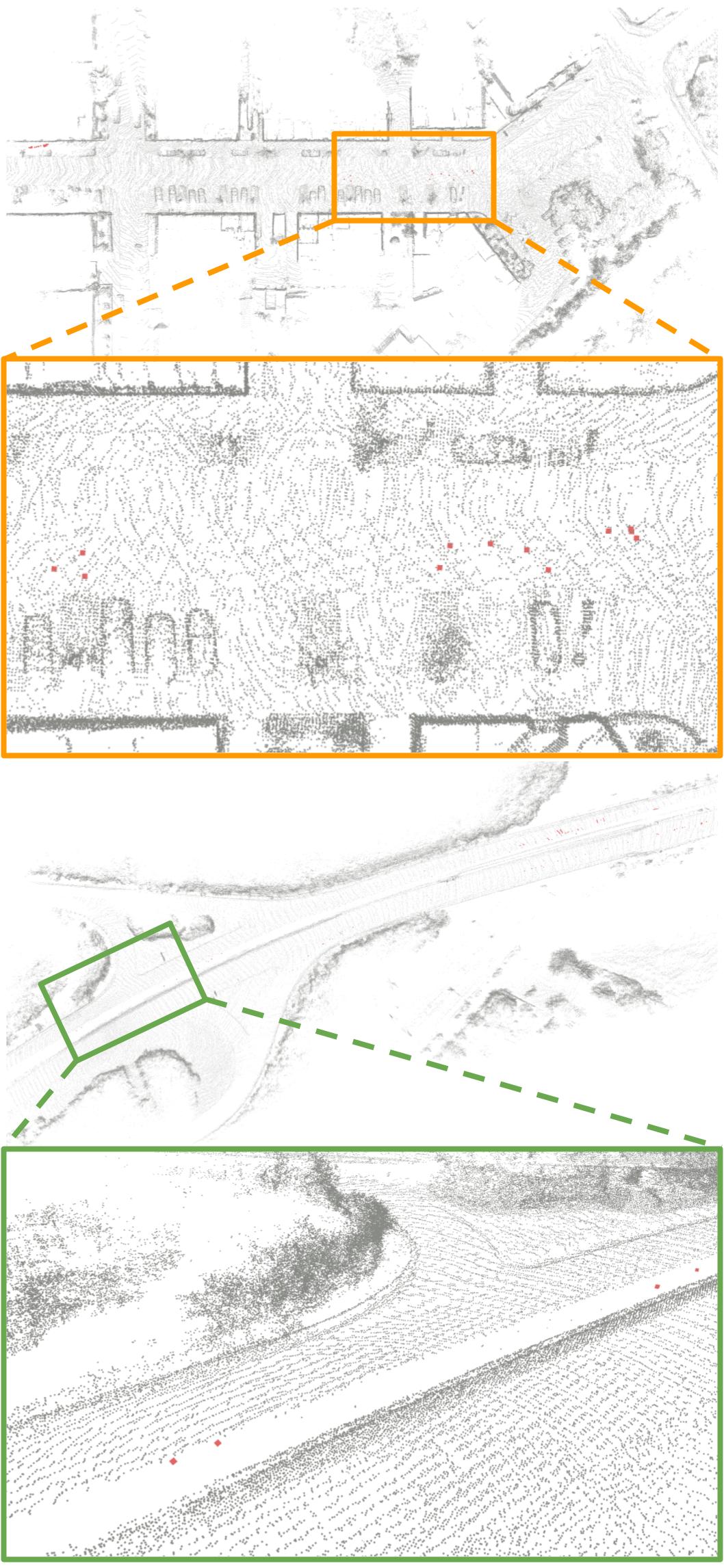}
\subcaption{OctoMap~\cite{hornung2013octomap}\centering}
\label{fig:kitti_comparison_octomap}
\end{minipage}
\begin{minipage}[t]{0.30\linewidth}
\centering
\includegraphics[width=1\linewidth]{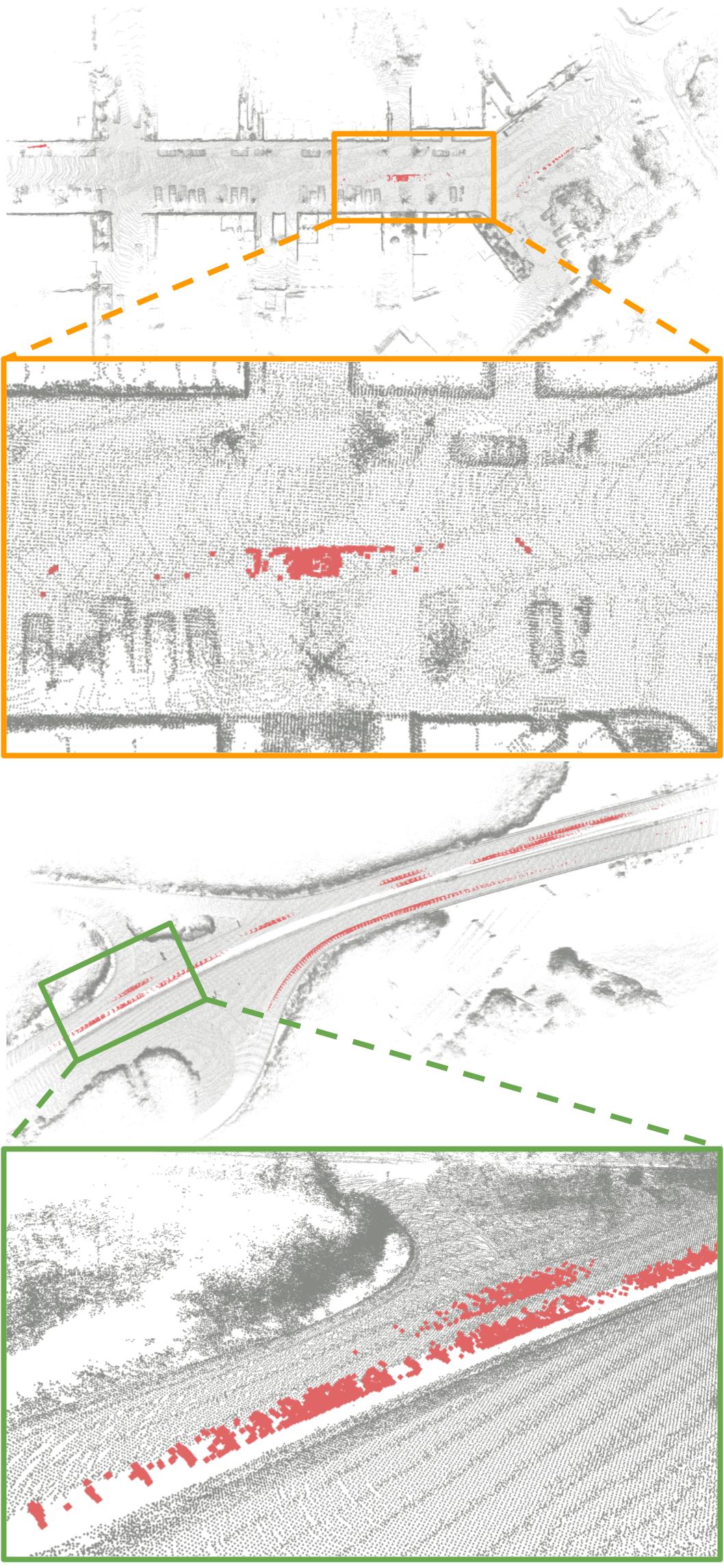}
\subcaption{Removert~\cite{kim2020remove}\centering}
\label{fig:kitti_comparison_removert}
\end{minipage}
\begin{minipage}[t]{0.30\linewidth}
\centering
\includegraphics[width=1\linewidth]{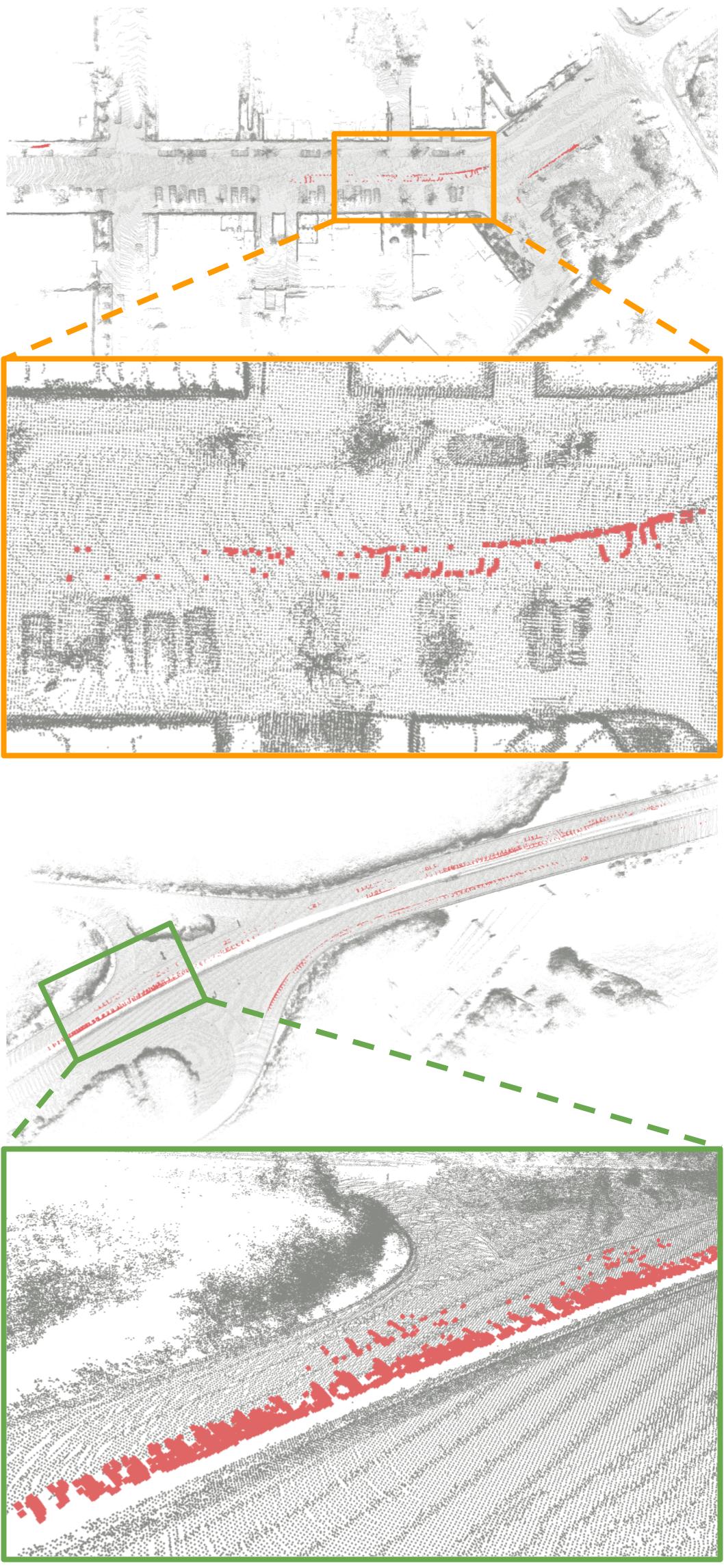}
\subcaption{4DMOS\centering~\cite{mersch2022receding}}
\label{fig:kitti_comparison_4dmos}
\end{minipage}
\begin{minipage}[t]{0.30\linewidth}
\centering
\includegraphics[width=1\linewidth]{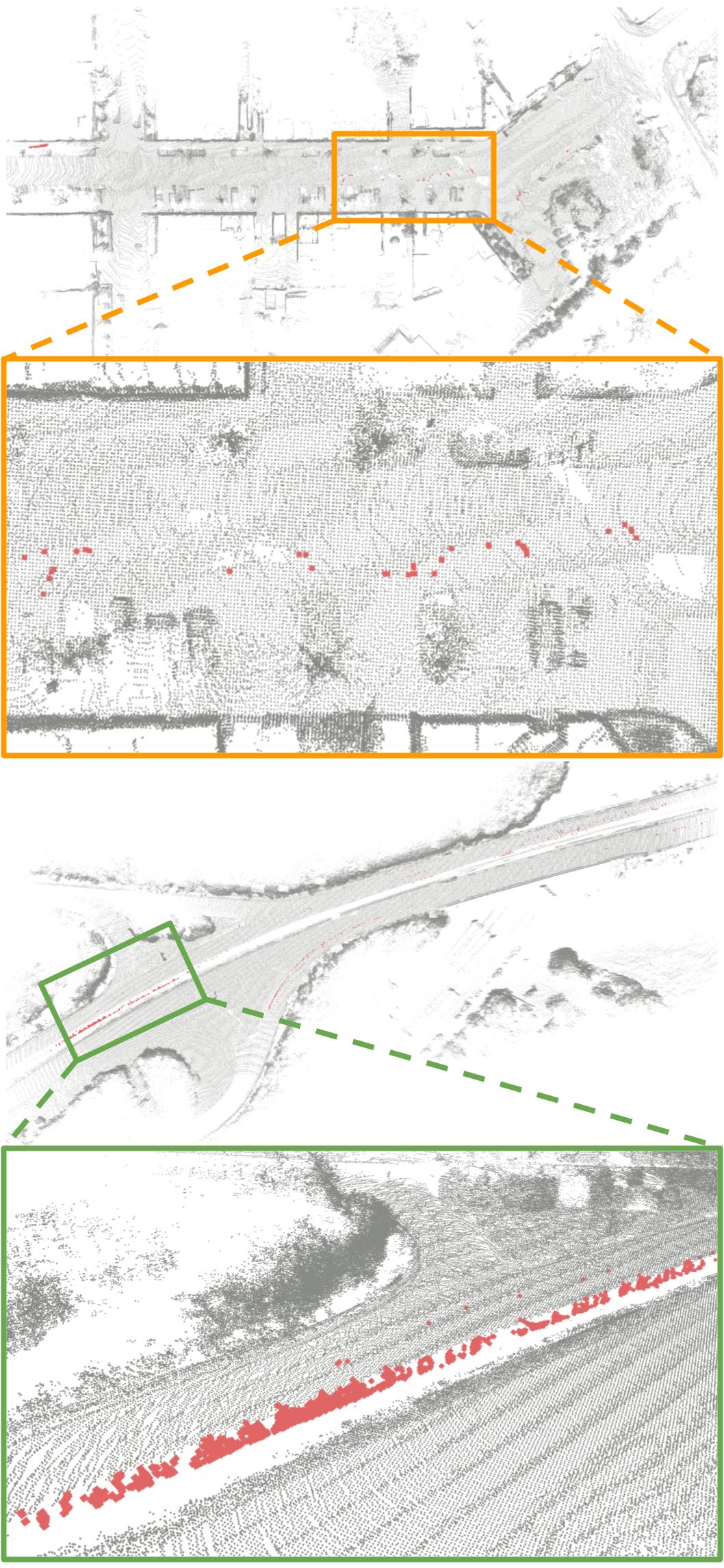}
\subcaption{ERASOR~\cite{lim2021erasor}\centering}
\label{fig:kitti_comparison_erasor}
\end{minipage}
\begin{minipage}[t]{0.30\linewidth}
\centering
\includegraphics[width=1\linewidth]{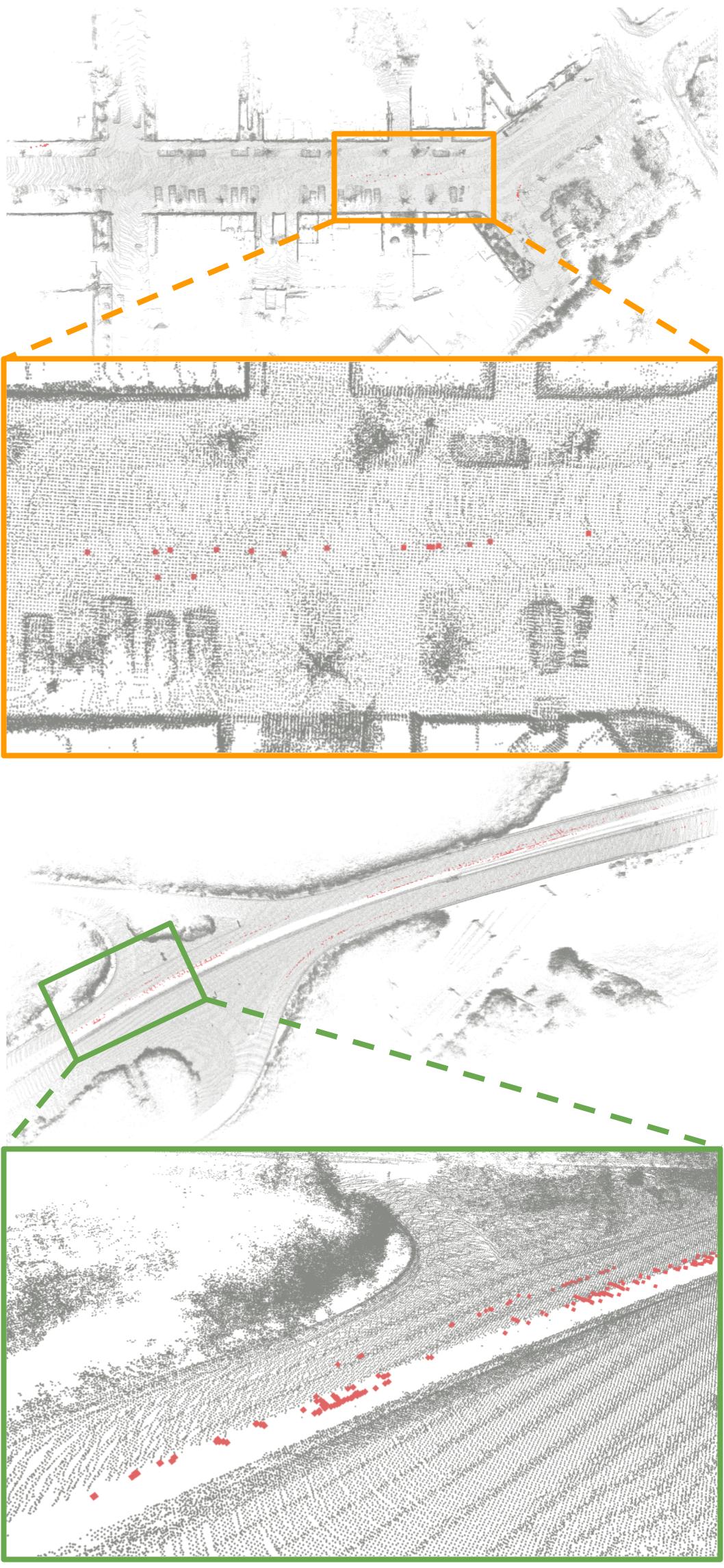}
\subcaption{\textnameSMAT{} (ours)\centering}
\end{minipage}
\caption{Qualitative comparison of \textnameSMAT{} and state-of-the-art methods for static map generation on sequences 00 and 01 of the SemanticKITTI dataset. Grey points represent static points, and a higher density of these points indicates greater efficiency in preserving static features. Red points, on the other hand, represent dynamic points, and a lower density of these points indicates a method's superior ability to remove dynamic objects.}
\label{fig:kitti_comparison}
\end{figure*}

\prettyref{fig:kitti_comparison} summarizes a qualitative comparison of different methods. We can observe that OctoMap's dynamic point removal is too aggressive, resulting in a sparse static map with a significant number of static points incorrectly removed (\prettyref{fig:kitti_comparison_octomap}).
Removert and 4DMOS have many dynamic points incorrectly retained in their static maps (\prettyref{fig:kitti_comparison_removert} and \prettyref{fig:kitti_comparison_4dmos}).
Both ERASOR and \textnameSMAT{} generate high-quality static maps. However, ERASOR occasionally removes ground points erroneously, resulting in small holes in the static map. Additionally, ERASOR assumes that dynamic objects always contact the ground, which is generally not true. For example, vegetation shading can obscure ground areas, making it challenging for ERASOR to remove points from vehicles in such regions (\prettyref{fig:kitti_comparison_erasor}).

\begin{table}
\centering
\caption{Comparison of mapping quality between \textnameSMAT{} and state-of-the-art methods on the SemanticKITTI dataset.}
\label{tab:kitti}
\resizebox{0.48\textwidth}{!}{%
\begin{tabular}{clccc}
\hline
\rowcolor[HTML]{EFEFEF} 
\multicolumn{1}{l}{\cellcolor[HTML]{EFEFEF}Sequence \#}& Method & PR{[}\%{]}     & RR{[}\%{]}     & $\text{F}_1$ score \\ \hline
                     & OctoMap~\cite{hornung2013octomap}        & 76.73          & \textbf{99.12} & 0.865          \\
                     & Removert~\cite{kim2020remove}            & 86.83          & 90.62          & 0.887          \\
                     & ERASOR~\cite{lim2021erasor}              & 93.98          & 97.08          & 0.955          \\
                     & 4DMOS~\cite{mersch2022receding}          & \textbf{99.98} & 91.67          & 0.956          \\
                     & DynamicFilter~\cite{fan2022dynamicfilter} & 90.07         & 91.09          & 0.906          \\
\multirow{-6}{*}{00} & \textnameSMAT{} (ours)                                     & 95.00          & 98.14          & \textbf{0.966} \\ \hline
                     & OctoMap~\cite{hornung2013octomap}        & 53.16          & \textbf{99.66} & 0.693          \\
                     & Removert~\cite{kim2020remove}            & 95.82          & 57.08          & 0.715          \\
                     & ERASOR~\cite{lim2021erasor}              & 91.49          & 95.38          & 0.934          \\
                     & 4DMOS~\cite{mersch2022receding}          & \textbf{99.81} & 83.28          & 0.908          \\
                     & DynamicFilter~\cite{fan2022dynamicfilter} & 87.95         & 87.69          & 0.878          \\
\multirow{-6}{*}{01} & \textnameSMAT{} (ours)                                     & 94.47          & 96.97          & \textbf{0.957} \\ \hline
                     & OctoMap~\cite{hornung2013octomap}        & 54.11          & 98.77          & 0.699          \\
                     & Removert~\cite{kim2020remove}            & 83.29          & 88.37          & 0.858          \\
                     & ERASOR~\cite{lim2021erasor}              & 87.73          & 97.01          & 0.921          \\
                     & 4DMOS~\cite{mersch2022receding}          & \textbf{99.84} & 95.75          & \textbf{0.978} \\
                     & DynamicFilter~\cite{fan2022dynamicfilter} & 88.02         & 86.10          & 0.871          \\
\multirow{-6}{*}{02} & \textnameSMAT{} (ours)                                     & 87.13          & \textbf{99.20} & 0.928          \\ \hline
                     & OctoMap~\cite{hornung2013octomap}        & 76.34          & 96.79          & 0.854          \\
                     & Removert~\cite{kim2020remove}            & 88.17          & 79.98          & 0.839          \\
                     & ERASOR~\cite{lim2021erasor}              & 88.73          & \textbf{98.26} & 0.921          \\
                     & 4DMOS~\cite{mersch2022receding}          & \textbf{99.88} & 86.93          & 0.930          \\
                     & DynamicFilter~\cite{fan2022dynamicfilter} & 90.17         & 84.65          & 0.873          \\
\multirow{-6}{*}{05} & \textnameSMAT{} (ours)                                     & 96.01          & 96.27          & \textbf{0.961} \\ \hline
                     & OctoMap~\cite{hornung2013octomap}        & 77.84          & 96.94          & 0.863          \\
                     & Removert~\cite{kim2020remove}            & 82.04          & 95.50          & 0.883          \\
                     & ERASOR~\cite{lim2021erasor}              & 90.62          & \textbf{99.27} & 0.948          \\
                     & 4DMOS~\cite{mersch2022receding}          & \textbf{97.25} & 79.08          & 0.872          \\
                     & DynamicFilter~\cite{fan2022dynamicfilter} & 87.94         & 86.80          & 0.874          \\
\multirow{-6}{*}{07} & \textnameSMAT{} (ours)                                     & 92.57          & 97.59          & \textbf{0.950} \\ \hline
\end{tabular}%
}
\end{table}

\noindent{\textbf{Comparison in customized simulation scenario.}}
While the KITTI dataset has been extensively used to evaluate SLAM and perception algorithms, it lacks frequent instances of dynamic objects. Therefore, many state-of-the-art methods, such as those presented in~\cite{lim2021erasor}, can already achieve PR and RR values exceeding 90\% on this dataset. In other words, this dataset cannot comprehensively evaluate the accuracy of dynamic object removal.

\begin{figure}[ht]
\centering
\includegraphics[width=1.0\linewidth]{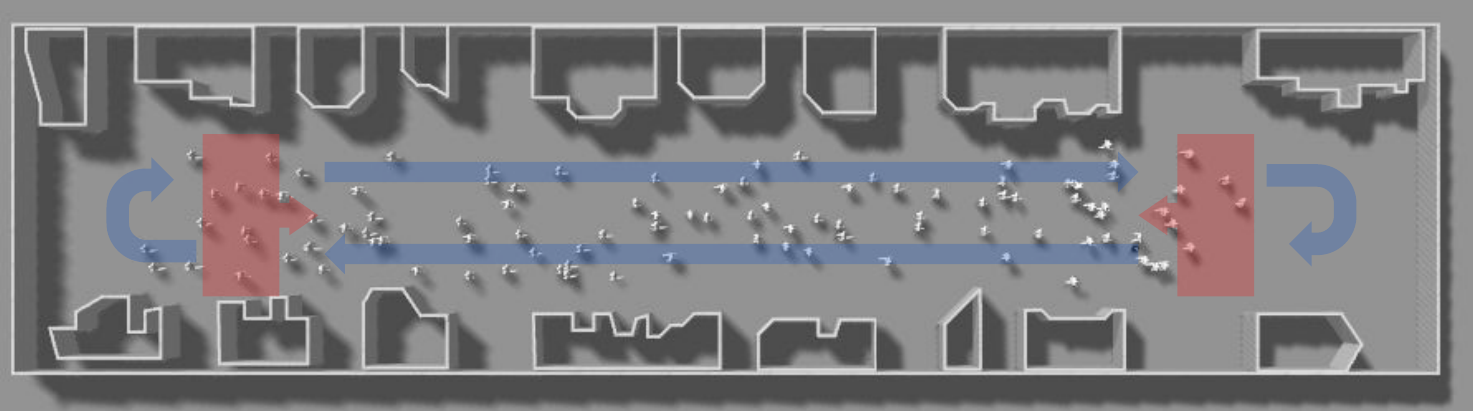}
\caption{Highly dynamic simulation scenario covering an area of \SI{70}{\meter} $\times$ \SI{10}{\meter}. The moving flow of pedestrians is indicated by the red arrow, while the robot's data collection trajectory is depicted by the blue arrow.}
\label{fig:simulation}
\end{figure}

To address this issue, we have developed a simulation environment that incorporates a larger number of dynamic objects, as depicted in \prettyref{fig:simulation}. This simulation environment utilizes the Gazebo simulator~\cite{Gazebo_simulator} to simulate a mobile robot equipped with a 3D LiDAR and employs the Menge framework~\cite{Menge} to simulate the movement of pedestrians. We conducted experiments with varying numbers of pedestrians: 50, 100, and 150. 
In these experiments, the robot follows a full loop path to collect LiDAR scans. Among the total collected points, the dynamic points account for 54.8\%, 58.5\%, and 59.3\% in these three experiments, respectively.

\begin{table}
\centering
\caption{Comparison of mapping quality between \textnameSMAT{} and state-of-the-art methods in simulation.}
\label{tab:gazebo_exp}
\resizebox{0.48\textwidth}{!}{%
\begin{tabular}{clccc}
\hline
\rowcolor[HTML]{EFEFEF}
\multicolumn{1}{l}{pedestrians \#}    & Method                            & PR{[}\%{]}     & RR{[}\%{]}     & F1 score       \\ \hline
                                      & OctoMap~\cite{hornung2013octomap} & 95.90          & \textbf{99.73} & 0.978          \\
                                      & Removert~\cite{kim2020remove}     & 80.77          & 86.62          & 0.836          \\
                                      & ERASOR~\cite{lim2021erasor}       & 70.12          & 80.92          & 0.751          \\
                                      & 4DMOS~\cite{mersch2022receding}   & \textbf{99.50} & 23.72          & 0.383          \\
                                      & DynamicFilter~\cite{fan2022dynamicfilter} & 91.91  & 96.09          & 0.940          \\
\multirow{-6}{*}{50}                  & \textnameSMAT{} (ours)                              & 96.11          & 99.68          & \textbf{0.979} \\ \hline
                                      & OctoMap~\cite{hornung2013octomap} & 95.59          & 99.71          & 0.976          \\
                                      & Removert~\cite{kim2020remove}     & 83.14          & 80.26          & 0.817          \\
                                      & ERASOR~\cite{lim2021erasor}       & 68.99          & 81.38          & 0.747          \\
                                      & 4DMOS~\cite{mersch2022receding}   & \textbf{99.40} & 19.36          & 0.324          \\
                                      & DynamicFilter~\cite{fan2022dynamicfilter} & 91.74  & 91.73          & 0.917          \\
\multirow{-6}{*}{100}                 & \textnameSMAT{} (ours)                              & 96.01          & \textbf{99.72} & \textbf{0.978} \\ \hline
                                      & OctoMap~\cite{hornung2013octomap} & \textbf{95.76} & 99.60          & \textbf{0.976} \\
                                      & Removert~\cite{kim2020remove}     & 85.25          & 71.18          & 0.776          \\
                                      & ERASOR~\cite{lim2021erasor}       & 68.17          & 77.68          & 0.726          \\
                                      & 4DMOS~\cite{mersch2022receding}   & \textbf{99.44} & 17.38          & 0.296          \\
                                      & DynamicFilter~\cite{fan2022dynamicfilter} & 90.12  & 89.26          & 0.897          \\
\multirow{-6}{*}{150}                 & \textnameSMAT{} (ours)                              & 95.13          & \textbf{99.61} & 0.973          \\ \hline
\end{tabular}%
}
\end{table}

\begin{table*}[ht]
\centering
\caption{Comparison of tracking quality between \textnameSMAT{} and state-of-the-art methods on JRDB 3D MOT benchmarks.}
\label{tab:jrdb_results}
\resizebox{0.99\linewidth}{!}{%
\begin{tabular}{l|cccc|cccc|ccc}
\hline
\rowcolor[HTML]{EFEFEF}
Method                       & MOTA{[}\%{]} $\uparrow$ & FN $\downarrow$ & FP $\downarrow$ & IDSW $\downarrow$ & IDF1{[}\%{]} $\uparrow$ & IDTP $\uparrow$ & IDFN $\downarrow$ & IDFP $\downarrow$ & HOTA{[}\%{]} $\uparrow$ & DetA $\uparrow$ & AssA $\uparrow$   \\ \hline
AB3DMOT~\cite{weng20203d}    & 19.34                   & 777968          & \textbf{13686}  & 6179          & 10.67                   & 64771           & 924383           & \textbf{160101}  
 & 11.81                   & 12.07           & 12.05 \\
JRMOT~\cite{shenoi2020jrmot} & 20.15                   & 765901          & 19705           & \textbf{4216} & 12.32                   & 75889           & 913265           & 167069           
 & 13.05                   & 12.60           & 13.96 \\ 
SS3D-MOT~\cite{liu2022ss3d}  & 22.96                   & \textbf{690001} & 52041           & 19973         & 14.48                   & 97041           & 892113           & 254153           
 & 15.80                   & \textbf{23.22 } & 10.89  \\ 
\textnameSMAT{} (ours)                         & \textbf{24.21}          & 702364          & 42577           & 4739          & \textbf{21.34}          & \textbf{140702} & \textbf{848452}  & 188665           
 & \textbf{18.93}          & 17.06          & \textbf{21.13}  \\ \hline
\end{tabular}%
}
\end{table*}

We compared \textnameSMAT{} with other methods in simulated experiments, and the results are summarized in \prettyref{tab:gazebo_exp}. Removert effectively removes dynamic points and preserves static features when there are 50 pedestrians, but its performance in dynamic point removal significantly declines as the number of pedestrians increases. For instance, when there are 150 pedestrians, Removert achieves only an RR of 71.18\%. 4DMOS cannot detect dynamic objects well in the simulation scene because its perceptual model is trained on the KITTI dataset. Additionally, its RR performance is poor because it does not have optimization steps for multi-frame mapping, and any dynamic objects missed by detection will be permanently included in the map. ERASOR, as a model-dependent method that relies on ground fitting to eliminate dynamic points above the ground, is ineffective in highly dynamic scenarios where determining the ground plane becomes difficult. 

\begin{figure}[ht]
\centering 
\begin{minipage}[t]{0.99\linewidth}
\centering
\includegraphics[width=0.99\linewidth]{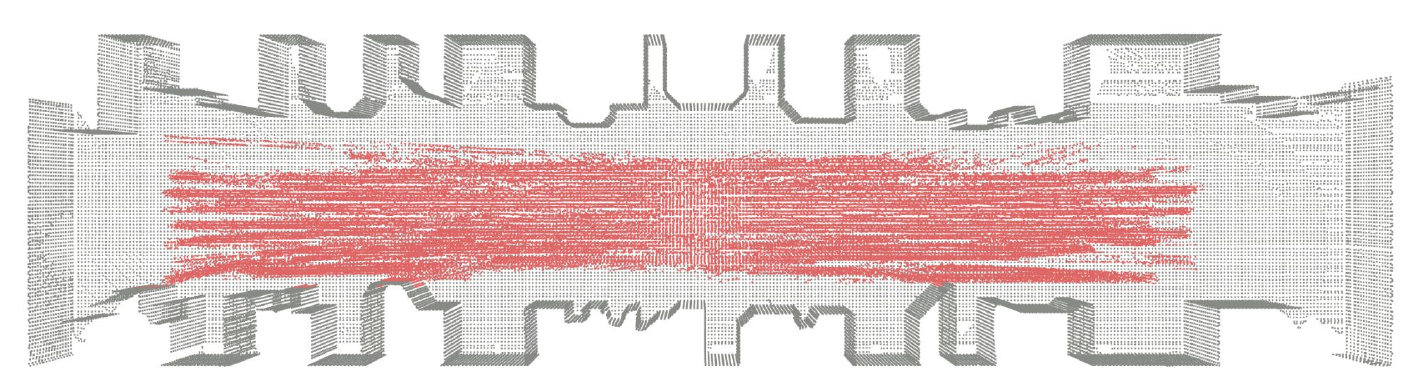}
\subcaption{Groud-truth map (grey) and without dynamic object removal (red) \centering}
\end{minipage}
\begin{minipage}[t]{0.99\linewidth}
\centering
\includegraphics[width=0.99\linewidth]{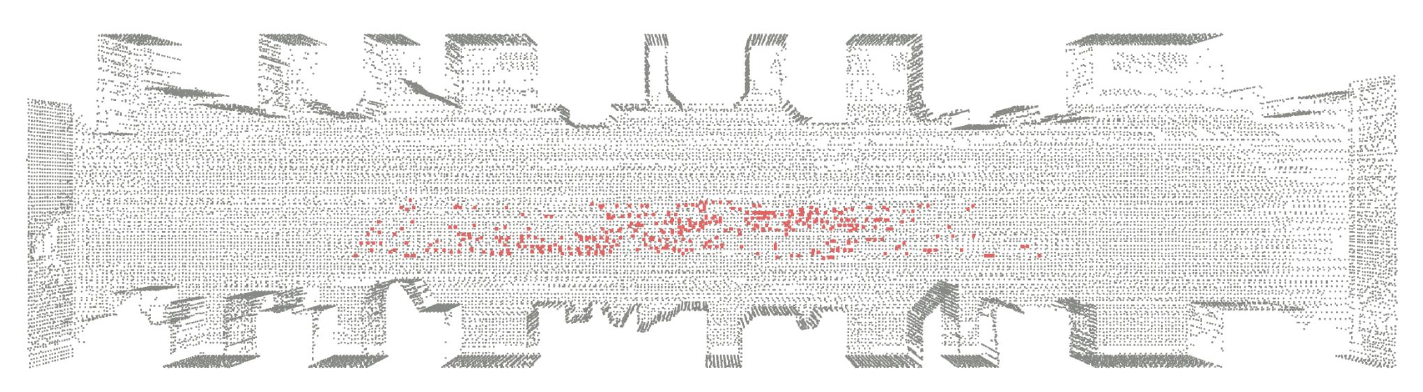}
\subcaption{OctoMap~\cite{hornung2013octomap}\centering}
\end{minipage}
\begin{minipage}[t]{0.99\linewidth}
\centering
\includegraphics[width=0.99\linewidth]{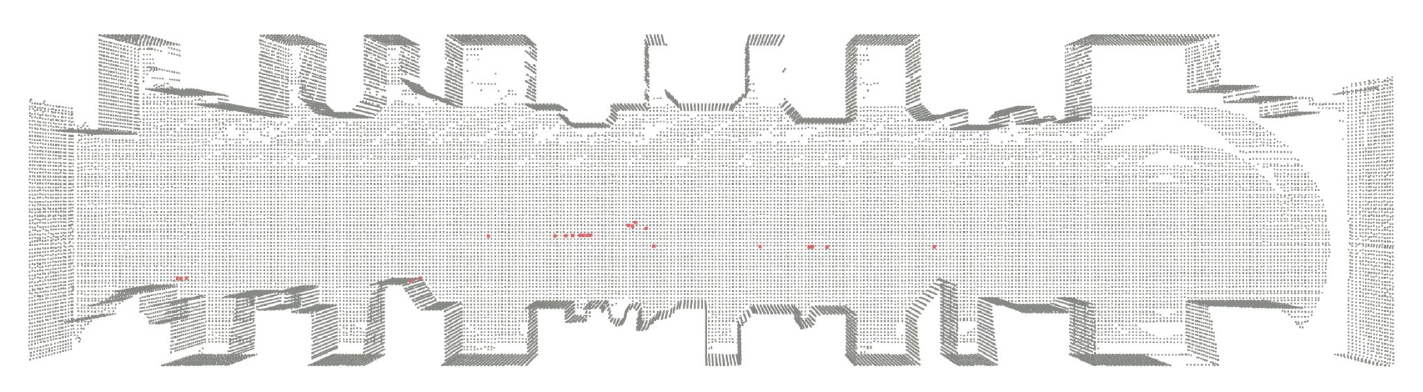}
\subcaption{\textnameSMAT{} (ours)\centering}
\end{minipage}
\caption{Qualitative comparison of static map produced by (a) SLAM without dynamic object removal, (b) Octomap \protect\cite{hornung2013octomap} and (c) \textnameSMAT{}, in simulation with 50 pedestrians. Grey points represent static points, and a higher density of grey points indicates a method's superior ability to retain static features. Red points represent dynamic points, and a lower density of red points indicates a method's better ability to remove dynamic objects.}
\label{fig:simulation_comparison}
\end{figure}

Both \textnameSMAT{} and OctoMap consistently achieve PR values exceeding 95\% and successfully remove nearly all dynamic points in all three experiments. However, OctoMap's good performance heavily relies on loop closure. It updates the occupancy probability of all voxels in the global map by accumulating multiple observations of the scene when the robot moves back and forth. Due to transient occupancy, it occasionally misclassifies static points as dynamic points and removes them incorrectly. The long-range correlation provided by loop closure is crucial to compensate for this error. To assess the performance of \textnameSMAT{} and Octomap in cases without loop closure, we designed one experiment where the robot moves from left to right in the dynamic scene of \prettyref{fig:simulation} with 50 pedestrians. Octomap has a PR of 98.30\%, but its RR drops significantly to 82.58\%, leading to an $\text{F}_1$ of 90.1\%. In contrast, \textnameSMAT{} consistently achieves a PR of 90.93\% and an RR of 99.53\%, leading to an much better $\text{F}_1$ of 95.1\%. As shown in \prettyref{fig:simulation_comparison}, Octomap produces a static map with sparser ground coverage and retains more dynamic points than \textnameSMAT{}'s static map.

\subsection{Multi-object tracking performance in human-populated scenes}
\label{sec:mot_eval}
Multi-object tracking (MOT) aims to detect dynamic objects and establish associations between detections over time based on object features. To assess the MOT performance among different methods, we use three commonly employed metrics: MOTA \cite{bernadin2008Evaluating}, IDF1 \cite{ristani2016performance}, and HOTA \cite{luiten2021hota}. Each metric focuses on different aspects of MOT evaluation. MOTA emphasizes the accuracy of object detection, IDF1 emphasizes the effectiveness of association, and HOTA combines both detection and association errors in a balanced manner \cite{luiten2021hota}. For a comprehensive discussion of these metrics, please refer to \prettyref{apd:mot_metrics}.

\begin{figure}
\centering 
\includegraphics[trim={90 60 95 80},clip,width=1\linewidth]{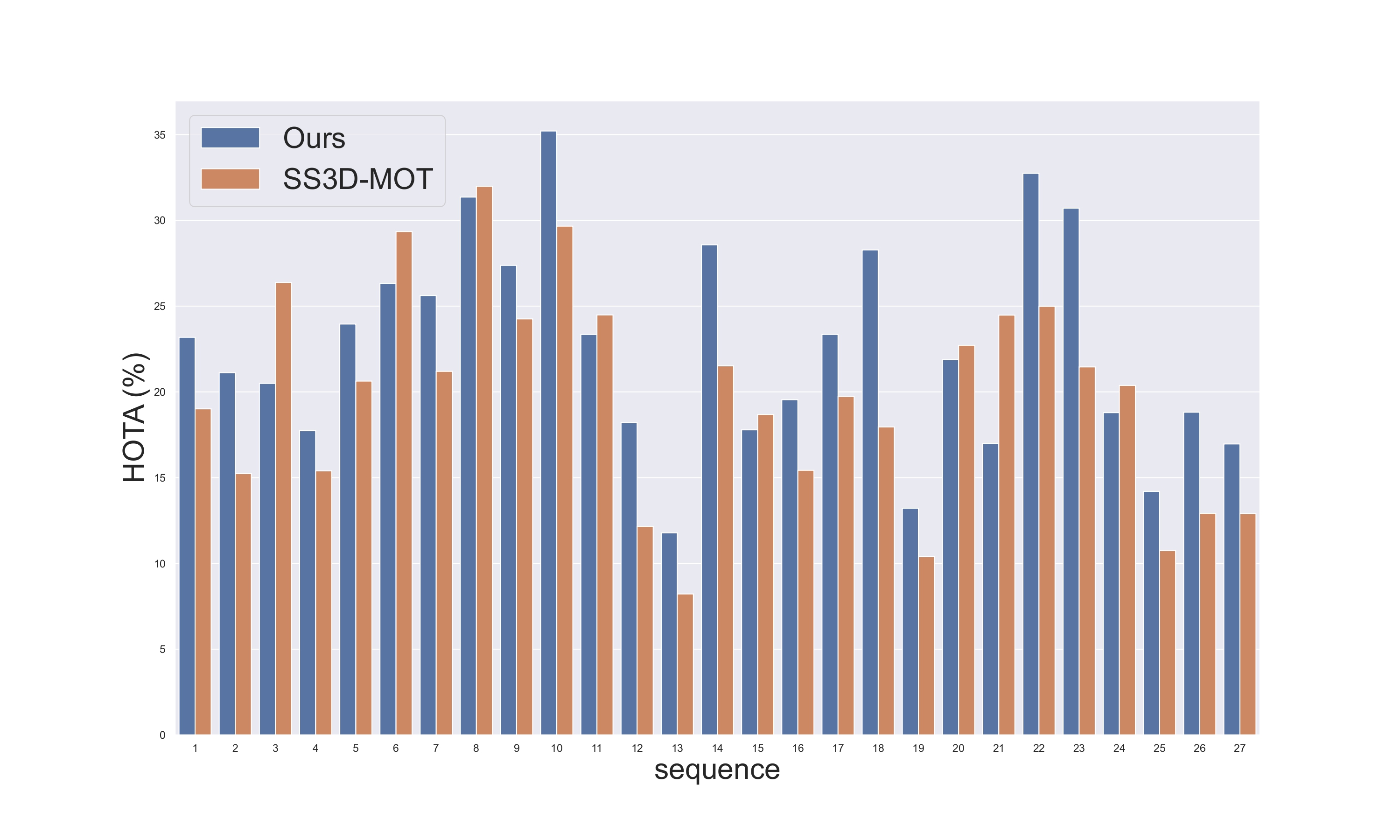}
\caption{HOTA of \textnameSMAT{} and SS3D-MOT \protect\cite{liu2022ss3d} on all test sequences of the JRDB dataset.}
\label{fig:jrdb_hota}
\end{figure}

\begin{figure*}[ht]
\centering
\begin{minipage}[t]{0.41\linewidth}
\centering
\includegraphics[width=0.99\linewidth]{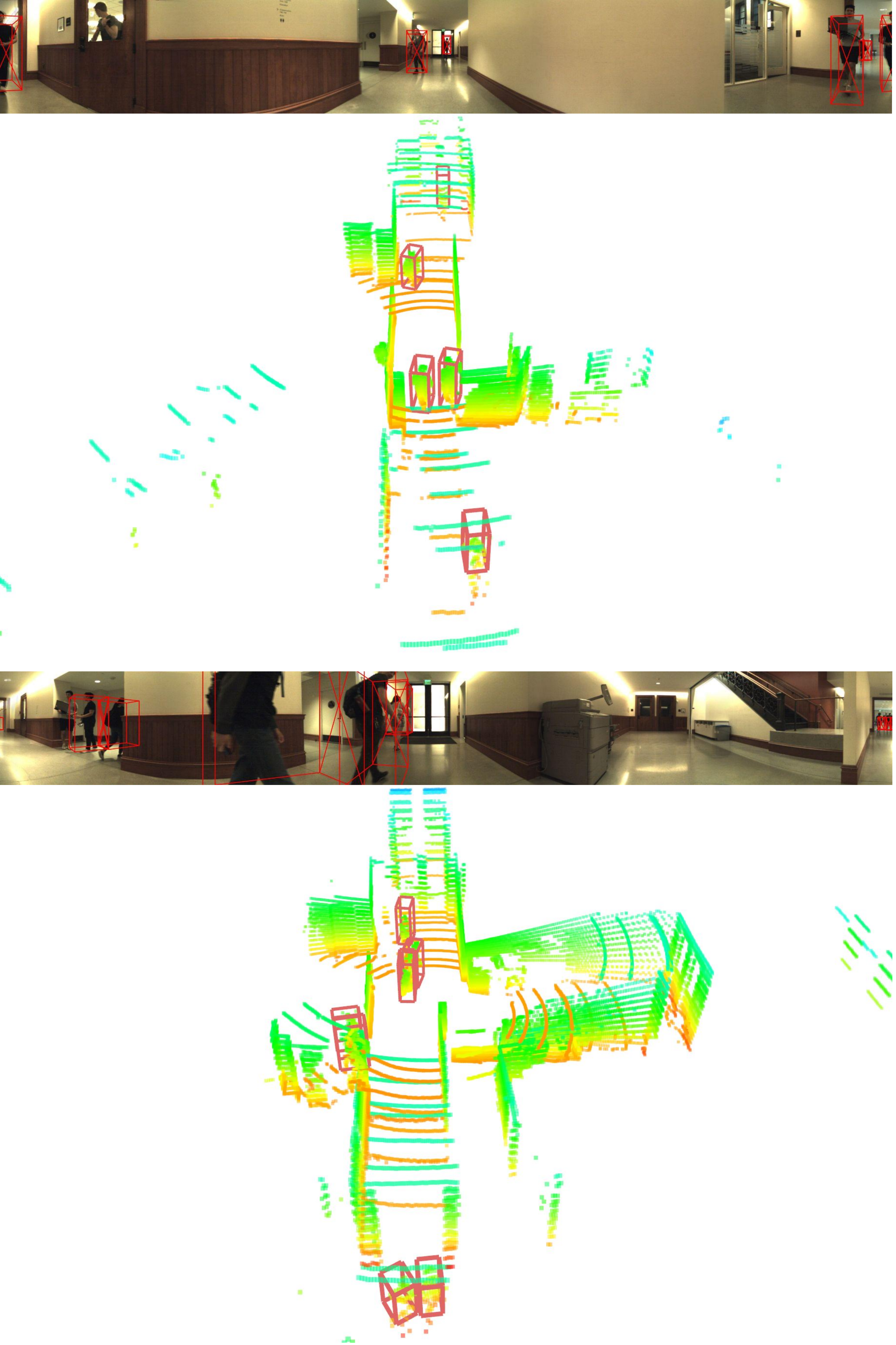}
\subcaption{\centering sequence 23}
\end{minipage}
\begin{minipage}[t]{0.41\linewidth}
\centering
\includegraphics[width=0.99\linewidth]{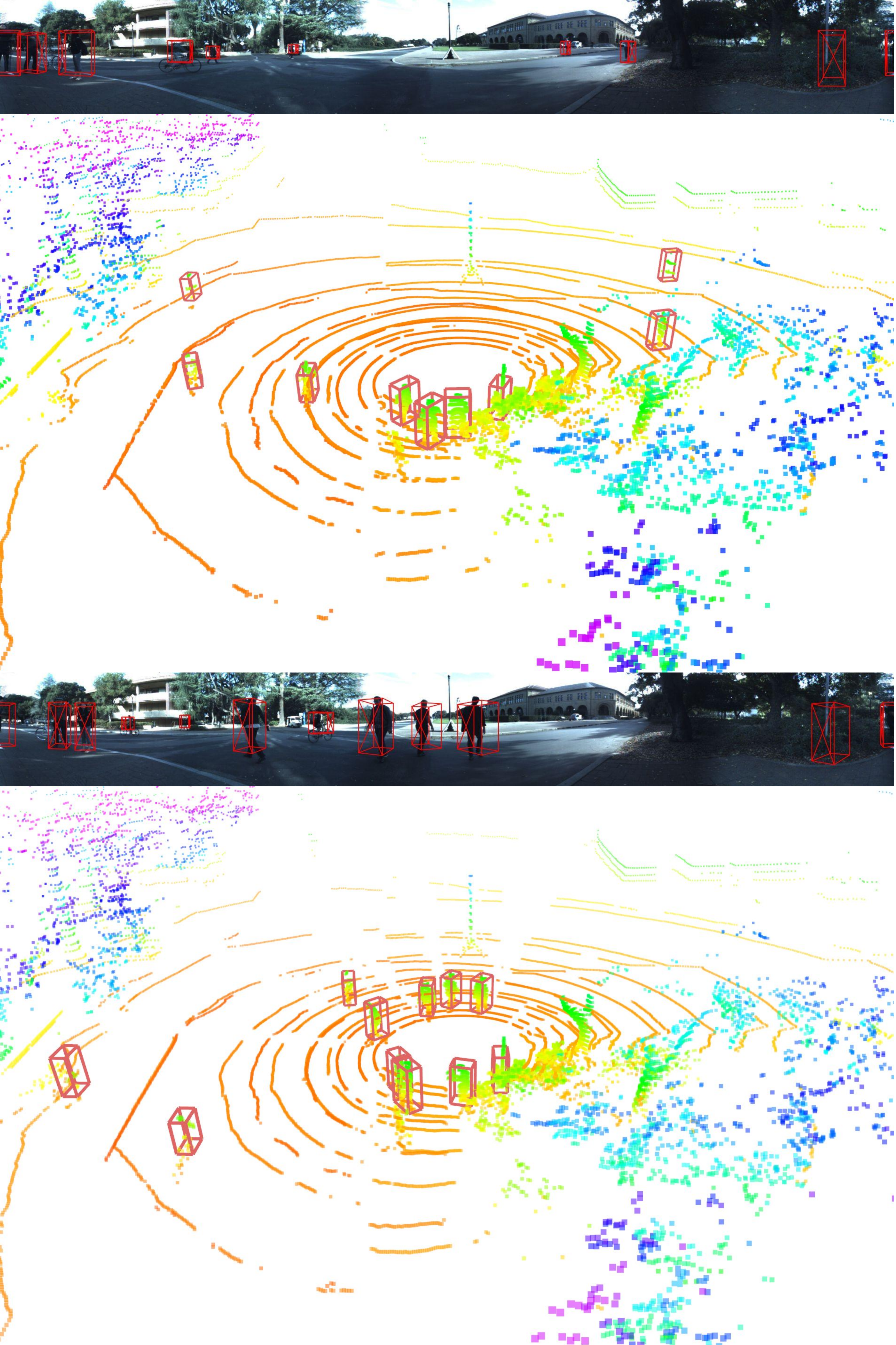}
\subcaption{\centering sequence 14}
\end{minipage} \\
\begin{minipage}[t]{0.41 \linewidth}
\centering
\includegraphics[width=0.99\linewidth]{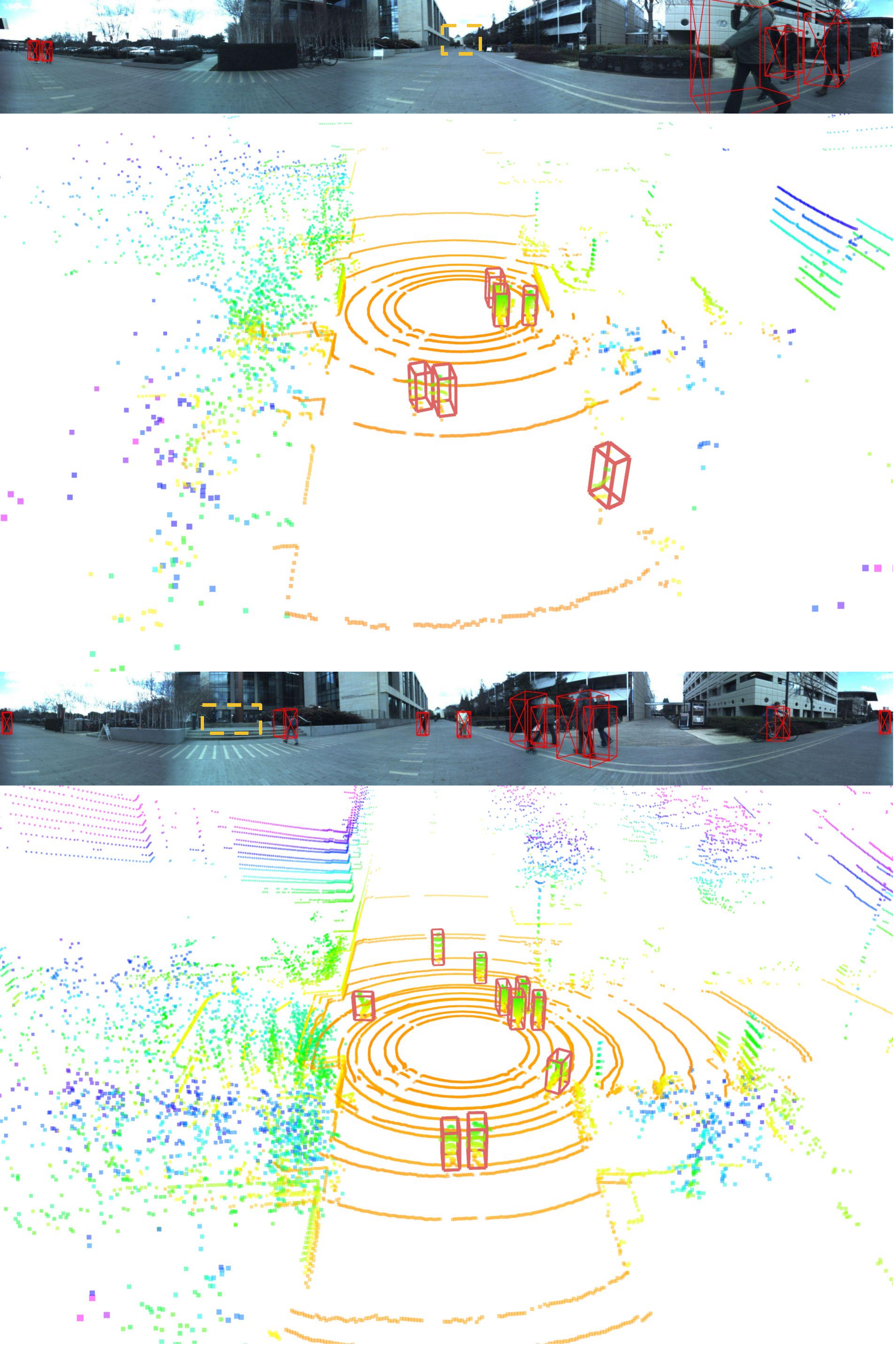}
\subcaption{\centering sequence 3}
\end{minipage}
\begin{minipage}[t]{0.41 \linewidth}
\centering
\includegraphics[width=0.99\linewidth]{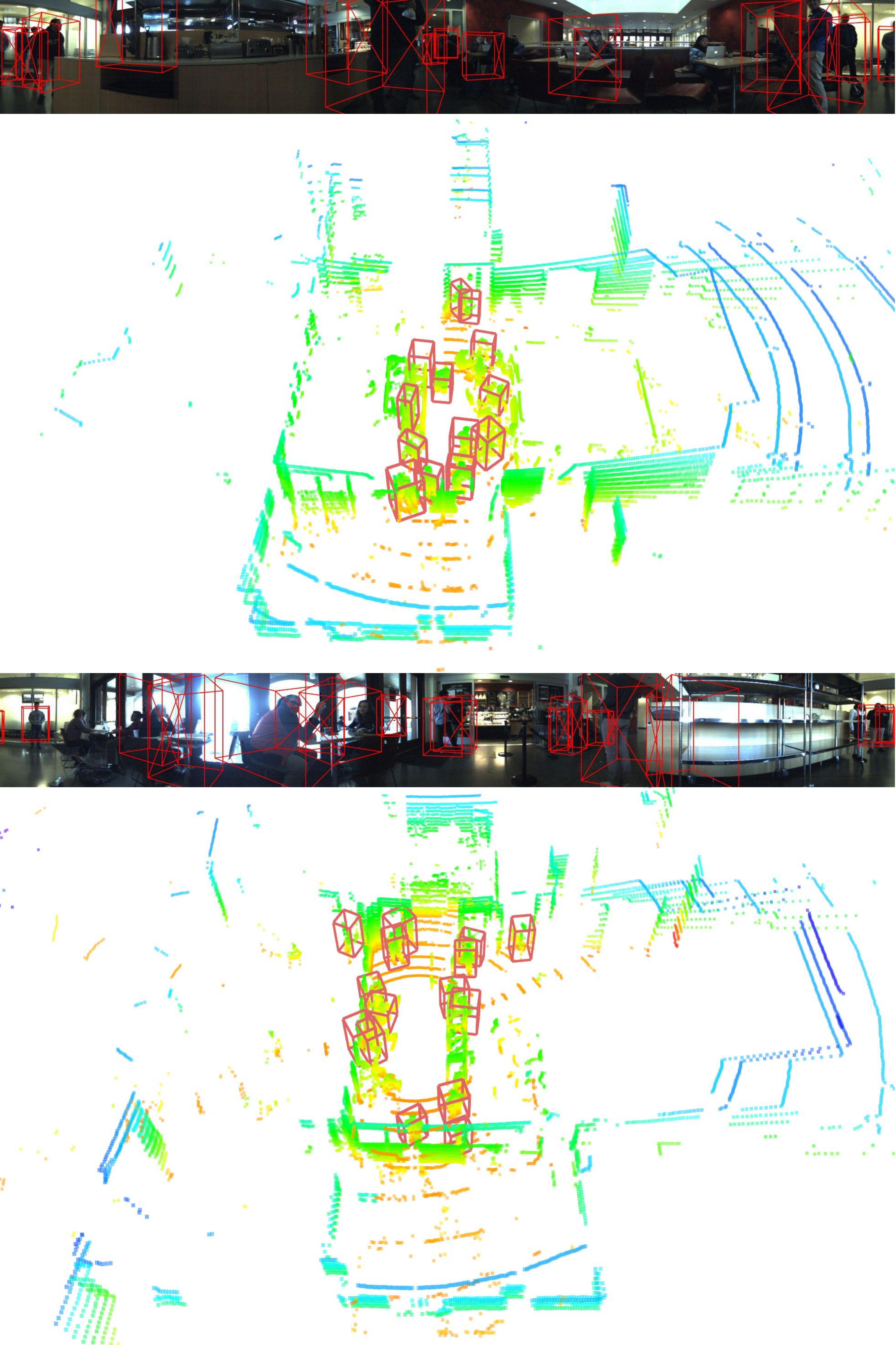}
\subcaption{\centering sequence 13}
\end{minipage}
\caption{Qualitative tracking results of \textnameSMAT{} using point clouds only on test sequences of the JRDB dataset. 
For each sequence, we display results for two frames. Each frame includes an RGB image (top) and corresponding point clouds (bottom). The red bounding boxes in the point clouds are \textnameSMAT{}'s results. They are then projected onto the RGB image based on the relative poses of the cameras and LiDAR, resulting in bounding boxes in RGB. (a) and (b) demonstrate indoor and outdoor situations where \textnameSMAT{} performs well. (c) is a situation where \textnameSMAT{} exhibits relatively poor performance, with pedestrians in yellow boxes on the RGB images being missed due to their distance from the robot. (d) is a highly dynamic and unstructured scene, which poses challenges for all methods, including \textnameSMAT{}.
}
\label{fig:jrdb_visualization}
\end{figure*}

We evaluate the MOT performance on the JRDB dataset \cite{martin2021jrdb}, which is a comprehensive perception dataset specifically designed for social robots. The dataset includes 64 minutes of sensor data, including RGB images and 3D point clouds captured in various campus scenarios, both indoors and outdoors. With a focus on human social environments, this dataset provides over 1.8 million annotated 3D human bounding boxes. It consists of a total of 54 sequences, with 27 sequences for training and the remaining sequences for testing.

\noindent{\textbf{Remarks on dynamic object detection.}} Although the JRDB dataset includes both RGB images and 3D point clouds, our method relies solely on point clouds for object detection. However, due to the sparsity of point clouds compared to images, objects located at far distances may not be visible using point clouds alone. To overcome this limitation, the 3D MOT benchmarks provided by JRDB offer a \emph{public detection} generated by JRMOT \cite{shenoi2020jrmot} for users to incorporate into their tracking algorithms. Therefore, we first merge the public detection with \textnameSMAT{}’s point-cloud-only detection and then utilize the merged detection in \textnameSMAT{}’s perception component for tracking. It is important to note that we only use the detection from RGB images in this specific dataset to ensure a fair comparison with other tracking algorithms.

JRMOT's public detection provides estimated human bounding boxes accompanied by confidence scores ranging from 0 to 1. In our implementation, we consider bounding boxes with confidence scores higher than 0.3. \textnameSMAT{}'s point-cloud-only detection result is denoted as $\mathbb{B}_o = \{\mathcal{B}_o^1, \mathcal{B}_o^2, \cdots, \mathcal{B}_o^m\}$, while the selected public detection result is denoted as $\mathbb{B}_p = \{\mathcal{B}_p^1, \mathcal{B}_p^2, \cdots, \mathcal{B}_p^n\}$. If a bounding box $\mathcal{B}_o^i$ in $\mathbb{B}_o$ overlaps with some bounding boxes in $\mathbb{B}_p$, $\mathcal{B}_o^i$ will be replaced by the overlapped bounding box $\mathcal{B}_p^j$ from $\mathbb{B}_p$ that has the maximum intersection with $\mathcal{B}_o^i$. Additionally, if the confidence score of a bounding box in $\mathbb{B}_p$ exceeds 0.8 and it does not overlap with any bounding boxes in $\mathbb{B}_o$, it will also be added to $\mathbb{B}_o$. In this way, the more accurate appearance-based detection results are merged into \textnameSMAT{}'s detection results.


\textnameSMAT{} is specifically designed to track moving objects, while the 3D MOT benchmark provided by JRDB requires tracking all pedestrians, regardless of their movements. As a result, the merged detection may include some bounding boxes that do not meet our criteria for stable tracking hypotheses, as they may correspond to stationary humans. These bounding boxes will not be directly incorporated into \textnameSMAT{} but will be tracked in a parallel process using \emph{tracking by detection} that we described in \prettyref{sec:frontend}.

\noindent{\textbf{Comparison on JRDB dataset.}}
\prettyref{tab:jrdb_results} compares \textnameSMAT{} with other baselines on the 3D MOT benchmark provided by JRDB. \textnameSMAT{} stands out as the leader among all online 3D MOT methods, surpassing the previous benchmark leader SS3D-MOT \cite{liu2022ss3d} by significant margins across multiple evaluation metrics: 5.44\% in MOTA, 47.38\% in IDF1, and 19.81\% in HOTA. In terms of detection error, \textnameSMAT{} effectively reduces the number of false negatives (FN) by 63,537 compared to the JRMOT baseline, while introducing a modest increase in false positives (FP) by 22,872. This improvement can be attributed to \textnameSMAT{}'s exceptional capability to detect moving pedestrians by analyzing the differences between the current frame and the real-time spatial structure map. It also enables \textnameSMAT{} to mitigate instances where stationary objects, such as tables and chairs, are mistakenly identified as individuals.


\textnameSMAT{} achieves the best performance among all methods in terms of both dynamic object detection accuracy measured by DetA (Detection Accuracy) and dynamic object association accuracy measured by IDF1 and AssA (Association Accuracy). \textnameSMAT{} demonstrates a substantial 35.40\% improvement in DetA compared to baselines, securing the second-highest DetA among all methods. Additionally, \textnameSMAT{} outperforms all other methods according to the AssA and IDF1 metrics, with a 47.38\% improvement in IDF1 and a 51.36\% improvement in AssA. \textnameSMAT{}'s strong performance can be partially attributed to the detection robustness provided by \emph{detection by tracking} in challenging situations. For example, when an object that has been consistently tracked suddenly disappears in a frame, \emph{detection by tracking} generates an object hypothesis through motion estimation. This hypothesis is treated as a valid moving object with sufficient dynamic points, effectively addressing tracking discontinuity caused by missed detections.

\begin{figure}[ht]
\centering 
\includegraphics[width=1\linewidth]{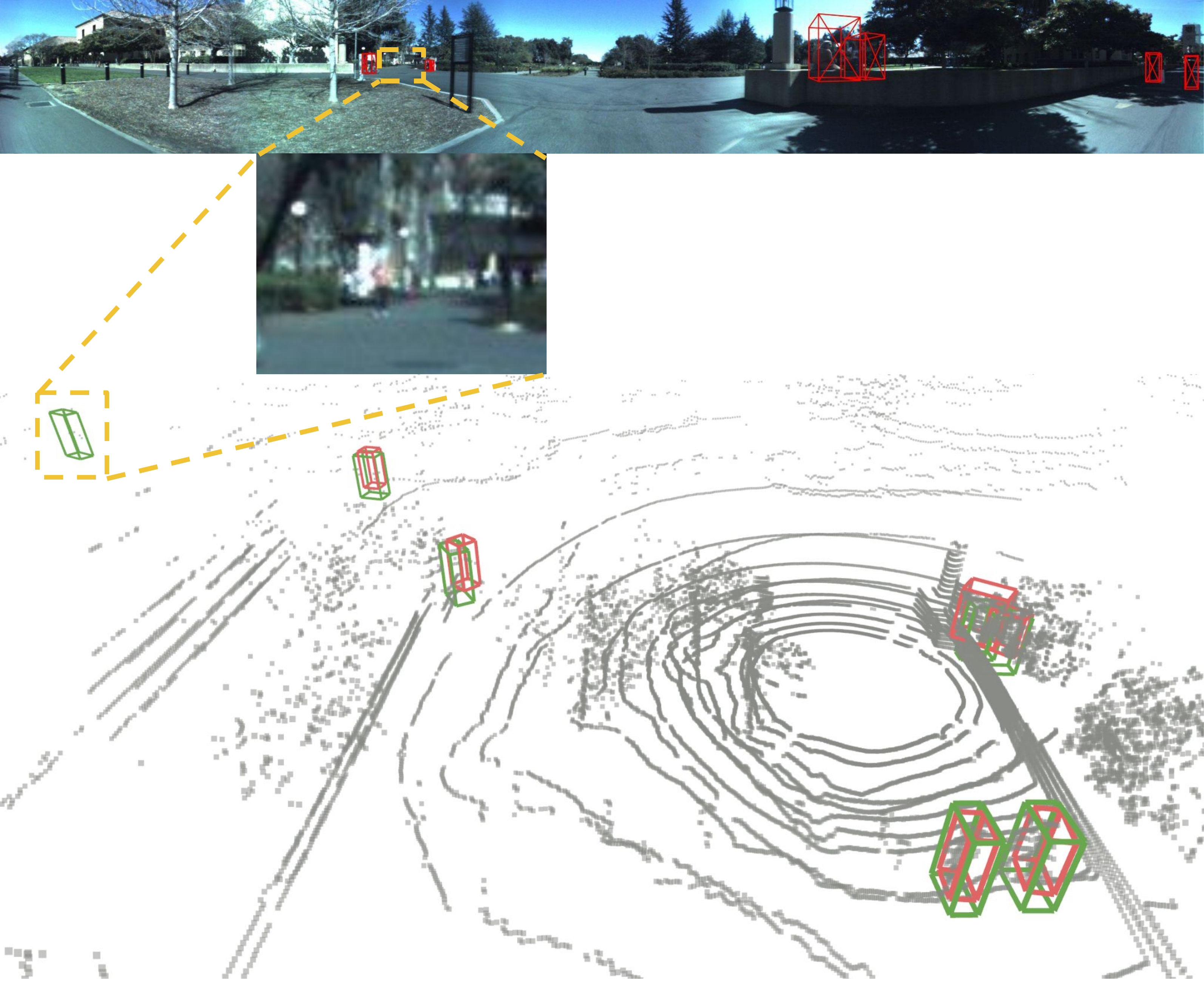}
\caption{\textnameSMAT{}'s detection missing due to sparse point clouds. This situation is taken from a sequence in the JRDB dataset. The red bounding boxes in the point clouds represent \textnameSMAT{}'s detection of dynamic objects, while the green bounding boxes indicate the ground truth. \textnameSMAT{} fails to detect the pedestrian on the leftmost side due to its distance of about \SI{20}{\metre} from the robot. In this case, the LiDAR cannot provide a sufficiently dense point cloud for accurate detection.}
\label{fig:jrdb_problem}
\end{figure}

\noindent{\textbf{Success and failure case studies.}}
The JRDB dataset contains a variety of campus scenarios, and \prettyref{fig:jrdb_hota} presents the HOTA scores of \textnameSMAT{} and SS3D-MOT on each test sequence. \textnameSMAT{} consistently outperforms SS3D-MOT in most scenarios, particularly in situations where there is a higher concentration of dynamic individuals near the robot. \prettyref{fig:jrdb_visualization}a and \prettyref{fig:jrdb_visualization}b showcase two typical indoor and outdoor cases where \textnameSMAT{} performs well. 

However, \textnameSMAT{}'s MOT performance may lag behind the learning-based SS3D-MOT when individuals are far from the robot. This is because \textnameSMAT{} employs clustering for object detection, which requires a certain density of points. When individuals are far away, LiDAR data is too sparse to describe distant objects adequately with sufficient points. For example, as shown in the RGB image on top of \prettyref{fig:jrdb_visualization}c, individuals within the yellow box cannot be detected and tracked by \textnameSMAT{} due to their sparse point clouds. Another example is in \prettyref{fig:jrdb_problem}, where individuals in the upper-left corner of the figure are not detected because they are located approximately 20 meters away from the robot, and thus their ground truth bounding box contains only a few points. However, such failure to detect and track objects at far distances is generally not an issue in practice, because robots have sufficient time to respond to these objects. 

\textnameSMAT{} may also exhibit unsatisfactory MOT performance in highly challenging and unstructured scenarios where SS3D also fails. For instance, \prettyref{fig:jrdb_visualization}d depicts a scenario with a multitude of individuals, tables, and chairs, where individuals may be detected as part of the furniture or furniture may be mistakenly identified as individuals.

\subsection{Ablation Study}
\label{sec:ablation_study}

To further investigate \textnameSMAT{}, we conducted a detailed ablation study in simulation with 150 pedestrians. We considered two variants of our framework: i) the \textit{front-end only} approach that only uses \textnameSMAT{}'s front-end, and ii) the \textit{back-end only} approach that only uses \textnameSMAT{}'s back-end. First, we compared the \textit{back-end only} approach with its variants to demonstrate the advantages of integrating both occupancy probability and visibility checking. Next, we conducted experiments to analyze the significance of coupling the front-end and back-end and to evaluate the effectiveness of the self-reinforcing mechanism.

\begin{table}
\centering
\caption{Comparison of \textit{back-end only} variants in simulation.}
\label{tab:ablation_back_end}
\resizebox{0.48\textwidth}{!}{%
\begin{tabular}{lcccc}
\hline
\rowcolor[HTML]{EFEFEF}
Method   & PR{[}\%{]} & RR{[}\%{]} & F1 score   &Runtime/scan{[}ms{]}\\ \hline
visibility check only & 95.53          & 72.43          & 0.824          & 7     \\
occupancy probability only      & 93.08          & \textbf{83.31} & 0.879          & 84    \\
full back-end only             & \textbf{96.87} & 80.90          & \textbf{0.882} & 10    \\ \hline
\end{tabular}%
}
\end{table}

\noindent{\textbf{Back-end module:}}
We compared \textit{back-end only} with two different variants: 
\begin{itemize}
    \item \textit{Visibility checking only}, which determines whether a voxel is occupied solely based on visibility checking, without taking into account occupancy probability. If an approximated ray passes through a voxel, it is considered free.
    \item \textit{Occupancy probability only} method, which computes occupancy probability using~\cite{hornung2013octomap}, without using visibility checking to approximate ray tracing.
\end{itemize}

As shown in \prettyref{tab:ablation_back_end}, calculating occupancy probability has a significant impact on RR. However, performing ray tracing for every scan without approximation is time-consuming, making it difficult for the \textit{occupancy probability only} approach to achieve real-time performance. On the other hand, the visibility checking's approximation of ray tracing effectively reduces computation while still maintaining a relatively high RR. Additionally, the visibility checking considers the occurrence of large incident angles and occlusion issues, resulting in a 3.79\% improvement in PR.

\begin{table}\tiny
\centering
\caption{Comparison of different \textnameSMAT{} variants in simulation.}
\label{tab:ablation_overall}
\resizebox{0.48\textwidth}{!}{%
\begin{tabular}{lcccc}
\hline
\rowcolor[HTML]{EFEFEF}
Method                     & PR{[}\%{]}     & RR{[}\%{]}     & F1 score         \\ \hline
front-end only             & \textbf{99.16} & 43.25          & 0.602            \\
back-end only              & 96.87          & 80.90          & 0.882            \\
DynamicFilter~\cite{fan2022dynamicfilter}               & 90.12          & 89.26          & 0.897            \\
\textnameSMAT{} w/o background subtraction & 86.66          & \textbf{99.78} & 0.928            \\
\textnameSMAT{}   & 95.13          & 99.61          & \textbf{0.973}   \\ \hline
\end{tabular}%
}
\end{table}

\begin{figure}[ht]
\centering
\begin{minipage}[t]{0.495\linewidth}
\centering
\includegraphics[trim={0 0 45 0},clip,width=1\linewidth]{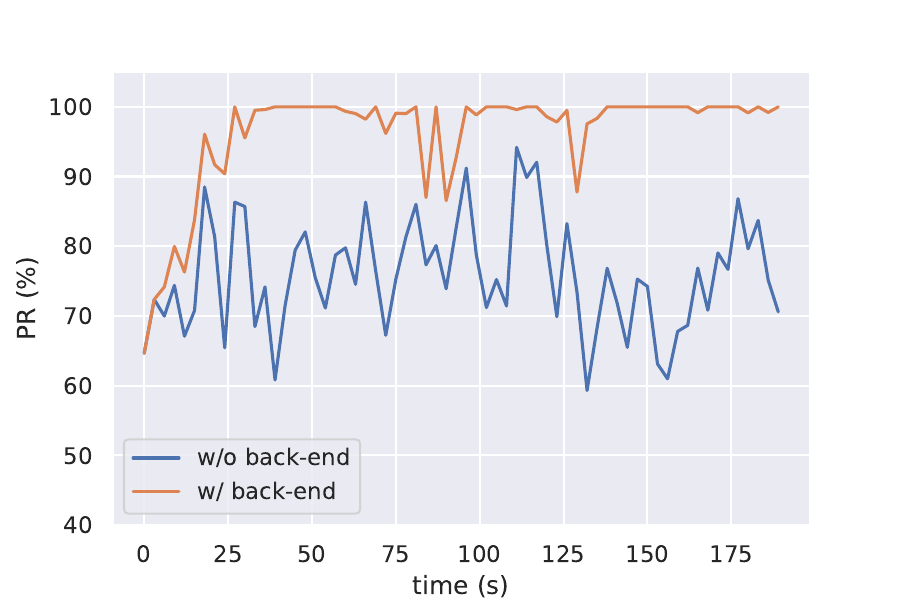}
\subcaption{PR of the front-end module\centering}
\label{fig:ablation_study_front_end_pr}
\end{minipage}
\begin{minipage}[t]{0.495\linewidth}
\centering
\includegraphics[trim={0 0 45 0},clip,width=1\linewidth]{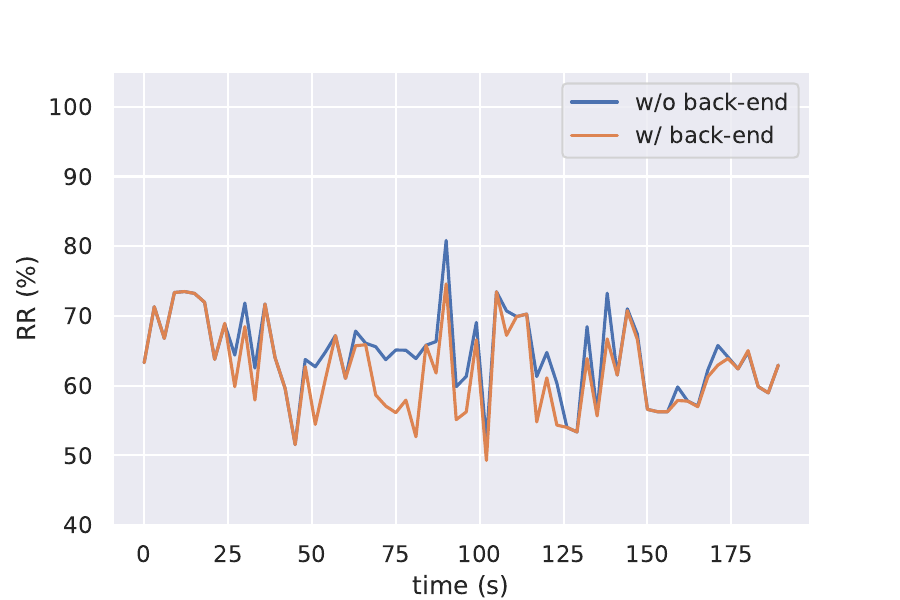}
\subcaption{RR of the front-end module\centering}
\label{fig:ablation_study_front_end_rr}
\end{minipage}
\caption{PR and RR of static scans generated by front-end with and without the back-end module, plotted over time.}
\label{fig:ablation_study_front_end}
\end{figure}

\begin{figure}[ht]
\centering
\begin{minipage}[t]{0.495\linewidth}
\centering
\includegraphics[trim={0 0 45 0},clip,width=1\linewidth]{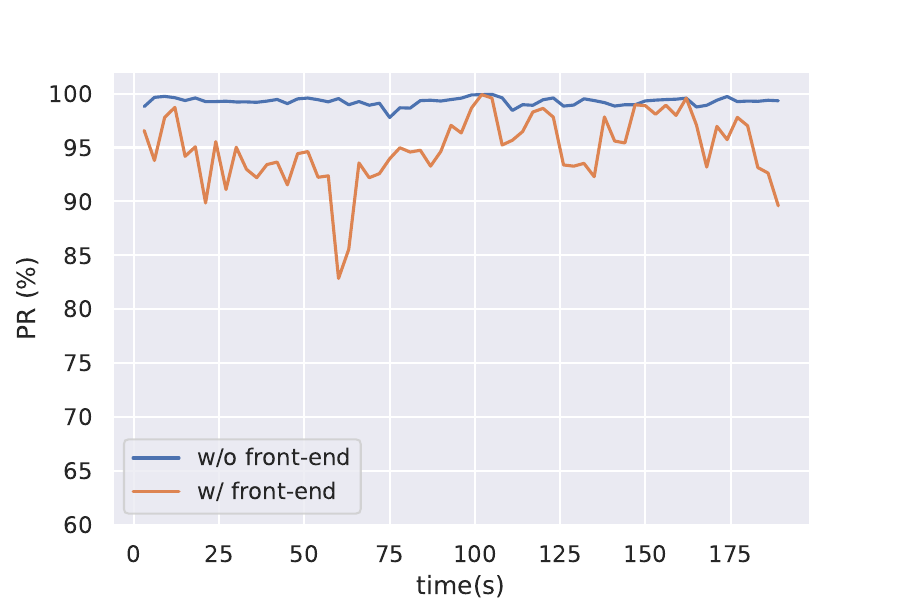}
\subcaption{PR of the back-end module\centering}
\label{fig:ablation_study_back_end_pr}
\end{minipage}
\begin{minipage}[t]{0.495\linewidth}
\centering
\includegraphics[trim={0 0 45 0},clip,width=1\linewidth]{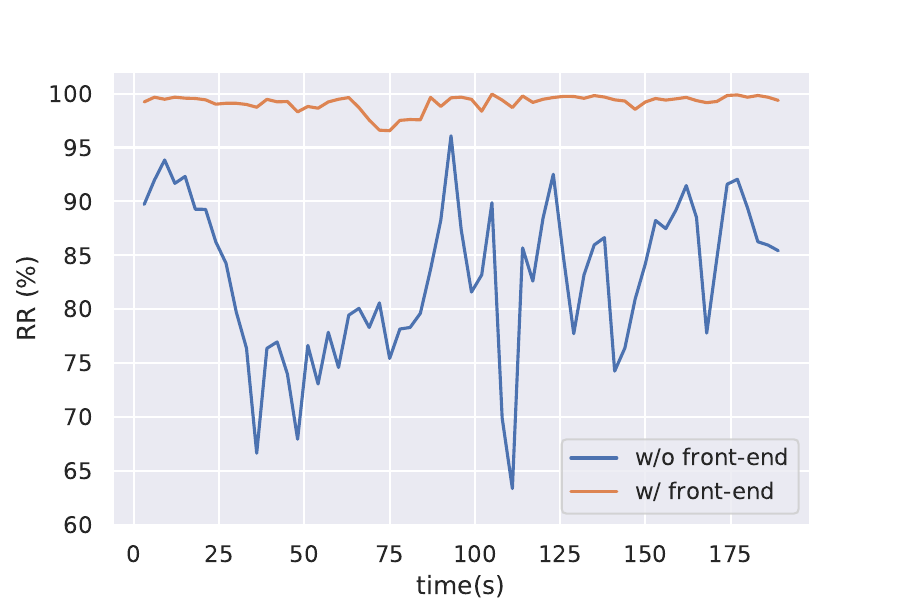}
\subcaption{RR of the back-end module\centering}
\label{fig:ablation_study_back_end_rr}
\end{minipage}
\caption{PR and RR of static submaps generated by back-end with and without front-end, plotted over time. PR and RR values are only computed for submaps within a \SI{6}{\meter} distance from the sensor at different times.}
\label{fig:ablation_study_back_end}
\end{figure}

\noindent{\textbf{Front-end and back-end coupling:}}
The importance of coupling the front-end and back-end is highlighted in \prettyref{tab:ablation_overall}. Achieving near 100\% RR by detecting all dynamic objects in each scan is difficult in highly dynamic scenarios. Simply stacking scans into a map can lead to the accumulation of dynamic points misclassified as static, resulting in inferior RR performance for \textit{front-end only}. On the other hand, \textit{back-end only}’s performance also degrades when there are many dynamic points in the scan. This is because many dynamic objects may be beyond the end of rays during the approximation of ray tracing, leading to the back-end’s incorrect preservation of dynamic points.

Our previous work DynamicFilter~\cite{fan2022dynamicfilter} also incorporates both front-end and back-end. However, unlike \textnameSMAT{}, DynamicFilter's front-end relies solely on visibility-based dynamic point removal and does not consider the inherent relationship between dynamic object tracking and mapping. As a result, while DynamicFilter outperforms \textit{front-end only} and \textit{back-end only}, there is still a significant performance gap compared to \textnameSMAT{}.

The background subtraction step in \textnameSMAT{} also plays an important role because it utilizes the prior map from back-end to improve detection performance in front-end. \textnameSMAT{} without background subtraction would have a weak interplay between the back-end and front-end, leading to a performance similar to DynamicFilter but lower than \textnameSMAT{}.

\begin{figure*}
\centering
\includegraphics[width=1.0\linewidth]{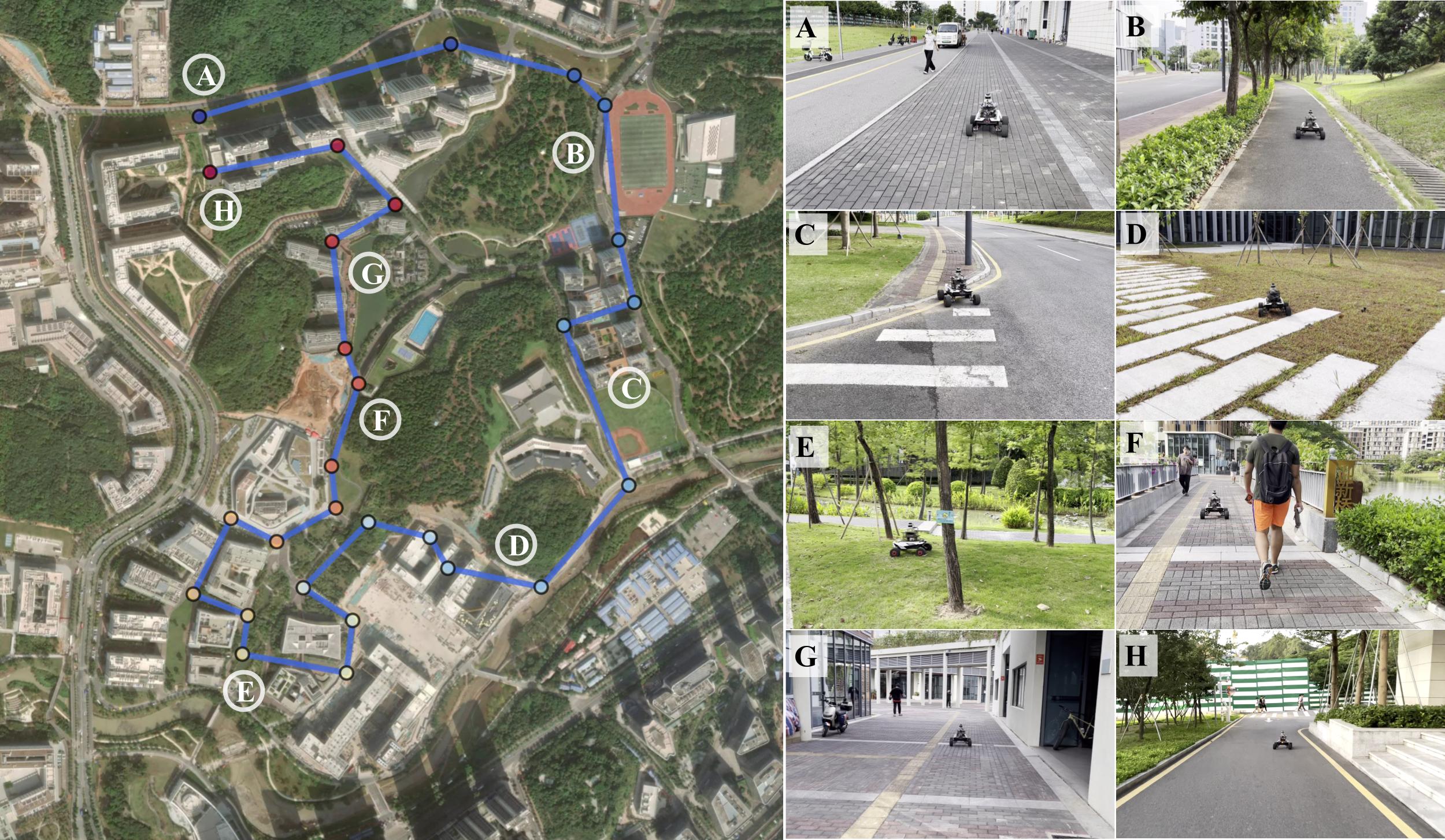}
\caption{Campus tour experiment. The map on the left shows the checkpoints in the visit order. The proposed long-range navigation system has been extensively tested in various scenarios, including unbounded urban environments like roads, sidewalks, and bicycle lanes (A-C, H), uneven grass regions (D-E), urban blocks (G), and narrow structures with pedestrians (F). With \textnameSMAT{}, the robot was able to visit the entire campus without human assistance and pre-built maps and only with imprecise GPS.} 
\label{fig:experiment-campus}
\end{figure*}

\noindent{\textbf{Self-reinforcing mechanism.}}
\textnameSMAT{} effectively eliminates nearly all dynamic points while preserving the spatial structure with its self-reinforcing mechanism, as shown in \prettyref{fig:ablation_study_front_end} and \prettyref{fig:ablation_study_back_end}, demonstrating the mutual reinforcement between the front-end and back-end.

As illustrated in \prettyref{fig:ablation_study_front_end_pr}, the utilization of the prior map generated by the back-end significantly improves the PR of the front-end at the current time step. The incorporation of a more accurate prior static map from the back-end would be beneficial to the preservation of more static features and reduction of type I error. Additionally, even when the front-end starts with poor PR, the back-end can quickly improve it in about 30 seconds and stabilize the front-end's PR afterward. However, as shown in \prettyref{fig:ablation_study_front_end_rr}, the RR of the static scan generated by the front-end remains consistently low, regardless of the presence of the back-end. This is because the back-end primarily helps the front-end reduce false detection of dynamic objects (type I error) and yield a high PR, but cannot directly address the missing detection of dynamic objects (type II error), which explains why RR remains low. Fortunately, the back-end has two additional checks, visibility checking and occupancy probability computation, to improve the identification and subsequent filtering of dynamic points, leading to a lower type II error. After these two checks, the RR of the back-end's output map can surpass 95\%, as shown in \prettyref{fig:ablation_study_back_end_rr}.

It is inevitable that the front-end will incorrectly filter out certain static points (type I error), leading to a slight negative impact on the back-end's PR, as shown in \prettyref{fig:ablation_study_back_end_pr}. Nonetheless, this slight negative impact remains within an acceptable range. According to the comparison between \textnameSMAT{} and \textit{back-end only} in \prettyref{tab:ablation_overall}, the inclusion of the front-end in \textnameSMAT{} only results in a 1.74\% reduction in the PR of the final static map generated by the back-end, while concurrently achieving a significantly improved RR of 99.61\%.

Thus, \textnameSMAT{}'s self-reinforcing mechanism leverages the inherent synergy between tracking and mapping. The back-end significantly enhances the front-end's ability to preserve static features, leading to substantial improvements in PR. Conversely, the front-end effectively assists the back-end in filtering dynamic points, resulting in enhanced RR. Thanks to both components, \textnameSMAT{} adeptly retains the spatial structure of the scene while efficiently filtering out the majority of dynamic points, thereby achieving exceptional performance in both tracking and mapping.

\section{Real-world experiments}
\label{sec:real_exp}

In this section, we present two real-world experiments conducted to evaluate the proposed \textnameSMAT{} framework on a mobile robot platform. The platform utilized a versatile four-wheeled mini chassis and an onboard computer with an AMD R7-5800H CPU for the real-time \textnameSMAT{} computation. The sensor suite included an Ouster OS1-32 3D LiDAR, a 9-axis inertial measurement unit (IMU), and a commodity U-Blox M8N GPS receiver. The system's onboard runtime performance is reported in \prettyref{tab:onboard_runtime}.

\begin{table}[]
\centering
\caption{Runtime of components.}
\label{tab:onboard_runtime}
\resizebox{0.48\textwidth}{!}{%
\begin{tabular}{lcc}
\hline
\rowcolor[HTML]{EFEFEF} 
\cellcolor[HTML]{EFEFEF}                             & \multicolumn{2}{c}{\cellcolor[HTML]{EFEFEF}Runtime / scan}                                                           \\
\rowcolor[HTML]{EFEFEF} 
\multirow{-2}{*}{\cellcolor[HTML]{EFEFEF}Components} & \multicolumn{1}{l}{\cellcolor[HTML]{EFEFEF}Mean {[}ms{]}} & \multicolumn{1}{l}{\cellcolor[HTML]{EFEFEF}Std {[}$\text{ms}${]}} \\ \hline
background subtraction in front-end                  & 2.77                                                      & 0.88                                                    \\
MOT system in front-end                              & 21.57                                                     & 9.50                                                     \\
back-end optimization                                & 12.61                                                     & 2.40                                                     \\ \hline
\end{tabular}
}
\end{table}

The objective of these real-world experiments was twofold: 1) to assess the capability of the \textnameSMAT{} framework in achieving robust perception in highly dynamic unknown urban environments, and 2) to determine if the online perception results could support long-range navigation without the need for pre-mapping. To achieve this, the mobile platform navigated through unknown environments (without pre-built maps) using only a series of GPS coordinates received from its commodity GPS receiver.

For a comprehensive and extensive evaluation, we selected two large-scale scenarios: a university campus and a seaside park, both characterized by the presence of pedestrians and other traffic agents. The robot platform successfully completed both tasks, covering a remarkable distance of over \SI{7}{km} during the longest run. These experiments provide strong evidence of \textnameSMAT{}'s effectiveness and robustness in enhancing the robot's perception in unknown, human-populated environments. More detailed information about the real-world experiments can be found on our project website: \url{https://sites.google.com/view/smat-nav}, and additional details about the long-range navigation system are provided in \prettyref{apd:long-range}.

\subsection{Campus tour}
\label{sec:campus-tour}

\begin{figure}[ht]
\centering 
\includegraphics[width=1\linewidth]{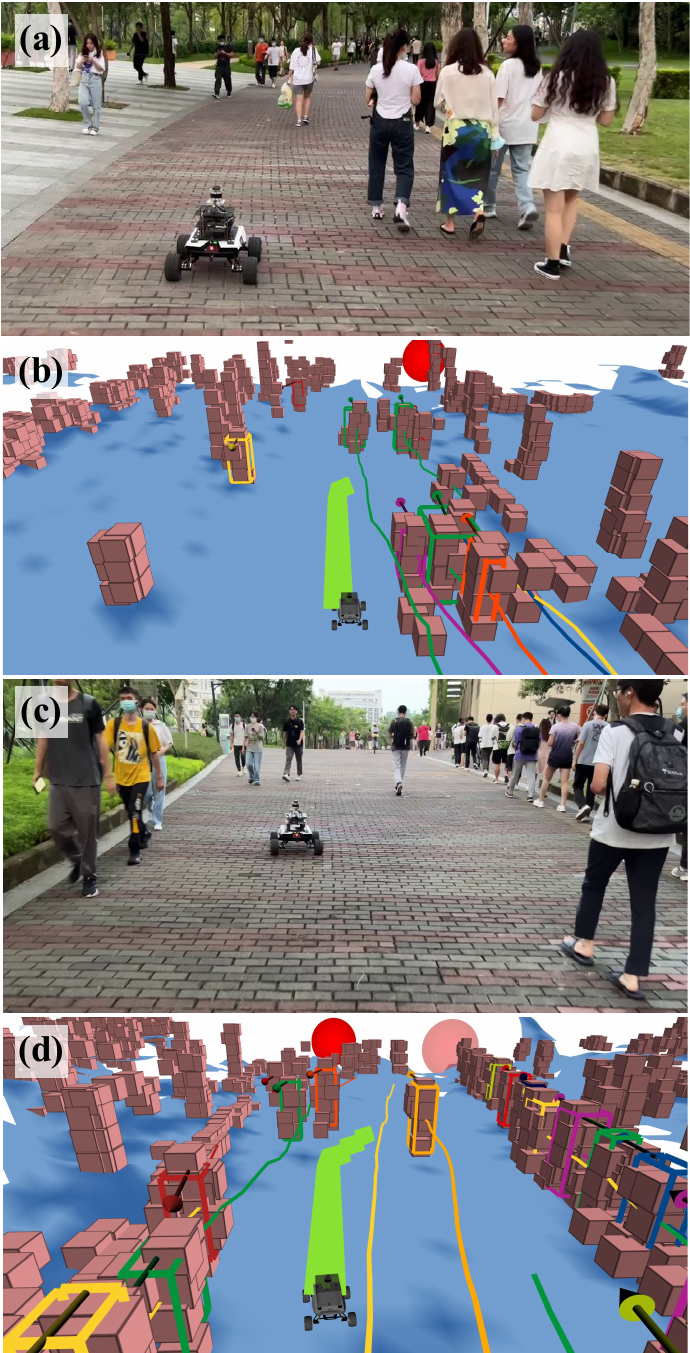}
\caption{Successful navigation through highly populated regions with \textnameSMAT{}. (a) and (c) show the actual situations, while (b) and (d) show \textnameSMAT{}'s online mapping results, based on which the robot can generate a drivable path (green) to reach the local goal position (red ball).}
\label{fig:pedestrains}
\end{figure}

\begin{figure*}
\centering
\includegraphics[width=1.0\linewidth]{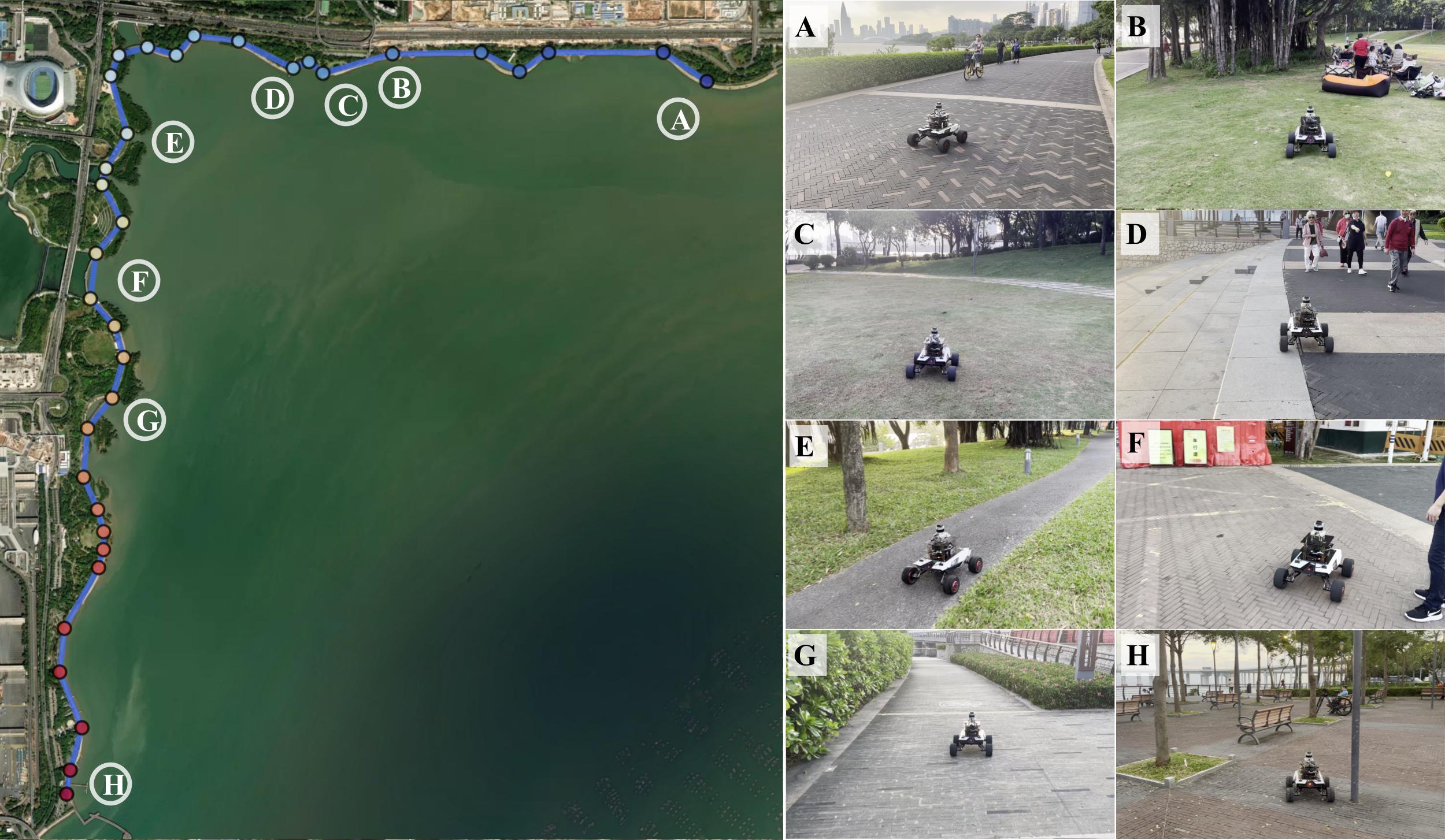}
\caption{Park tour along the coastline. GGuided by predefined checkpoints and relying on \textnameSMAT{}, the robot successfully traversed the coastline, encountering various scenarios including roads, sidewalks with pedestrians (A, D, E, H), and meadows (B, C). In the case where a temporary construction site blocked the regular sidewalk (F), the robot utilized \textnameSMAT{}'s local mapping to identify a bicycle lane as an alternative path (F-G) to continue its progress forward.
}
\label{fig:experiment-park}
\end{figure*}


The first experiment was conducted on a university campus, as shown in \prettyref{fig:experiment-campus}. The robot visited all prominent landmarks on the campus, such as a playground, residence halls, rivers, canteens, and libraries, by following imprecise GPS checkpoints. Throughout the experiment, it traveled about \SI{3.5}{km}, encountering diverse rigid and dynamic obstacles, and different terrain types. Notably, the robot \textit{solely} relied on imprecise GPS checkpoints and coarse GPS signals, without accessing a pre-built map, which posed a significant challenge to the robustness of online perception. Nevertheless, \textnameSMAT{} adeptly extracted spatial structures and concurrently tracked moving objects during the navigation. The autonomous campus tour was completed within approximately 45 minutes, with an average speed of \SI{1.7}{\meter/\second}, without requiring human intervention.

The campus tour began near a building (\prettyref{fig:experiment-campus}A), and the robot followed bicycle lanes, crossed rivers, and passed canteens, libraries, and residence halls. Finally, it returned to the same building via a different route (\prettyref{fig:experiment-campus}H). The robot traversed both structured and unstructured regions, avoiding obstacles such as bushes, ditches (\prettyref{fig:experiment-campus}B), and the riverbank (\prettyref{fig:experiment-campus}E). In cases where the robot deviated into a meadow between urban blocks due to rough checkpoint localization (\prettyref{fig:experiment-campus}D), it intelligently found the best frontier for exploration and eventually returned to the correct route. \textnameSMAT{} enabled online recovery of the drivable map, even in areas with high-density pedestrians, as shown in \prettyref{fig:pedestrains}. The extracted drivable map had high accuracy without ghost effects resulting from pedestrians, even on narrow bridges (\prettyref{fig:experiment-campus}F) and \prettyref{fig: bridge-passing}, allowing for safe and effective navigation without getting stuck or colliding. 

\begin{figure*}[ht]
\centering
\begin{minipage}[t]{0.33\linewidth}
\centering
\includegraphics[width=0.99\linewidth]{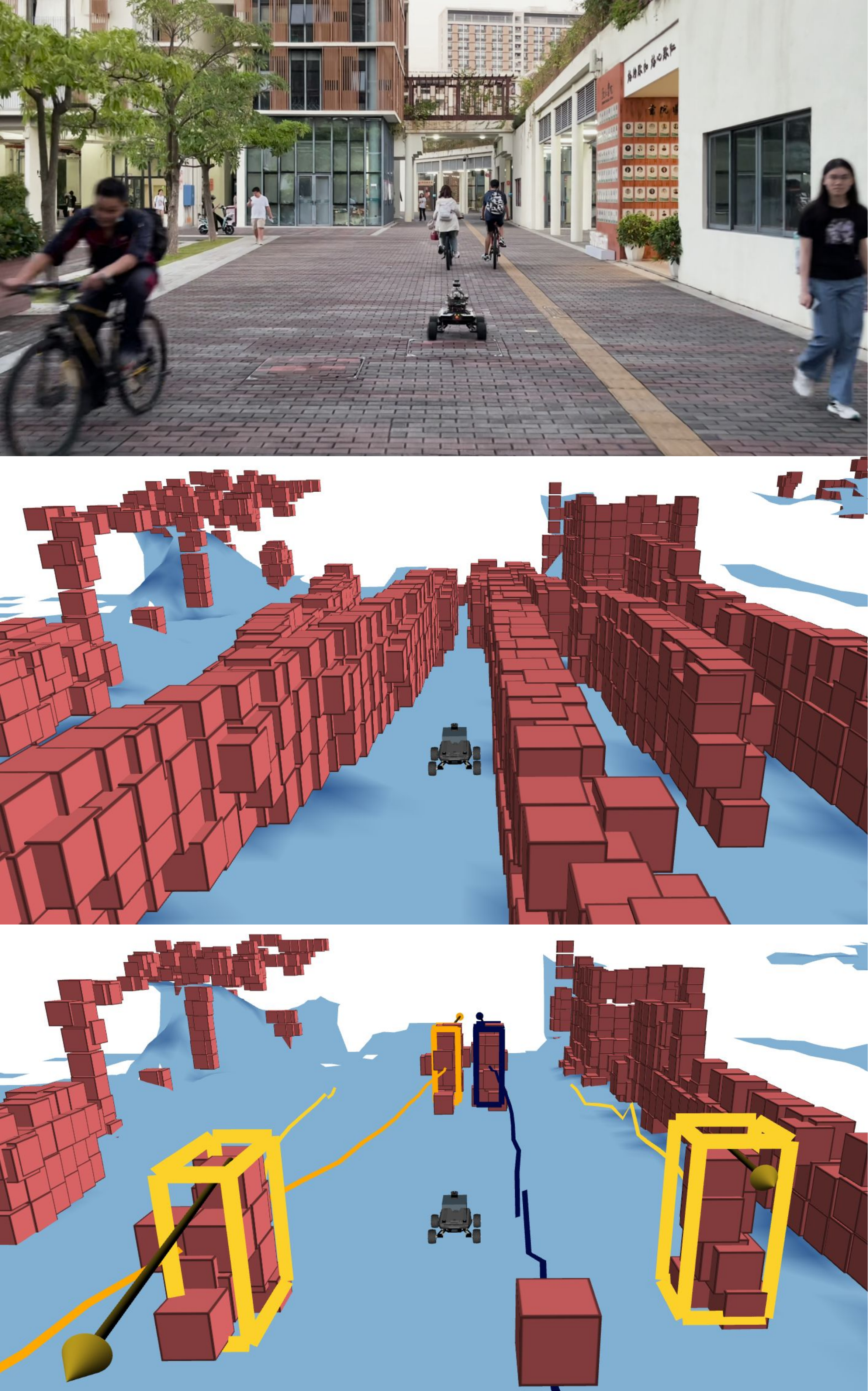}
\subcaption{\centering }
\end{minipage}
\begin{minipage}[t]{0.33\linewidth}
\centering
\includegraphics[width=0.99\linewidth]{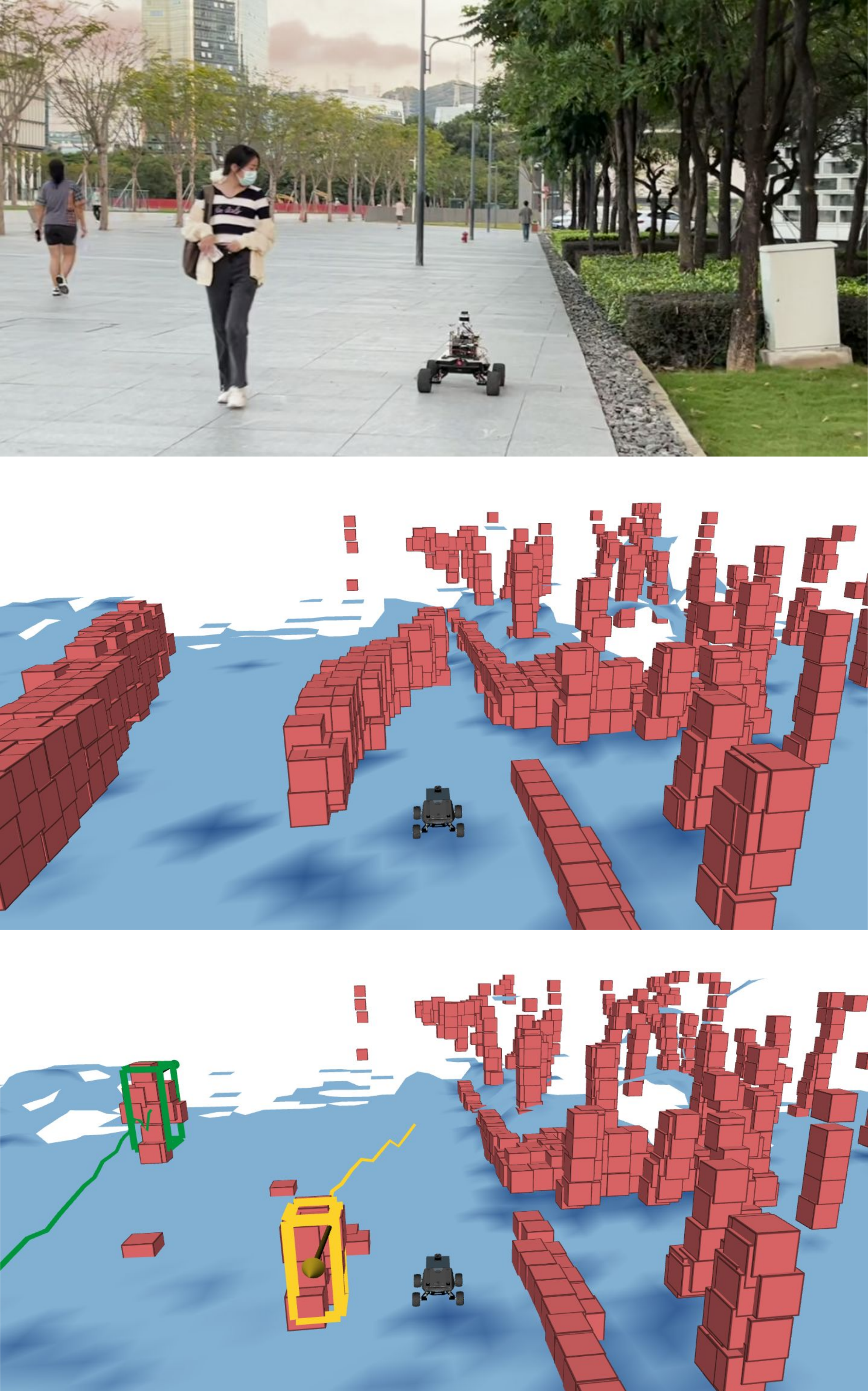}
\subcaption{\centering }
\end{minipage} 
\begin{minipage}[t]{0.33\linewidth}
\centering
\includegraphics[width=0.99\linewidth]{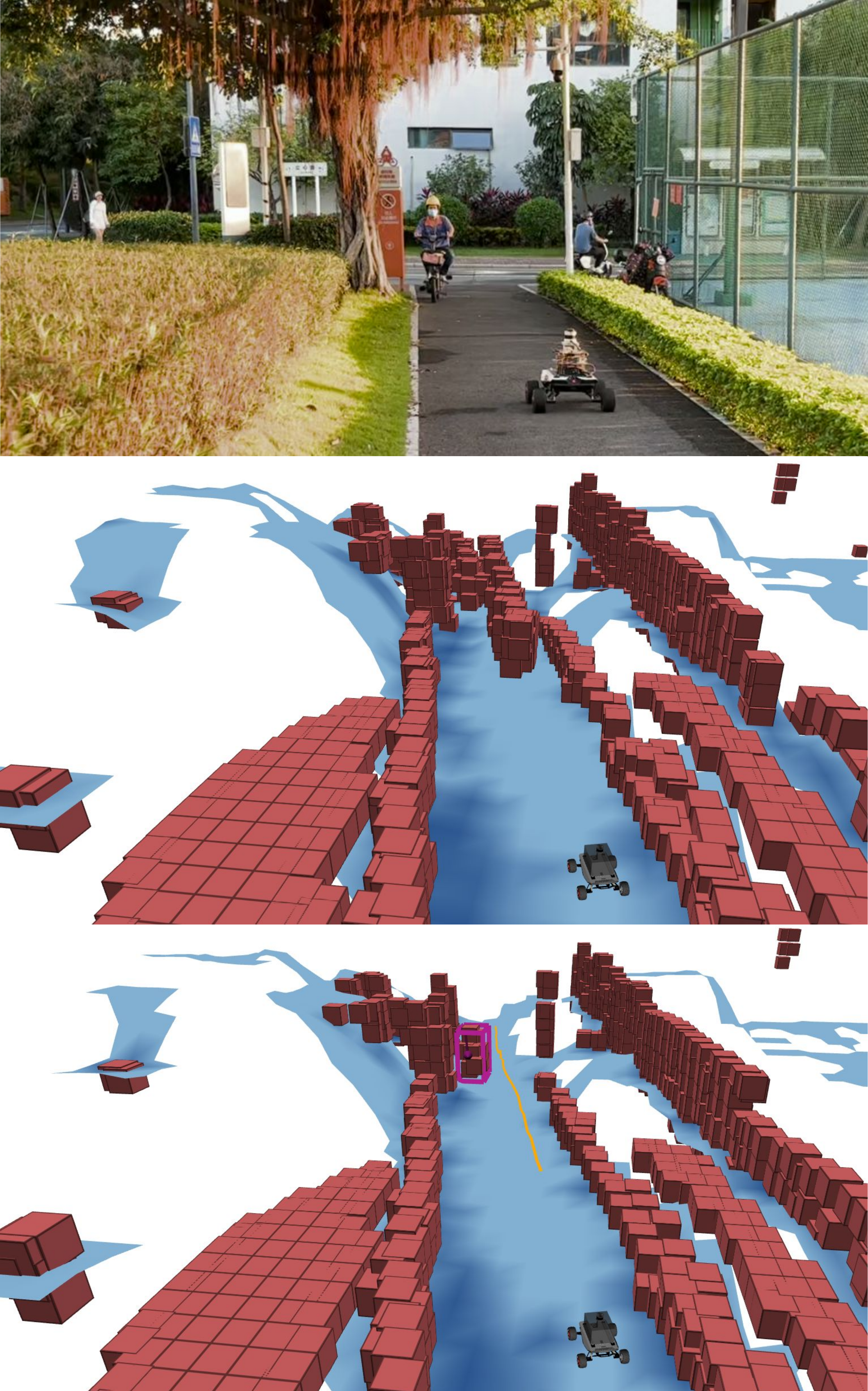}
\subcaption{\centering }
\end{minipage} \\
\begin{minipage}[t]{0.33\linewidth}
\centering
\includegraphics[width=0.99\linewidth]{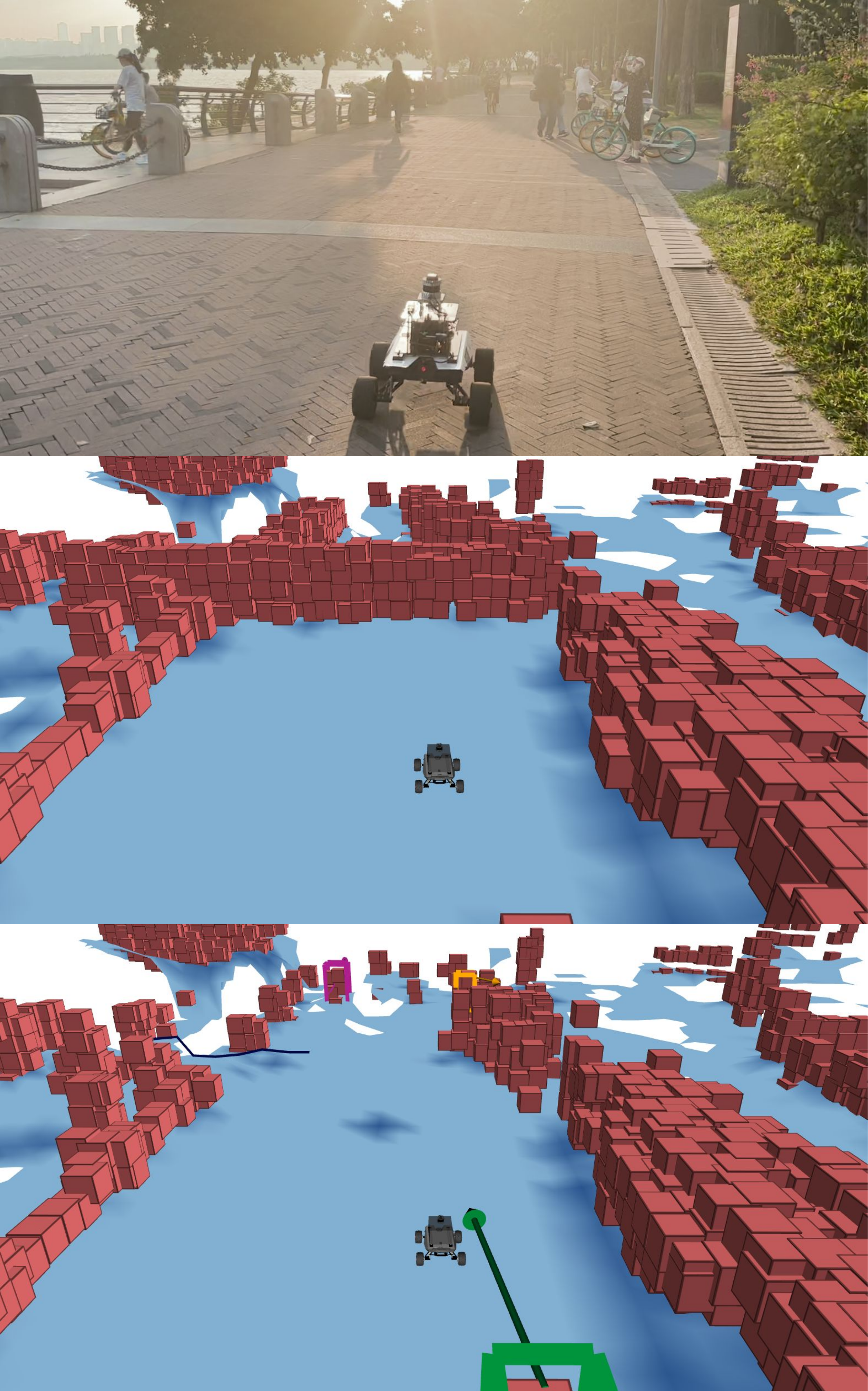}
\subcaption{\centering }
\end{minipage}
\begin{minipage}[t]{0.33\linewidth}
\centering
\includegraphics[width=0.99\linewidth]{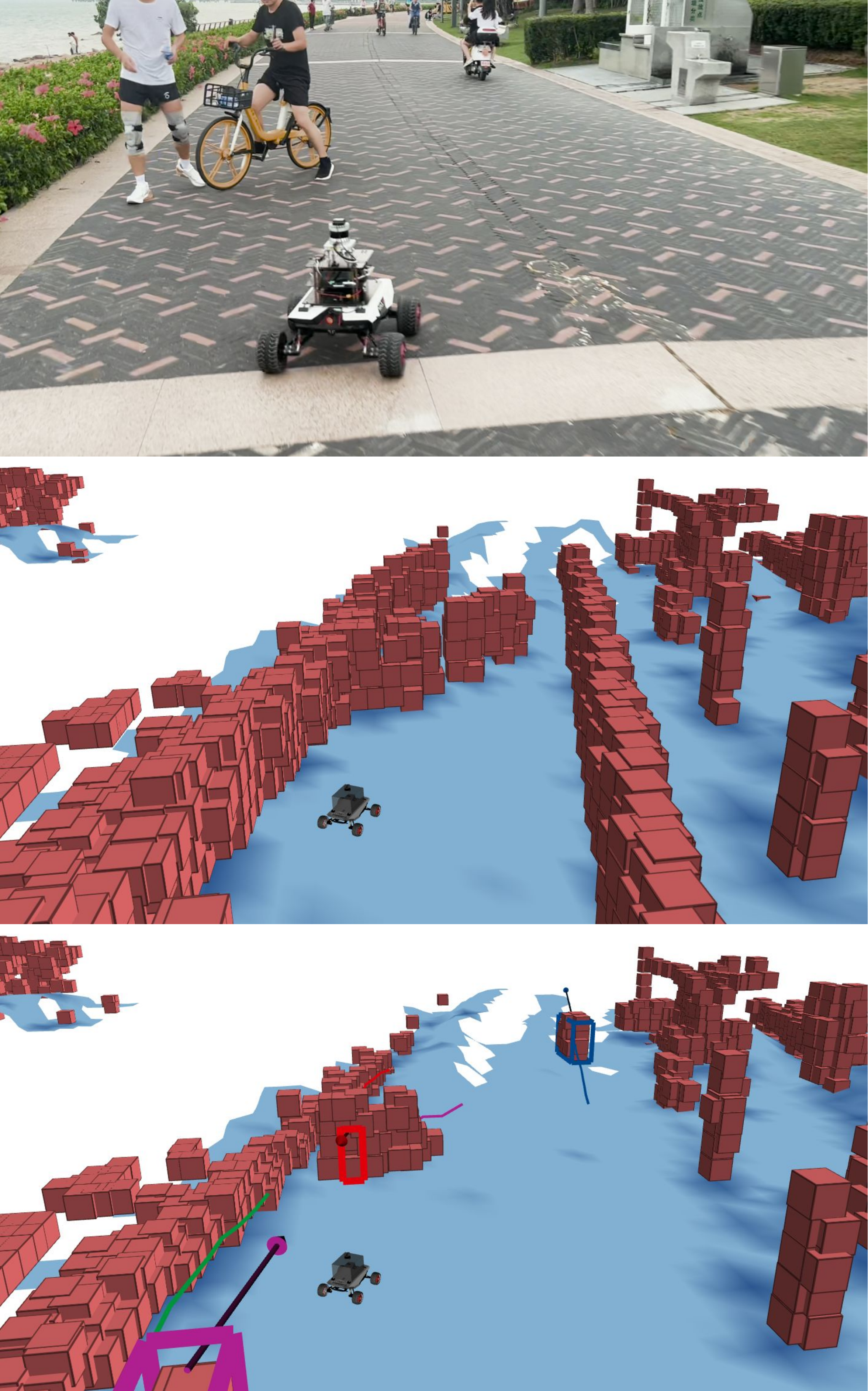}
\subcaption{\centering }
\end{minipage} 
\begin{minipage}[t]{0.33\linewidth}
\centering
\includegraphics[width=0.99\linewidth]{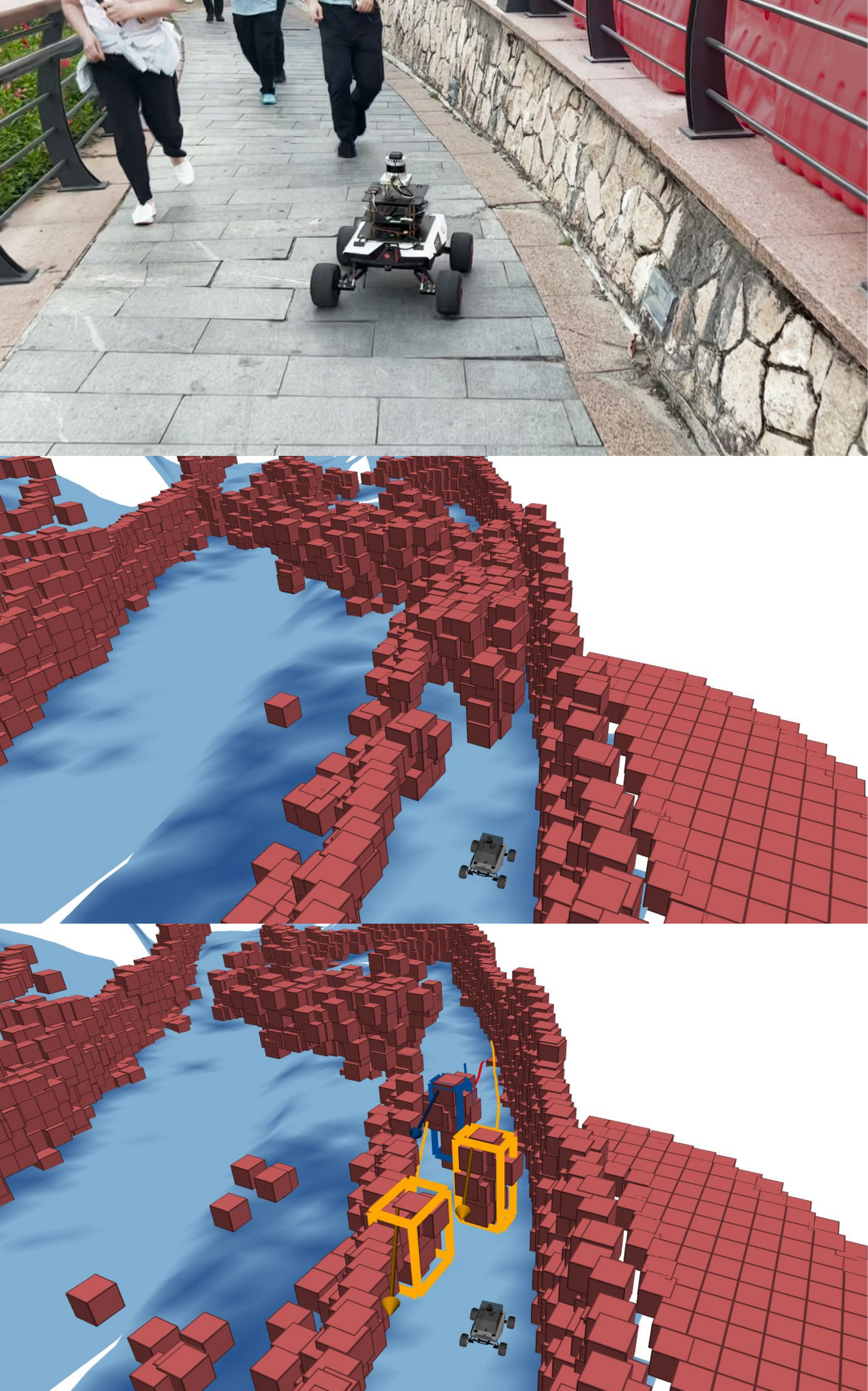}
\subcaption{\centering }
\end{minipage}
\caption{Qualitative results of \textnameSMAT{} in six snapshots of real-world urban scenarios. Each snapshot includes the actual scene (first row), the local mapping generated by a state-of-the-art SLAM \cite{xu2021fast} (second row), and local mapping by \textnameSMAT{} (third row). Inaccurate mapping results from the SLAM system can compromise the performance of downstream navigation tasks, such as frontier finding and local planning. In contrast, \textnameSMAT{} accurately preserves the environment's geometric outline while simultaneously tracking moving objects, resulting in superior performance.
}
\label{fig:traversability-performance}
\end{figure*}

\prettyref{fig:traversability-performance}a-c shows \textnameSMAT{} tracking and mapping details for three snapshots in the campus tour. In \prettyref{fig:traversability-performance}a and b, we observe that \textnameSMAT{} achieves real-time and accurate tracking of pedestrians and bicycles, while also constructing a high-quality static map. In contrast, the state-of-the-art SLAM method only produces a static map mixed with dynamic points, leading to the robot's inability to plan a safe trajectory. \prettyref{fig:traversability-performance}c demonstrates \textnameSMAT{} successfully computing the drivable area through a narrow corridor, effectively tracking and filtering out a motorbike ridden by an individual in blue. Although \textnameSMAT{} misses one person on the right side of the corridor due to the LiDAR's sparse perception of distant objects, this omission does not impact the robot's safe navigation.


\subsection{Park tour along coastline}
\label{sec:park-tour}

Another real-world experiment was conducted in an open park to further investigate \textnameSMAT{}'s mapping performance. Unlike the enclosed campus environment, this experiment aimed to verify whether \textnameSMAT{} could perform robustly in a more complex environment without a prior map. In this scenario, the robot continuously perceived traversable areas in real-time, including pedestrian paths, coastlines, and grassy regions, and interacted with cyclists, joggers, and pedestrians. Despite the increasing complexity of the environment, the robot equipped with \textnameSMAT{} achieved a long-distance travel of \SI{7}{km}, relying only on a sequence of imprecise GPS checkpoints as navigation guidance, as depicted in \prettyref{fig:experiment-park}.

During the 75-minute trip, our robot exhibited exceptional performance without any human intervention, demonstrating both flexibility and robustness in all challenging scenarios, even in the presence of sparse checkpoints and inaccurate GPS signals. \textnameSMAT{} accurately perceived traversable areas in most cases involving sidewalks and bicycle lanes (\prettyref{fig:experiment-park}A, D, and E), enabling the robot to navigate smoothly and steadily over long distances, unaffected by the presence of moving objects or potential obstacles. The system precisely identified static and dynamic obstacles depicted in \prettyref{fig:experiment-park}, with the robot deviating from sidewalks due to minor guidance errors. In \prettyref{fig:experiment-park}B-C, the rough checkpoint incorrectly guided the robot towards a camping area. The robot first perceived a straight path through the campsites (\prettyref{fig:experiment-park}B) and then returned to sidewalks (\prettyref{fig:experiment-park}C).

In a specific case depicted in \prettyref{fig:experiment-park}F, the regular sidewalk was obstructed by temporary construction sites, leading to a large deviation from the predefined checkpoints. This presented a considerable challenge for navigation systems relying on offline maps. In contrast, \textnameSMAT{} intelligently perceived potential forward routes without a pre-built map. The robot first moved toward the right side to assess the availability of a forward path. When the right side was blocked, \textnameSMAT{} dynamically detected the next best boundary on the left side of the construction sites, guiding the robot towards the bicycle lane and effectively bypassing the temporary construction area. Similarly, in \prettyref{fig:experiment-park}G, \textnameSMAT{} initially perceived a passable region ahead but encountered a dead end. Leveraging \textnameSMAT{}'s online perception, the robot planned a path back to the ramp area on the right side, successfully returning to the regular sidewalk. By relying on real-time mapping and perception rather than depending on offline maps, our system demonstrates excellent adaptability and responsiveness to dynamic environments, allowing the robot to identify alternative paths and make informed decisions on the fly, resulting in improved navigation performance. 


\prettyref{fig:traversability-performance}d-f shows \textnameSMAT{} tracking and mapping details in three park tour snapshots. In \prettyref{fig:traversability-performance}d and e, \textnameSMAT{} identifies stationary pedestrians or bicycles as static obstacles, such as the bicycles that stopped for photos (\prettyref{fig:traversability-performance}d) and pedestrians engaged in conversation (\prettyref{fig:traversability-performance}e). Fortunately, these errors do not impact navigation safety as the movable objects behave as static objects. \prettyref{fig:traversability-performance}f showcases \textnameSMAT{} performing well in a challenging situation where the robot encounters pedestrians approaching from the opposite direction in a narrow aisle. In contrast, the state-of-the-art SLAM method fails to provide a satisfactory static map in all these situations.

\subsection{\textnameSMAT{}'s other advantages}

\textnameSMAT{} enables the robot to navigate efficiently and safely in unknown, highly dynamic, and large-scale scenarios without relying on a pre-built global map. Deploying and maintaining such a map for these scenarios presents several challenges. First, there is the issue of memory consumption. In the scenarios we tested, the global map required memory ranging from hundreds of megabytes (for campus tour in \prettyref{sec:campus-tour}) to several gigabytes (for park tour in \prettyref{sec:park-tour}). Second, pre-built maps need continuous update to avoid inconsistencies between the map description and online perception. For example, a temporary construction site that appears in our park tour (\prettyref{fig:experiment-park}F) but is not available in the pre-built map could confuse or hinder a robot relying on a global map. In contrast, \textnameSMAT{}'s online mapping can instruct the robot to check all feasible paths and adaptively select an alternative solution.

\textnameSMAT{}'s memory and computational scalability are particularly crucial for long-range navigation and exploration in large urban regions. Compared to notable previous works in long-range navigation \cite{kummerle2015autonomous} and large-scale exploration \cite{cao2022exploration, cao2021tare}, \textnameSMAT{} offers advantages in terms of map independence and the scale of experimental sites. \cite{kummerle2015autonomous} evaluated their approach on a \SI{7.4}{km} through densely populated urban zones in Freiburg, Germany. Our park tour covered a similar distance of about $\SI{7}{km}$, but \textnameSMAT{} avoids heavy reliance on pre-built maps as used in \cite{kummerle2015autonomous}. In~\cite{cao2022exploration}, the sizes of the four experimental sites ranged from $\SI{130}{\metre} \times \SI{100}{\metre}$ to $\SI{340}{\metre} \times \SI{340}{\metre}$, with a maximum traveling distance of approximately \SI{1.8}{km} for the robot. In comparison, our campus tour covered the entire $\SI{1000}{\metre} \times \SI{700}{\metre}$ campus, with a traveling distance of around \SI{3.5}{km}. Our experiments are more complex than those conducted in \cite{cao2022exploration} and \cite{cao2021tare} in terms of the types and density of dynamic objects and unstructured obstacles, demonstrating the robustness and scalability of \textnameSMAT{}.

\textnameSMAT{} offers important advantages over learning-based tracking or perception methods, making it more suitable for mobile robot tasks. First, learning-based methods require a significant amount of training data to perform well, and transferring a model learned for one region to another can be challenging due to differences in data distribution for dynamic objects, unstructured obstacles, and their motion/appearances in difficult regions such as campuses, CBDs, downtown areas, and uptowns of different cities. Collecting real-world data for long-range navigation tasks in diversified unknown urban environments can be expensive or difficult. In contrast, \textnameSMAT{} does not rely on training data and can be deployed anywhere in a plug-and-play manner. Second, \textnameSMAT{} allows the robot to continuously and autonomously collect data for training learning-based systems, enabling \textit{cold start} capabilities that are useful for continuous learning or recovering from errors or mistakes. Additionally, \textnameSMAT{} can be deployed on a single CPU with real-time performance, while learning-based methods typically depend on resource-intensive GPUs that may not be available or energy-efficient for onboard computers. Moreover, \textnameSMAT{} solely utilizes LiDAR perception, which better safeguards pedestrian privacy, whereas many learning-based methods require image modality, which can be restrictive in security or government applications. Lastly, \textnameSMAT{} already achieves state-of-the-art dynamic object tracking performance that surpasses certain learning-based methods that leverage both LiDAR and images (see \prettyref{tab:jrdb_results}). Learning-based methods complement \textnameSMAT{} and can be integrated into \textnameSMAT{} to further enhance performance.

\section{Conclusion}
\label{sec:conclusion}

This paper introduces \textnameSMAT{}, a simultaneous mapping and tracking framework. \textnameSMAT{} effectively utilizes the reciprocal relationship between dynamic and static LiDAR points and static structural mapping to enhance performance in challenging tasks such as tracking multiple moving objects and mapping in highly dynamic urban scenarios. Through tests on various datasets, simulations, and physical robots, \textnameSMAT{} demonstrates state-of-the-art performance in both tasks. It enables real-time and robust mapping in unknown, dynamic, and large-scale urban scenarios without relying on a prior global map, expensive GPU resources, or image modalities that may compromise privacy. Future work could focus on enhancing \textnameSMAT{} with learning-based scene understanding, other sensory modalities, and semantic-aware mapping using large foundation models.

\section*{Appendix}

\appendix
\renewcommand{\appendixname}{Appendix~\Alph{section}}

\section{\textnameSMAT{}'s hyperparameters}

The hyperparameter values used in all of our experiments are listed in \prettyref{tab:parameters}. For a robot moving at low speeds ($v < \SI{3.0}{\metre / \second}$), a neighborhood of up to \SI{20}{\metre} is sufficient for perception. Therefore, we set $R_{\text{bound}}$ to \SI{20}{\metre} to process only the points within \SI{20}{\metre} of the robot's current position for each LiDAR scan. 
To determine stable tracking metrics, since LiDAR's frequency is \SI{10}{Hz}, \textnameSMAT{} can detect and track target objects in 10 consecutive scans in \SI{1}{\second}, which is sufficiently long to determine objects are being tracked stably or not. Thus, we set $t_{\text{val}}$ to \SI{1}{\second}. We consider an object stably tracked if it is detectable and trackable most of the time, has a certain speed, and does not vary significantly in appearance or volume. Accordingly, we set $\rho_{\min}$ to 0.7, $v_{\min}$ to \SI{1.0}{\metre / \second}, and $\Delta \mathbb{V}_{\min}$ to \SI{3}{\metre ^3}. 
When multiple static scans generate a local static map, only scans close to the robot's current location can provide useful information due to sparser LiDAR data at longer distances. Hence, we set $R_{\text{localmap}}$ to \SI{5}{\metre}. For the occupancy probability threshold $p_{\text{occ}}$ in \prettyref{alg:visibility_check}, we set it to a medium value $0.5$. The higher threshold will result in more conservative static maps.

\begin{table}[ht]
\caption{\textnameSMAT{}'s hyper-parameters used in \prettyref{sec:perc_method}.}
 \begin{tabularx}{0.49\textwidth}{l|X}
   \hline
  Parameter & Value  \\
   \hline
   \text{$R_{\text{bound}}$} & \SI{20}{\metre} \\
   \text{$t_{\text{val}}$} & \SI{1}{\second} \\
   \text{$\rho_{\min}$} & 0.7 \\
   \text{$v_{\min}$} & \SI{1.0}{\metre / \second} \\
   \text{$\Delta \mathbb{V}_{\min}$} & \SI{3}{\metre ^3} \\
   \text{$R_{\text{localmap}}$} & \SI{5}{\metre} \\ 
   \text{$p_{\text{occ}}$} & 0.5 \\
   \hline
 \end{tabularx}

\label{tab:parameters}
\end{table}

\section{Multi-object tracking (MOT) metrics}
\label{apd:mot_metrics}


Here we briefly describe the MOT metrics used in \prettyref{sec:mot_eval}: MOTA~\cite{bernadin2008Evaluating}, IDF1~\cite{ristani2016performance}, and HOTA~\cite{luiten2021hota}. They assess the tracking results by comparing the ground truth and predicted trajectories. 
Ground truth trajectories (gtTrajs) are represented by a set of detections (gtDets) in each frame. Each gtDet is assigned a unique id (gtID). The gtIDs remain consistent over time for detections from the same ground truth object and are unique within each frame. 
Predicted trajectories (prTrajs) are similar to the ground truth data, consisting of a set of predicted detections (prDets) with unique predicted ids (prIDs). Similar to gtIDs, prIDs are unique within each frame and remain consistent over time for detections from the same predicted object.

MOTA~\cite{bernadin2008Evaluating} calculates the number of true positives (TP), false negatives (FN), and false positives (FP) between the sets of gtDets and prDets. After a bijective mapping, pairs consisting of a prDet and its corresponding gtDet that are similar enough are considered TPs. IoU, which measures the ratio between the intersection area of two bounding boxes and the area of their union, is commonly used to assess similarity between detection results in terms of bounding boxes. Any gtDets that are not TPs are classified as FNs, while any prDets that are not TPs are classified as FPs.
To quantify association errors, MOTA introduces IDSW, which is a TP whose prID differs from the prID of the previous TP with the same gtID. MOTA is calculated as:
\begin{equation*}
\text{MOTA} = 1 - \frac{|\text{FN}|+|\text{FP}|+|\text{IDSW}|}{|\text{gtDet}|}.
\end{equation*}
IDSW only measures association errors w.r.t. the previous TP with the same gtID and does not account for errors where the same prID switches to a different gtID \cite{luiten2021hota}. Thus, $|\text{IDSW}|$ is typically smaller than $|\text{FN}|$ and $|\text{FP}|$, and MOTA primarily reflects the detection performance.

IDF1 \cite{ristani2016performance} uses IDTP, IDFN, and IDFP instead of TP, FN, and FP in MOTA. IDTP, IDFN, and IDFP are calculated between gtTrajs and prTrajs. For a pair of a gtDet and a prDet in TPs, if their gtID and prID correspond to the same trajectory, this pair is an IDTP. Compared to the definition of TP, the definition of IDTP is stricter and emphasizes tracking continuity. If a ground truth trajectory is tracked into several predicted trajectories, only the pairs of prDets in the best predicted trajectory and their mapped gtDets are considered as IDTPs. Any gtDets that are not IDTPs are IDFNs, and any prDets that are not IDTPs are IDFPs, similar to the definition of FN and FP. IDF1 is calculated as:
\begin{equation*}
\text{IDF1} = \cfrac{|\text{IDTP}|}{|\text{IDTP}| + 0.5 |\text{IDFN}| + 0.5 |\text{IDFP}|}.
\end{equation*}
IDF1 only considers the best set of matching trajectories for IDTPs. Any predicted trajectory that is not in this set is counted as IDFP and reduces IDF1, even if it contributes correct detections. Thus, IDF1 is mainly indicate the association performance.

HOTA \cite{luiten2021hota} employs the same approach as MOTA to calculate TP, FN, and FP. However, instead of calculating these metrics using a fixed similarity threshold, HOTA calculates them at each valid similarity threshold $\alpha$. For each $\alpha$, HOTA computes $\text{HOTA}_{\alpha}$ using the corresponding TP, FN, and FP values. The final HOTA score is obtained by integrating the $\text{HOTA}_{\alpha}$ values across the range of $\alpha$ from 0 to 1.

$\text{HOTA}_{\alpha}$ improves upon MOTA by introducing a better measure of association error. It can be divided into a detection accuracy score, $\text{DetA}_{\alpha}$, and an association accuracy score, $\text{AssA}_{\alpha}$. Specifically, $\text{DetA}_{\alpha} = \cfrac{|\text{TP}|}{|\text{TP}| + |\text{FN}| + |\text{FP}|}$, while $\text{AssA}_{\alpha} = \cfrac{1}{|\text{TP}|}\sum_{c \in \text{TP}}\mathcal{A}(c)$, where $\mathcal{A}(c)$ evaluates the association performance of each prDet in the set of TPs and details for its calculations can be found in \cite{luiten2021hota}. The final DetA, AssA, and HOTA scores are:
\begin{equation*}
\begin{split}
&{\text{DetA}} = \int_0^1{\text{DetA}_{\alpha}}d\alpha, \\
&{\text{AssA}} = \int_0^1{\text{AssA}_{\alpha}}d\alpha, \\
&{\text{HOTA}} = \int_0^1{\text{HOTA}_{\alpha}}d\alpha, {\text{HOTA}_{\alpha}} = \sqrt{{\text{DetA}_{\alpha} \cdot \text{AssA}_{\alpha}}}.
\end{split}
\end{equation*}

\begin{figure}[t]
\centering
\includegraphics[width=1.0\linewidth]{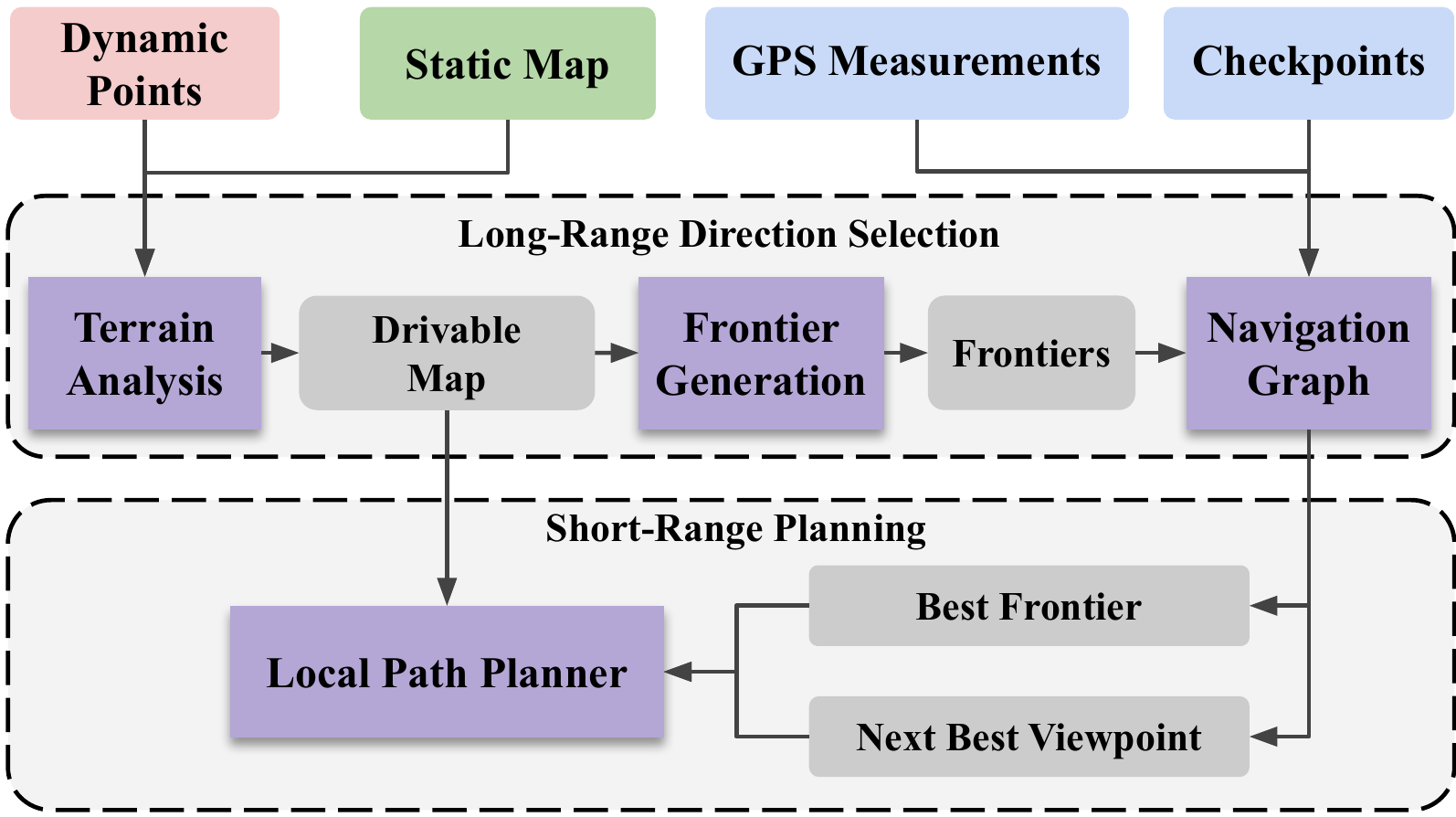}
\caption{Architecture of the long-range navigation system. First, the long-range direction selection module receives GPS guidance and perception information from \textnameSMAT{}. It analyzes the terrain quality, generates frontiers, and constructs the navigation graph. Its output includes a drivable map, the next best viewpoint, and the best frontier. These inputs are then passed on to the short-range planning module, which generates paths to control the robot.}
\label{fig:concept-navigation}
\end{figure}

\section{Long-Range Navigation System}
\label{apd:long-range}

\begin{figure}[t]
\centering
\includegraphics[width=1.0\linewidth]{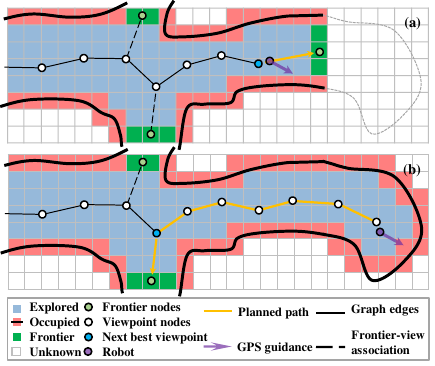}
\caption{The configuration of the navigation graph and the selection of the next best viewpoint and frontier node with the given reference direction. In case (a), the robot can simply select the nearest viewpoint node and the front frontier to drive forward. However, in case (b) when faced with a dead end, the robot needs to approach the next best viewpoint first and then drives to the corresponding best frontier later.}
\label{fig:navigation-graph}
\end{figure}

Here we present the long-range navigation system based on our \textnameSMAT{} perception. To estimate the local motion of the robot, we utilize the LiDAR SLAM algorithm \cite{xu2021fast}. Note that, the SLAM estimated pose may experience drift in global coordination during long-range navigation. However, its precision is sufficient for local mapping. For navigation guidance, we utilize the global information provided by GPS, which offers a rough estimate.

To accomplish the long-range navigation task with rough guidance, we propose a hierarchical framework that includes long-range direction selection and short-range path planning. The long-range direction selection utilizes information from \textnameSMAT{} to generate drivable maps and frontiers. We have developed a frontier-based algorithm for long-range direction selection, which is guided by imprecise checkpoints and rough GPS signals. To store and evaluate frontiers, we employ a navigation graph, allowing the robot to intelligently select the driving direction and backtrack when deviating from the correct path. For real-time collision avoidance with static and dynamic objects, we use a state lattice motion planner for short-range path planning. The architecture of the navigation system is illustrated in \prettyref{fig:concept-navigation}.

\noindent\textbf{Long-range direction selection} comprises three modules: terrain analysis, frontier generation, and navigation graph. Terrain analysis generates a drivable map by utilizing the current scan's dynamic points and the static map provided by \textnameSMAT{}. Both static and dynamic points are merged and organized into 2D grid cells. For each cell, the reference height is determined as the lower quartile of the height values of all points within it. The traversing cost of each point in a cell is then calculated as the difference between its height value and the reference height of the cell. In frontier generation, a probabilistic occupancy mapping algorithm~\cite{thrun2002probabilistic} is employed to incrementally separate the known and unknown regions in the drivable map. The extracted frontiers correspond to the known cells located at the boundary between the known and unknown regions.

The navigation graph stores and evaluates the extracted frontiers, consisting of two types of nodes: viewpoint nodes and frontier nodes. \prettyref{fig:navigation-graph} illustrates that viewpoint nodes are incrementally sampled along the robot's historical path at a predefined distance. When a new viewpoint node is sampled, it connects to its nearest viewpoint neighbor. Simultaneously, frontier nodes, which are centroids of frontier clusters, connect to their nearest viewpoint neighbor. Once the graph is constructed, we evaluate the priority scores of both frontier and viewpoint nodes. The computation of priority scores for frontier nodes considers both the reference and driving directions. The reference direction is the normalized vector from the current GPS location to the next checkpoint. The driving direction of a frontier is the normalized vector from its connected viewpoint to itself. We calculate the inner product of the two normalized direction vectors and transform the result into a score ranging from 0 to 1. Each viewpoint node's priority score is initialized with the maximum score among its frontier neighbors. Then, the max neighborhood aggregation operation is repeatedly applied among viewpoint nodes and corresponding edges until convergence. The aggregation decision is made as follows:
\begin{equation*}
    s_i^{*} = \max\big(s_i, \gamma\cdot\max_{j\in\mathcal{N}_i}(s_j)\big) \ ,
\end{equation*}
where $s_i$ and $s_j$ are the priority scores of viewpoint nodes $i$ and $j$, respectively. $s_i^*$ represents the priority score after an aggregation iteration. $\mathcal{N}_i$ represents the neighbor viewpoint nodes of node $i$. $\gamma$ is the discounted factor that controls the navigation behavior, maintaining a balance between driving forward and backtracking. Once convergence is achieved, we select the nearest viewpoint with the highest local priority score from those in the graph as the next best viewpoint. The best frontier is determined as the neighbor of the best viewpoint with the highest priority score.

\noindent \textbf{Short-range planning} involves creating a feasible path from the robot's current position, passing through the best viewpoint node, and ending at the best frontier node, once the next best viewpoint and best frontier have been identified. To determine a short-range goal on the planned path, a sliding window approach is utilized. In order to ensure a collision-free route to the short-range goal, a lattice sampling method~\cite{Zhang2020Falco} is employed. To forecast the future trajectory of each tracklet in the upcoming second, we utilize the extended Kalman filter with the constant velocity model. This enables the transformation of the trajectory into an unintrusive social space, thereby improving the robot's collision avoidance efficiency and enhancing the social compliance of motion planning.

\bibliographystyle{SageH}
\bibliography{references.bib}

\begin{thebibliography}{53}
\providecommand{\natexlab}[1]{#1}
\providecommand{\url}[1]{\texttt{#1}}
\providecommand{\urlprefix}{URL }
\expandafter\ifx\csname urlstyle\endcsname\relax
  \providecommand{\doi}[1]{DOI:\discretionary{}{}{}#1}\else
  \providecommand{\doi}{DOI:\discretionary{}{}{}\begingroup
  \urlstyle{rm}\Url}\fi

\bibitem[{Azim and Aycard(2012)}]{azim2012detection}
Azim A and Aycard O (2012) Detection, classification and tracking of moving
  objects in a 3d environment.
\newblock In: \emph{IEEE Intelligent Vehicles Symposium}. IEEE, pp. 802--807.

\bibitem[{Behley et~al.(2019)Behley, Garbade, Milioto, Quenzel, Behnke,
  Stachniss and Gall}]{behley2019semantickitti}
Behley J, Garbade M, Milioto A, Quenzel J, Behnke S, Stachniss C and Gall J
  (2019) Semantickitti: A dataset for semantic scene understanding of lidar
  sequences.
\newblock In: \emph{IEEE/CVF International Conference on Computer Vision}. pp.
  9297--9307.

\bibitem[{Bernardin and Stiefelhagen(2008)}]{bernadin2008Evaluating}
Bernardin K and Stiefelhagen R (2008) Evaluating multiple object tracking
  performance: The clear mot metrics.
\newblock \emph{EURASIP Journal on Image and Video Processing} 2008.
\newblock \doi{10.1155/2008/246309}.

\bibitem[{Bogoslavskyi and Stachniss(2016)}]{bogoslavskyi2016fast}
Bogoslavskyi I and Stachniss C (2016) Fast range image-based segmentation of
  sparse 3d laser scans for online operation.
\newblock In: \emph{IEEE/RSJ International Conference on Intelligent Robots and
  Systems}. IEEE, pp. 163--169.

\bibitem[{Cao et~al.(2021)Cao, Zhu, Choset and Zhang}]{cao2021tare}
Cao C, Zhu H, Choset H and Zhang J (2021) Tare: A hierarchical framework for
  efficiently exploring complex 3d environments.
\newblock In: \emph{Robotics: Science and Systems}, volume~5.

\bibitem[{Cao et~al.(2022)Cao, Zhu, Yang, Xia, Choset, Oh and
  Zhang}]{cao2022exploration}
Cao C, Zhu H, Yang F, Xia Y, Choset H, Oh J and Zhang J (2022) Autonomous
  exploration development environment and the planning algorithms.
\newblock In: \emph{IEEE International Conference on Robotics and Automation}.
  pp. 8921--8928.
\newblock \doi{10.1109/ICRA46639.2022.9812330}.

\bibitem[{Chen et~al.(2021)Chen, Li, Mersch, Wiesmann, Gall, Behley and
  Stachniss}]{chen2021lidar-mos}
Chen X, Li S, Mersch B, Wiesmann L, Gall J, Behley J and Stachniss C (2021)
  Moving object segmentation in 3d lidar data: A learning-based approach
  exploiting sequential data.
\newblock \emph{IEEE Robotics and Automation Letters} 6(4): 6529--6536.

\bibitem[{Chen et~al.(2022)Chen, Mersch, Nunes, Marcuzzi, Vizzo, Behley and
  Stachniss}]{chen2022automatic}
Chen X, Mersch B, Nunes L, Marcuzzi R, Vizzo I, Behley J and Stachniss C (2022)
  Automatic labeling to generate training data for online lidar-based moving
  object segmentation.
\newblock \emph{IEEE Robotics and Automation Letters} 7(3): 6107--6114.

\bibitem[{Cortinhal et~al.(2020)Cortinhal, Tzelepis and
  Aksoy}]{cortinhal2020salsanext}
Cortinhal T, Tzelepis G and Aksoy EE (2020) Salsanext: fast, uncertainty-aware
  semantic segmentation of lidar point clouds for autonomous driving.
\newblock \emph{arXiv preprint arXiv:2003.03653} .

\bibitem[{Curtis et~al.(2016)Curtis, Best and Manocha}]{Menge}
Curtis S, Best A and Manocha D (2016) Menge: A modular framework for simulating
  crowd movement.
\newblock \emph{Collective Dynamics} 1: 1--40.

\bibitem[{Dewan et~al.(2016{\natexlab{a}})Dewan, Caselitz, Tipaldi and
  Burgard}]{dewan2016motion}
Dewan A, Caselitz T, Tipaldi GD and Burgard W (2016{\natexlab{a}}) Motion-based
  detection and tracking in 3d lidar scans.
\newblock In: \emph{IEEE International Conference on Robotics and Automation}.
  IEEE, pp. 4508--4513.

\bibitem[{Dewan et~al.(2016{\natexlab{b}})Dewan, Caselitz, Tipaldi and
  Burgard}]{dewan2016rigid}
Dewan A, Caselitz T, Tipaldi GD and Burgard W (2016{\natexlab{b}}) Rigid scene
  flow for 3d lidar scans.
\newblock In: \emph{IEEE/RSJ International Conference on Intelligent Robots and
  Systems}. IEEE, pp. 1765--1770.

\bibitem[{Fan et~al.(2022)Fan, Shen, Chen, Zhang and
  Pan}]{fan2022dynamicfilter}
Fan T, Shen B, Chen H, Zhang W and Pan J (2022) Dynamicfilter: an online
  dynamic objects removal framework for highly dynamic environments.
\newblock In: \emph{International Conference on Robotics and Automation}. IEEE,
  pp. 7988--7994.

\bibitem[{Francis et~al.(2020)Francis, Faust, Chiang, Hsu, Kew, Fiser and
  Lee}]{francis2020long}
Francis A, Faust A, Chiang HTL, Hsu J, Kew JC, Fiser M and Lee TWE (2020)
  Long-range indoor navigation with prm-rl.
\newblock \emph{IEEE Transactions on Robotics} 36(4): 1115--1134.

\bibitem[{Geiger et~al.(2012)Geiger, Lenz and Urtasun}]{geiger2012KITTI}
Geiger A, Lenz P and Urtasun R (2012) Are we ready for autonomous driving? the
  kitti vision benchmark suite.
\newblock In: \emph{IEEE Conference on Computer Vision and Pattern
  Recognition}. IEEE, pp. 3354--3361.

\bibitem[{Hornung et~al.(2013)Hornung, Wurm, Bennewitz, Stachniss and
  Burgard}]{hornung2013octomap}
Hornung A, Wurm KM, Bennewitz M, Stachniss C and Burgard W (2013) Octomap: An
  efficient probabilistic 3d mapping framework based on octrees.
\newblock \emph{Autonomous Robots} 34(3): 189--206.

\bibitem[{Huang et~al.(2022)Huang, Gojcic, Huang, Wieser and
  Schindler}]{huang2022dynamic}
Huang S, Gojcic Z, Huang J, Wieser A and Schindler K (2022) Dynamic 3d scene
  analysis by point cloud accumulation.
\newblock In: \emph{European Conference on Computer Vision}. Springer, pp.
  674--690.

\bibitem[{Kaestner et~al.(2012)Kaestner, Maye, Pilat and
  Siegwart}]{kaestner2012generative}
Kaestner R, Maye J, Pilat Y and Siegwart R (2012) Generative object detection
  and tracking in 3d range data.
\newblock In: \emph{IEEE International Conference on Robotics and Automation}.
  IEEE, pp. 3075--3081.

\bibitem[{Kim and Kim(2020)}]{kim2020remove}
Kim G and Kim A (2020) Remove, then revert: Static point cloud map construction
  using multiresolution range images.
\newblock In: \emph{IEEE/RSJ International Conference on Intelligent Robots and
  Systems}. IEEE, pp. 10758--10765.

\bibitem[{Koenig and Howard(2004)}]{Gazebo_simulator}
Koenig N and Howard A (2004) Design and use paradigms for gazebo, an
  open-source multi-robot simulator.
\newblock In: \emph{IEEE/RSJ International Conference on Intelligent Robots and
  Systems}, volume~3. pp. 2149--2154.

\bibitem[{K{\"u}mmerle et~al.(2015)K{\"u}mmerle, Ruhnke, Steder, Stachniss and
  Burgard}]{kummerle2015autonomous}
K{\"u}mmerle R, Ruhnke M, Steder B, Stachniss C and Burgard W (2015) Autonomous
  robot navigation in highly populated pedestrian zones.
\newblock \emph{Journal of Field Robotics} 32(4): 565--589.

\bibitem[{Lang et~al.(2019)Lang, Vora, Caesar, Zhou, Yang and
  Beijbom}]{lang2019pointpillars}
Lang AH, Vora S, Caesar H, Zhou L, Yang J and Beijbom O (2019) Pointpillars:
  Fast encoders for object detection from point clouds.
\newblock In: \emph{IEEE/CVF Conference on Computer Vision and Pattern
  Recognition}. pp. 12697--12705.

\bibitem[{Lim et~al.(2021)Lim, Hwang and Myung}]{lim2021erasor}
Lim H, Hwang S and Myung H (2021) Erasor: Egocentric ratio of pseudo
  occupancy-based dynamic object removal for static 3d point cloud map
  building.
\newblock \emph{IEEE Robotics and Automation Letters} 6(2): 2272--2279.

\bibitem[{Liu et~al.(2022)Liu, Gao, Liu, Liu, Meng and Gao}]{liu2022ss3d}
Liu C, Gao C, Liu F, Liu J, Meng D and Gao X (2022) Ss3d: Sparsely-supervised
  3d object detection from point cloud.
\newblock In: \emph{IEEE/CVF Conference on Computer Vision and Pattern
  Recognition}. pp. 8428--8437.

\bibitem[{Liu et~al.(2019)Liu, Qi and Guibas}]{liu2019flownet3d}
Liu X, Qi CR and Guibas LJ (2019) Flownet3d: Learning scene flow in 3d point
  clouds.
\newblock In: \emph{IEEE/CVF Conference on Computer Vision and Pattern
  Recognition}. pp. 529--537.

\bibitem[{Luiten et~al.(2021)Luiten, Osep, Dendorfer, Torr, Geiger,
  Leal-Taix{\'e} and Leibe}]{luiten2021hota}
Luiten J, Osep A, Dendorfer P, Torr P, Geiger A, Leal-Taix{\'e} L and Leibe B
  (2021) Hota: A higher order metric for evaluating multi-object tracking.
\newblock \emph{International Journal of Computer Vision} 129(2): 548--578.

\bibitem[{Manyika et~al.(2017)Manyika, Lund, Chui, Bughin, Woetzel, Batra, Ko
  and Sanghvi}]{manyika2017jobs}
Manyika J, Lund S, Chui M, Bughin J, Woetzel J, Batra P, Ko R and Sanghvi S
  (2017) Jobs lost, jobs gained: Workforce transitions in a time of automation.
\newblock \emph{McKinsey Global Institute} 150.

\bibitem[{Martin-Martin et~al.(2021)Martin-Martin, Patel, Rezatofighi, Shenoi,
  Gwak, Frankel, Sadeghian and Savarese}]{martin2021jrdb}
Martin-Martin R, Patel M, Rezatofighi H, Shenoi A, Gwak J, Frankel E, Sadeghian
  A and Savarese S (2021) Jrdb: A dataset and benchmark of egocentric robot
  visual perception of humans in built environments.
\newblock \emph{IEEE Transactions on Pattern Analysis and Machine Intelligence}
  : 1--1\doi{10.1109/TPAMI.2021.3070543}.

\bibitem[{Mersch et~al.(2022)Mersch, Chen, Vizzo, Nunes, Behley and
  Stachniss}]{mersch2022receding}
Mersch B, Chen X, Vizzo I, Nunes L, Behley J and Stachniss C (2022) {Receding
  Moving Object Segmentation in 3D LiDAR Data Using Sparse 4D Convolutions}.
\newblock \emph{IEEE Robotics and Automation Letters} 7(3): 7503--7510.

\bibitem[{Milioto et~al.(2019)Milioto, Vizzo, Behley and
  Stachniss}]{milioto2019rangenet++}
Milioto A, Vizzo I, Behley J and Stachniss C (2019) Rangenet++: Fast and
  accurate lidar semantic segmentation.
\newblock In: \emph{IEEE/RSJ International Conference on Intelligent Robots and
  Systems}. IEEE, pp. 4213--4220.

\bibitem[{Montemerlo et~al.(2008)Montemerlo, Becker, Bhat, Dahlkamp, Dolgov,
  Ettinger, Haehnel, Hilden, Hoffmann, Huhnke et~al.}]{montemerlo2008junior}
Montemerlo M, Becker J, Bhat S, Dahlkamp H, Dolgov D, Ettinger S, Haehnel D,
  Hilden T, Hoffmann G, Huhnke B et~al. (2008) Junior: The stanford entry in
  the urban challenge.
\newblock \emph{Journal of Field Robotics} 25(9): 569--597.

\bibitem[{Moosmann and Stiller(2013)}]{moosmann2013joint}
Moosmann F and Stiller C (2013) Joint self-localization and tracking of generic
  objects in 3d range data.
\newblock In: \emph{2013 IEEE International Conference on Robotics and
  Automation}. IEEE, pp. 1146--1152.

\bibitem[{Pagad et~al.(2020)Pagad, Agarwal, Narayanan, Rangan, Kim and
  Yalla}]{pagad2020robust}
Pagad S, Agarwal D, Narayanan S, Rangan K, Kim H and Yalla G (2020) Robust
  method for removing dynamic objects from point clouds.
\newblock In: \emph{IEEE International Conference on Robotics and Automation}.
  IEEE, pp. 10765--10771.

\bibitem[{Pfreundschuh et~al.(2021)Pfreundschuh, Hendrikx, Reijgwart, Dub{\'e},
  Siegwart and Cramariuc}]{pfreundschuh2021dynamic}
Pfreundschuh P, Hendrikx HF, Reijgwart V, Dub{\'e} R, Siegwart R and Cramariuc
  A (2021) Dynamic object aware lidar slam based on automatic generation of
  training data.
\newblock In: \emph{IEEE International Conference on Robotics and Automation}.
  IEEE, pp. 11641--11647.

\bibitem[{Pomerleau et~al.(2014)Pomerleau, Kr{\"u}si, Colas, Furgale and
  Siegwart}]{pomerleau2014long}
Pomerleau F, Kr{\"u}si P, Colas F, Furgale P and Siegwart R (2014) Long-term 3d
  map maintenance in dynamic environments.
\newblock In: \emph{IEEE International Conference on Robotics and Automation}.
  IEEE, pp. 3712--3719.

\bibitem[{Ristani et~al.(2016)Ristani, Solera, Zou, Cucchiara and
  Tomasi}]{ristani2016performance}
Ristani E, Solera F, Zou R, Cucchiara R and Tomasi C (2016) Performance
  measures and a data set for multi-target, multi-camera tracking.
\newblock In: \emph{European Conference on Computer Vision}. Springer, pp.
  17--35.

\bibitem[{Schauer and N{\"u}chter(2018)}]{schauer2018peopleremover}
Schauer J and N{\"u}chter A (2018) The peopleremover—removing dynamic objects
  from 3-d point cloud data by traversing a voxel occupancy grid.
\newblock \emph{IEEE Robotics and Automation Letters} 3(3): 1679--1686.

\bibitem[{Shan et~al.(2020)Shan, Englot, Meyers, Wang, Ratti and
  Rus}]{shan2020lio}
Shan T, Englot B, Meyers D, Wang W, Ratti C and Rus D (2020) Lio-sam:
  Tightly-coupled lidar inertial odometry via smoothing and mapping.
\newblock In: \emph{IEEE/RSJ International Conference on Intelligent Robots and
  Systems}. IEEE, pp. 5135--5142.

\bibitem[{Shenoi et~al.(2020)Shenoi, Patel, Gwak, Goebel, Sadeghian,
  Rezatofighi, Martin-Martin and Savarese}]{shenoi2020jrmot}
Shenoi A, Patel M, Gwak J, Goebel P, Sadeghian A, Rezatofighi H, Martin-Martin
  R and Savarese S (2020) Jrmot: A real-time 3d multi-object tracker and a new
  large-scale dataset.
\newblock In: \emph{IEEE/RSJ International Conference on Intelligent Robots and
  Systems}. pp. 10335--10342.
\newblock \doi{10.1109/IROS45743.2020.9341635}.

\bibitem[{Tanzmeister et~al.(2014)Tanzmeister, Thomas, Wollherr and
  Buss}]{tanzmeister2014grid}
Tanzmeister G, Thomas J, Wollherr D and Buss M (2014) Grid-based mapping and
  tracking in dynamic environments using a uniform evidential environment
  representation.
\newblock In: \emph{IEEE International Conference on Robotics and Automation}.
  IEEE, pp. 6090--6095.

\bibitem[{Thrun(2002)}]{thrun2002probabilistic}
Thrun S (2002) Probabilistic robotics.
\newblock \emph{Communications of the ACM} 45(3): 52--57.

\bibitem[{Urmson et~al.(2008)Urmson, Anhalt, Bagnell, Baker, Bittner, Clark,
  Dolan, Duggins, Galatali, Geyer et~al.}]{urmson2008autonomous}
Urmson C, Anhalt J, Bagnell D, Baker C, Bittner R, Clark M, Dolan J, Duggins D,
  Galatali T, Geyer C et~al. (2008) Autonomous driving in urban environments:
  Boss and the urban challenge.
\newblock \emph{Journal of Field Robotics} 25(8): 425--466.

\bibitem[{Ushani et~al.(2017)Ushani, Wolcott, Walls and
  Eustice}]{ushani2017learning}
Ushani AK, Wolcott RW, Walls JM and Eustice RM (2017) A learning approach for
  real-time temporal scene flow estimation from lidar data.
\newblock In: \emph{IEEE International Conference on Robotics and Automation}.
  IEEE, pp. 5666--5673.

\bibitem[{Wang et~al.(2007)Wang, Thorpe, Thrun, Hebert and
  Durrant-Whyte}]{wang2007simultaneous}
Wang CC, Thorpe C, Thrun S, Hebert M and Durrant-Whyte H (2007) Simultaneous
  localization, mapping and moving object tracking.
\newblock \emph{The International Journal of Robotics Research} 26(9):
  889--916.

\bibitem[{Wang et~al.(2015)Wang, Posner and Newman}]{wang2015model}
Wang DZ, Posner I and Newman P (2015) Model-free detection and tracking of
  dynamic objects with 2d lidar.
\newblock \emph{The International Journal of Robotics Research} 34(7):
  1039--1063.

\bibitem[{Wang et~al.(2022)Wang, Zhang, Kong, Zhu, Zheng, Zhuang and
  Xu}]{wang2022motion}
Wang H, Zhang L, Kong Q, Zhu W, Zheng J, Zhuang L and Xu X (2022) Motion
  planning in complex urban environments: An industrial application on
  autonomous last-mile delivery vehicles.
\newblock \emph{Journal of Field Robotics} 39(8): 1258--1285.

\bibitem[{Weng et~al.(2020)Weng, Wang, Held and Kitani}]{weng20203d}
Weng X, Wang J, Held D and Kitani K (2020) 3d multi-object tracking: A baseline
  and new evaluation metrics.
\newblock In: \emph{IEEE/RSJ International Conference on Intelligent Robots and
  Systems}. pp. 10359--10366.
\newblock \doi{10.1109/IROS45743.2020.9341164}.

\bibitem[{Wu et~al.(2020)Wu, Wang, Li, Liu and Fuxin}]{wu2020pointpwc}
Wu W, Wang ZY, Li Z, Liu W and Fuxin L (2020) Pointpwc-net: Cost volume on
  point clouds for (self-) supervised scene flow estimation.
\newblock In: \emph{European Conference on Computer Vision}. Springer, pp.
  88--107.

\bibitem[{Xu and Zhang(2021)}]{xu2021fast}
Xu W and Zhang F (2021) Fast-lio: A fast, robust lidar-inertial odometry
  package by tightly-coupled iterated kalman filter.
\newblock \emph{IEEE Robotics and Automation Letters} 6(2): 3317--3324.

\bibitem[{Yan et~al.(2017)Yan, Duckett and Bellotto}]{yan2017online}
Yan Z, Duckett T and Bellotto N (2017) Online learning for human classification
  in 3d lidar-based tracking.
\newblock In: \emph{IEEE/RSJ International Conference on Intelligent Robots and
  Systems}. IEEE, pp. 864--871.

\bibitem[{Yoon et~al.(2019)Yoon, Tang and Barfoot}]{yoon2019mapless}
Yoon D, Tang T and Barfoot T (2019) Mapless online detection of dynamic objects
  in 3d lidar.
\newblock In: \emph{Conference on Computer and Robot Vision}. IEEE, pp.
  113--120.

\bibitem[{Zhang et~al.(2020)Zhang, Hu, Chadha and Singh}]{Zhang2020Falco}
Zhang J, Hu C, Chadha RG and Singh S (2020) Falco: Fast likelihood-based
  collision avoidance with extension to human-guided navigation.
\newblock \emph{Journal of Field Robotics} 37(8): 1300--1313.

\bibitem[{Zhang and Singh(2014)}]{zhang2014loam}
Zhang J and Singh S (2014) Loam: Lidar odometry and mapping in real-time.
\newblock In: \emph{Robotics: Science and Systems}, volume~2. Berkeley, CA, pp.
  1--9.

\end{thebibliography}


\end{document}